\documentclass[11pt]{article} %
\usepackage{url}
\usepackage{smile}
\usepackage{graphicx} %
\usepackage{algorithm}
\usepackage{algorithmic}
\usepackage{todonotes}
\usepackage{epstopdf}
\usepackage{wrapfig}
\usepackage[colorlinks, linkcolor=blue, anchorcolor=blue, citecolor=blue]{hyperref}
\usepackage[margin=1in]{geometry}
\usepackage[normalem]{ulem}
\usepackage[export]{adjustbox}%
\usepackage{mathtools, cuted}
\usepackage{natbib}
\usepackage{enumerate}
\usepackage{enumitem}

\usepackage{kpfonts}
\DeclareMathAlphabet{\mathsf}{OT1}{cmss}{m}{n}

\SetMathAlphabet{\mathsf}{bold}{OT1}{cmss}{bx}{n}

\providecommand{\norm}[1]{\|#1\|}

\begin{document}

\title{\huge \bf SLA$^2$P: Self-supervised Anomaly Detection with Adversarial Perturbation\thanks{The code will be made publicly available after the acceptance of the paper for publication.}}

\author
{Yizhou Wang$^\dag$, Can Qin$^\dag$, Rongzhe Wei$^\ddag$, Yi Xu$^\dag$, Yue Bai$^\dag$ and Yun Fu$^\dag$ \\
$^\dag$ Northeastern University, $^\ddag$ Purdue University \\
\texttt{wyzjack990122@gmail.com}, \texttt{\{qin.ca, xu.yi, bai.yue\}@northeastern.edu}, \\
\texttt{wei397@purdue.edu}, \texttt{yunfu@ece.neu.edu}
}

\date{}

\maketitle

\begin{abstract}
   Anomaly detection is a fundamental yet challenging problem in machine learning due to the lack of label information. In this work, we propose a novel and powerful framework, dubbed as SLA$^2$P, for unsupervised anomaly detection. After extracting representative embeddings from raw data, we apply random projections to the features and regard features transformed by different projections as belonging to distinct pseudo classes. We then train a classifier network on these transformed features to perform self-supervised learning. Next we add adversarial perturbation to the transformed features to decrease their softmax scores of the predicted labels and design anomaly scores based on the predictive uncertainties of the classifier on these perturbed features. Our motivation is that because of the relatively small number and the decentralized modes of anomalies, {\bf 1)} the pseudo label classifier's training concentrates more on learning the semantic information of normal data rather than anomalous data; {\bf 2)} the transformed features of the normal data are more robust to the perturbations than those of the anomalies. Consequently, the perturbed transformed features of anomalies fail to be classified well and accordingly have lower anomaly scores than those of the normal samples. Extensive experiments on image, text and inherently tabular benchmark datasets back up our findings and indicate that SLA$^2$P achieves state-of-the-art results on unsupervised anomaly detection tasks consistently.
 \end{abstract}
 
 \section{Introduction}
 Anomalies, also known as outliers, are defined as “data instances that significantly deviate from the majority of data instances”~\citep{pang2020deep}. Correspondingly, anomaly detection (AD) refers to the process of finding these anomalous data points out in a data-driven fashion, which has long been a fundamental problem in machine learning and has various real-world applications, including medical health~\citep{min2017deep,khan2018review}, fraud detection~\citep{adewumi2017survey,shen2007application,fu2016credit}, cybersecurity~\citep{tan2011fast,kwon2019survey} and video surveillance~\citep{chen2015detecting,sultani2018real} etc.
 In view of whether and to what degree labels are available, anomaly detection tasks can be generally classified into three categories: {\bf 1)} Supervised anomaly detection involves training models on a labeled dataset consisting of both inliers and outliers and then applying them to test data. {\bf 2)} Semi-supervised anomaly detection (SSAD), or one-class classification, deals with the setting that the training dataset is only composed of normal data and the trained model is supposed to detect anomalous data in the testing phase. {\bf 3)} Unsupervised anomaly detection (UAD), which is the most common and challenging case, obeys the condition that solely unlabeled data with both inliers and outliers are provided and the anomaly detection technique is required to be capable of detecting the outliers~\citep{chandola2009anomaly}. These three categories consider anomalies as points that are intrinsically from distinct classes from normal points'. Besides, there exists another type of AD called out-of-distribution (OOD) detection ~\citep{liang2018enhancing,hendrycks2018deep,Hsu_2020_CVPR,liu2020energy}, which aims to distinguish samples that have disjoint distribution from training samples (usually from a different dataset).
 
 We mainly focus on the unsupervised setting, i.e., UAD, which is most widely applicable because it is costly to obtain labels for AD in many real-world scenarios~\citep{pang2020deep,chalapathy2019deep,chandola2007outlier}. UAD has wide-range applications in practice, including large-scale dataset construction, website management and
 news management etc. We must clarify that in some literatures, the so-called ``unsupervised anomaly detection'' actually refers to SSAD (e.g.,~\citep{somepalli2020unsupervised,kwon2020backpropagated,venkataramanan2019attention}) or OOD (e.g.,~\citep{schirrmeister2020understanding}) by our definition, which is beyond the scope of this paper.%

 In this work, we introduce a novel {\bf S}e{\bf L}f-supervised framework for unsupervised {\bf A}nomaly Detection using {\bf A}dversarial {\bf P}erturbation, which we name as \textit{SLA$^2$P}. With the extracted embeddings from unlabeled raw data, we project them into different subspaces by multiplicating random matrices. Despite having no explicit idea of the subspaces, we deem the transformation to be digging various unknown aspects of latent information of the data. Then we train a deep neural network (DNN) classifier on the transformed features in order to distinguish which subspace they are projected into. The training process of the classifier is equipped with early stopping technique to prevent overfitting and the reason behind this is that outliers fail to be trained well owing to their smaller population size and more various modes. The trained classifier is supposed to mostly learn useful information concerning characteristics of the normal data, hence its predictive uncertainties can be utilized to design anomaly scores, with anomalies having more uniform predictions while normal instances having sharper predictions. Such prediction distribution disparities can be further amplified by adding adversarial perturbations to these transformed features. The aim of the perturbations is to decrease the softmax scores of the predicted labels of the transformed features by the classifier. The practice boosts UAD performance empirically, and we are not aware of any other analogous approach in the anomaly detection field. To summarize, our contributions are four-fold:
 \begin{itemize}
    \item We propose a novel framework SLA$^2$P for UAD, which can be applied to image, text and inherently tabular datasets. Our approach is the first attempt to use {\bf feature-level transformation} to create pretext classification task as surrogate supervision for UAD task .
    \item We investigate UAD task by involving {\bf adversarial perturbation} technique in a {\bf self-supervised} fashion to achieve considerable performance improvement. 
    \item Our method achieves state-of-the-art (SOTA) results on $7$ challenging datasets, surpassing the current best methods by a considerable margin. Besides, it is {\bf higly robust} and can maintain excellent detection performance under particularly high or small anomaly ratios.
    \item We further theoretically analyze SLA$^2$P by illustrating the ``{\bf similarity-preserving}'' property of random projections, which bridges image-level and feature-level transformation based self-supervised methods.
 \end{itemize} %
 \section{Related Works}
 \paragraph{Unsupervised anomaly detection.} Traditional methods for UAD are mostly based on classic unsupervised learning tools, including density estimation methods~\citep{breunig2000lof,yang2009outlier,kim2012robust}, clustering methods~\citep{ester1996density,he2003discovering}, dimension reduction methods~\citep{shyu2003novel,paffenroth2018robust} and one-class support vector machine methods~\citep{scholkopf2000support,amer2013enhancing}. Thanks to the marvelous representation ability of DNNs, many reconstruction-based methods using DNNs are developed for UAD in recent years. They mainly employ deep generative models~\citep{zenati2018efficient,perera2019ocgan,lai2020novelty} or autoencoders (AE)~\citep{chen2017outlier,pidhorskyi2018generative,abati2019latent} to reconstruct data and determine abnormality of data via its reconstruction error. For instance, DAGMM~\citep{zong2018deep} feed the latent representations of the AE into a gaussian mixture model and jointly optimize them. RSRAE~\citep{lai2019robust} incorporates a robust PCA layer~\citep{lerman2018overview} into deep AEs, which is designed to project normal data into their subspace while leaving anomalies out. Sound as these carefully designed architectures seem, it is inevitable that such reconstruction-based methods focus more on low-level or element-wise error rather than high-level semantic information, as pointed out by~\cite{wang2019effective}.

 \paragraph{Anomaly detection with pretrained networks.} Transferring discriminative embeddings of pretrained nets to AD has been widely studied and achieved great success~\citep{sabokrou2018deep,rippeL2020modeling,ruff2020unifying}.~\citep{andrews2016transfer} shows that transfer-representation-learning approaches offer viable representations for AD tasks without prior knowledge of the data. Methods in~\citep{kumagai2019transfer,DBLP:conf/aaai/VercruyssenMD20} improve AD performance on target domains via transferring information of related domains.~\cite{burlina2019s} employs pretrained VGGNets~\citep{simonyan2014very} to perform novelty detection. Recently DNNs pretrained on ImageNet have been used to extract features for unsupervised anomaly detection and segmentation on images~\citep{bergmann2019mvtec,bergmann2020uninformed,bergman2020deep,venkataramanan2019attention} and videos~\citep{nazare2018pre,pang2020self}.

 \paragraph{Self-supervised learning.} Self-supervised Learning (SSL) has been an increasingly prevailing unsupervised learning method. SSL methods learn representations via designing a pretext task  between inputs and self-defined signals~\citep{bojanowski2017unsupervised,gidaris2018unsupervised} or contrastive learning~\citep{oord2018representation,chen2020simple}. Employing SSL techniques to assist semi-supervised AD has shown promising results recently~\citep{golan2018deep,bergman2020classification, tack2020csi, sehwag2021ssd}. In UAD task, E$^3$Outlier~\citep{wang2019effective} trains a discriminitive DNN via SSL and use the network outputs of the DNN to design anomaly scores. Nevertheless, such practice can only be applied to image data and require manually pre-defined transformations. By contrast, our method can be applied to both image and tabular data, and in our framework random transformations are able to exhibit promising performance.

 \paragraph{Input perturbation.}
 Input adversarial perturbation technique is first proposed by~\citep{goodfellow2014explaining} to generate adversarial data samples to fool the classifier. In OOD detection, several works utilize the opposite perturbation to the input raw data using the gradient of the maximum softmax score of the predicted label attained from pretrained network~\citep{liang2018enhancing,Hsu_2020_CVPR}.~\cite{NEURIPS2018_abdeb6f5} add perturbation to increase the proposed Mahalanobis distance-based confidence score for OOD detection. In anomaly/outlier detection area, there are few efforts in utilizing input perturbations to enhance AD performance as we can not directly utilize the softmax outputs of some pretrained network. To the best of our knowledge, we make the first attempt to wisely combine perturbation technique with self-supervised learning to address AD problems.

 \captionsetup{width=\textwidth}
 \begin{figure*}[t]
    \begin{center}
       \includegraphics[width=\linewidth]{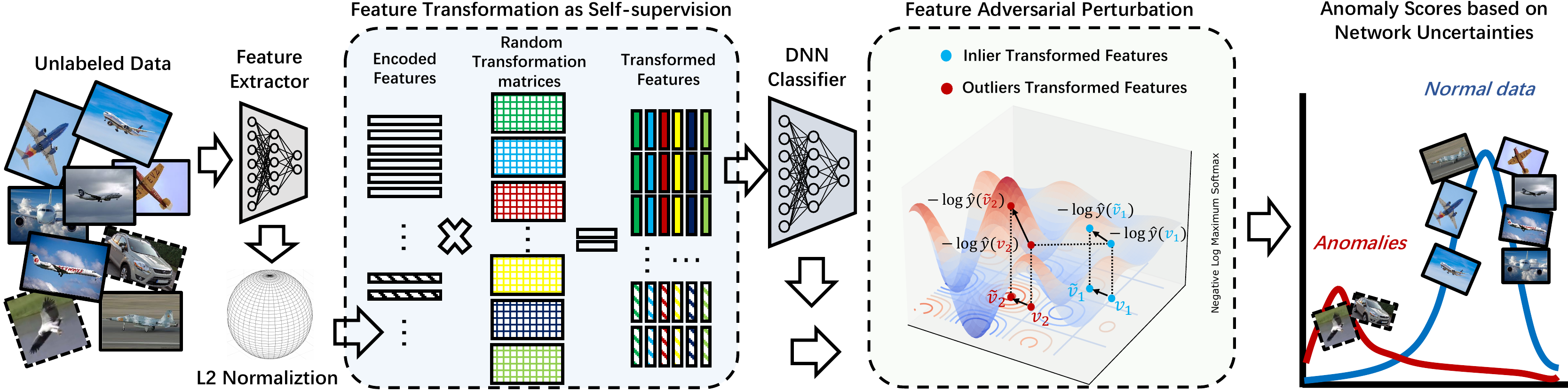}
    \end{center}
       \caption{Overview of our proposed framework SLA$^2$P on image data. Given unlabeled image data with both inliers and outliers, we first extract representative features from raw data and normalize them into a unit sphere vector space (features of anomalies are shaded). We then apply random projections to the embeddings by multiplicating matrices sampled randomly from standard normal distribution. Transformations by different matrices give rise to pseudo labels, on which we train a DNN classifier. Next we adversarially perturb the transformed samples using the gradients of the softmax scores of the predicted labels of the classifier. Finally, the anomaly scores are generated leveraging the predictive uncertainty estimates of the network on the perturbed transformed features. Best viewed in color.}
    \label{fig: structure}
 \end{figure*}
 \section{Problem Statement}
 We consider data space $\cX \subset \RR^d$ and we are given an unlabeled dataset $X \subset \cX$ . $X$ contains both normal data (inliers) $X_{\text{in}}$ of size $n$ and anomalous data (outliers) $X_{\text{out}}$ of size $np$, where $0<p<1$ is the anomaly rate defined as the ratio of anomalies to normal data. The essential objective of UAD is to design a universal indicator function $I(\xb): \cX \rightarrow \{0, 1\}$ such that $I(\xb)=1$ for $\xb \in X_{\text{in}}$ and $I(\xb)=0$ for $\xb \in X_{\text{out}}$. However, directly pursuing such function is hard and inefficient, as there is a trade-off between type-I error (rate of normal samples which are classified as anomalies) and type-II error (rate of anomalous samples that are classified as normal)~\citep{golan2018deep}. The standard practice to deal with this problem is to instead design a scoring function $S(\xb): \cX \rightarrow \RR$ s.t. higher scores indicate more normality while lower scores indicate more anomality.
 \section{Method}
 Our proposed framework SLA$^2$P is illustrated in Figure~\ref{fig: structure}, which can be partitioned into three sub-processes.
 \subsection{Feature Extraction with L2 normalization}
 Given a raw datum $\xb$, we employ a feature extractor $\fb$ to vectorize it into its discriminative embedding $\fb(\xb) \in \RR^d$. The choice of $f$ depends on the data type. We use a pretrained DNN for images and standard text pre-processing methods for text data. For inherently tabular data, the raw data is qualified to serve as its embedding, i.e., choosing $\fb$ to be identity mapping is enough. The detailed setting for different datasets is elaborated in Sec.~\ref{sec: exp}. After feature extraction, we apply L2 normalization to the embeddings:
 \begin{equation*}
    \vb(\xb) = \frac{\fb(\xb)}{\norm{\fb(\xb)}_2}.
 \end{equation*} 
 L2 normalization projects the embeddings into a sphere in $\RR^d$, which only reduces the feature dimension by one but eliminates the differences of magnitudes (norms) among the unlabeled data~\citep{gu2020spherical}. Our idea is that anomalies differ from normal data in intrinsic and latent distribution, not magnitudes. Therefore the normalization step is supposed to help the subsequent modules of SLA$^2$P to concentrate on semantic level discrepancies and attenuate low-level variances between normal and anomalous data. Our theoretical argument also corroborates its effectiveness in that the proof of our similarity perserving Theorem~\ref{thm: main} requires the L2-norm of the processed vectors to be $1$ (see Sec.~\ref{sec: theory} for more details). %
 
 \subsection{Feature-level Self-supervised Learning}
 \paragraph{Motivation from geometric transformations of images.} Geometric transformations as self-supervision have obtained tremendous success on image unsupervised learning tasks~\citep{golan2018deep,gidaris2018unsupervised,wang2019effective} and the core idea is that DNNs are considered to be able to excavate useful information of image data if they are able to recoginize their transformed patterns~\citep{gidaris2018unsupervised}. Nevertheless, the majority of data in machine learning tasks is in tabular form that does not possess the same structure as 2D images, and there are not in-hand ``geometric transformations'' for data in vector forms. In fact, geometric transformations are usually performed by applying affine transformations to homogeneous coordinates of images. Given a coordinate pair $(x,y)^\top$, it is firstly represented as a three-dimensional vector by adding an additional dimension with element 1. Then the new coordinates $(x',y')^\top$ are yielded by multiplicating a transformation matrix:
 $$
 \left ( \begin{matrix}
    x' \\
    y' \\
    1
 \end{matrix}\right )
 =\left( \begin{matrix}
    a&b&c \\
    d&e&f \\ 
    0&0&1
 \end{matrix}\right)
 \left( \begin{matrix}
    x \\
    y \\
    1
 \end{matrix}\right). 
 $$
 When $a=\cos\theta, b=\sin\theta, d=-\sin\theta, e=\cos\theta, c=f=0$, the manipulation performs rotation by angle $\theta$; when $a=e=1, b= d=0, c=x_0, f=y_0$, it performs translating image along x axis by $x_0$ and along y axis by $y_0$.
 
 Spurred by this, we propose to extend such kind of transformation to vector form data via replacing the coordinate vector with the vectors themselves. In this way, we can perform transformations in the feature-level to conduct self-supervised learning. The designing of transformation matrix still remains a problem since a rotation matrix of vectors does not have the same meanings as a rotation matrix for image 2D coordinates. Fortunately, our experiments show that utilizing random matrices with each element i.i.d. sampled from standard normal distribution is enough to exhibit outstanding results. Next, we give a detailed description of our method of using random matrices for self-supervision.

 \paragraph{Random matrix multiplications as surrogate supervision.} Let the random projection matrix set $\cA = \{\Ab^{(1)}, \Ab^{(2)}, \cdots, \Ab^{(M)}\}$. For any $1 \le m \le M$, $\Ab^{(m)}=(a_{ij}) \in \RR^{k \times d}$ ($k\ll d$) are independent and identically distributed (i.i.d.) sampled and the elements of the matrix are also i.i.d. sampled from standard normal distribution
 \begin{equation*}
    a_{ij} \sim \cN(0,1).
 \end{equation*}
 We design the random transformation to be multiplying the aforementioned random matrices to data embeddings:
 \begin{equation*}
    \vb^{(m)}= \Ab^{(m)}\vb,
 \end{equation*}
 thereby we have generated a self-labeled data set 
 \begin{equation*}
    D_{\cA} \triangleq \big\{\big(\vb^{(m)}(\xb), m\big)~\big |~\xb \in \cX, \Ab^{(m)} \in \cA\big\}.
 \end{equation*}
 Here we treat $m$ as the pseudo label of $\Ab^{(m)}\vb$, and hence we have a self-defined classification dataset with $mn(1+p)$ transformed features. The random projections not only project feature embeddings into different spaces, but also perform dimension reduction. We observe that a random matrix multiplication project one subspace or convex set in $\RR^d$ to another subspace or convex set in $\RR^k$, as claimed in Proposition~\ref{prop: project-preserve}.  Considering that extracted features of normal data tend to be in a certain subspace or convex set, Proposition~\ref{prop: project-preserve} guarantees that the transformed features of normal data are still in one subspace or convex set.

 \begin{proposition}\label{prop: project-preserve}
    For any real matrix $\Ab \in \RR^{k\times d }$, $\cV$ is a linear subspace in $\RR^d$ and $\cU$ is a convex set in $\RR^d$, then $\Ab\cV$ is a linear subspace in $\RR^k$ and $\Ab\cU$ is a convex set in $\RR^k$.
 \end{proposition}

 \paragraph{Training pseudo label classifier.} It is straightforward that we can train a multi-class classifier $C_{\btheta}$ with parameters $\btheta$ on the self-labeled dataset, i.e., the classifier is supposed to classify the transformed features $\{\vb^{(m)}(\xb_i)\}$ into pseudo-class $m$ for every $1\le m\le M$. The loss function is standard cross-entropy loss. We denote the softmax output vector of $C_{\btheta}$ as $\yb(\cdot|\btheta)$ with $m$th element $y^{(m)}(\cdot|\btheta)$ and then our self-defined pretext task can be formulated as

 \begin{equation*}
    \min_{\btheta} \frac{1}{n(1+p)}\sum_{i=1}^{n(1+p)}\cL(\xb_i|\btheta),
 \end{equation*}
 where 
 \begin{align*}
    \cL(\xb_i|\btheta) &= -\frac{1}{M}\sum_{m=1}^M \log\Big(y^{(m)}\big(\vb^{(m)}(\xb_i)|\btheta\big)\Big)\\
    & = -\frac{1}{M}\sum_{m=1}^M \log\Big(y^{(m)}\big(\Ab^{(m)}\vb(\xb_i)|\btheta\big)\Big).
 \end{align*}
 
 During the training stage, we employ an early stopping technique to refrain from overfitting. We set a hyperparameter $\mu$ as classification accuracy threshold such that the training stops as soon as the classification accuracy of the classifier in the current batch reaches $\mu \in (0,1)$. This strategy plays a vital role in our framework, since if the classifier $C_{\btheta}$ can distinguish all the transformed data the anomalous data fail to be detected easily by the classification result. %
 
 \subsection{Network Uncertainty Based Scoring with Adversarial Perturbation}
 We define our anomaly score function $S(\xb)$ based on the predictive uncertainties of the classifier $C_{\theta}$. Our goal is to make normal data have higher scores than anomalies. Instead of using the training results of the original transformed samples like~\citep{golan2018deep, wang2019effective} did, we add adversarial perturbations to the transformed features before feeding them into $C_{\theta}$.

 \paragraph{Adversarial perturbation to the transformed features.}
 We adversarially perturb the transformed features employing the gradient of the negative log softmax score of the predicted class of the trained classifier $C_{\theta}$ w.r.t. the input sample. Mathematically, for any $1\le m \le M$ and $\xb$, we let 
 \begin{equation}\label{eq: perturbation}
 \tilde{\vb}^{m}(\xb) = \vb^{(m)}(\xb) + \eta ( -\nabla_{\vb^{(m)}(\xb)} \log \hat{y}^{(m)}(\xb|\btheta)),
 \end{equation}
 where $\hat{y}^{(m)} (\xb|\btheta) = \max_{i} y^{(i)}(\vb^{(m)} (\xb)|\btheta )$, and $\eta$ is the perturbation magnitude.  Considering $C_\theta$ has been trained well on most transformed data samples, such practice aims to lower the softmax score of the pseudo class with the highest prediction probability, i.e., to make the projected features more difficult to classify. Note that our adversarial perturbation form is different from~\citep{goodfellow2014explaining} as we do not involve the pseudo labels when computing the gradient. Empirically, we discover that after such operation the anomaly scores of both inliers and outliers will decrease, but the inliers will be more robust to the perturbations than the outliers.

 \paragraph{Negative Brier Score.} Inspired by reconstruction-based AD methods~\citep{chen2017outlier,zong2018deep,lai2019robust} which employ L2 distance error of reconstruction and input as score, we utilize Euclidean distance for scoring by considering the one-hot encoding of the correct label as ground truth in the classification problem:
 \begin{equation*}
    S(\xb) = - \frac{1}{M}\sum_{m=1}^M \norm{\yb\big(\tilde{\vb}^{(m)}(\xb)|\btheta\big) - \eb_m}_2^2,
 \end{equation*}
 where $\eb_m$ is the $m$-th canonical basis vector representing the label vector of $m$-th pseudo class.
 $-S(\xb)$ refers to the mean squared error of the prediction outputs of $C_{\theta}$ and ground truth one-hot pseudo labels, which actually corresponds to a classic proper score rule known as Brier Score~\citep{brier1950verification}. The ideal case is that the normal data to have lower error and accordingly higher anomaly score.
 
 \section{Theoretical Foundation of SLA$^2$P}\label{sec: theory}
 
 Random projections do not manifestly preserve semantic information of data, which contrasts with geometric transformations ( e.g., a rotated or translated plane is still a plane, not a bird). Due to randomness, one may have no explicit idea of what subspace the features are projected into and whether the random transformations have an influence on the discrepancies among original features. Intriguingly, we find that as long as the dimension $k$ is sufficiently large, the L2 distance and inner product of transformed embeddings, which are two effective measures of similarities between high-dimensional vectors, are roughly in proportion to their original values by a constant factor $k$.
  
 \begin{theorem}[{\bf Similarity-preserving Property}]\label{thm: main}
    Given any fixed $m$ and index pair of the unlabeled dataset $(i,j)$, for any positive numbers $0<\epsilon<1$ and $0<\delta<1$, \\
    {\bf(1)} when $k>\frac{4\log\frac{2}{\delta}}{\epsilon^2-\epsilon^3}$, we have 
    with probability at least $1-\delta$ over the random sampling of matrix $\Ab^{(m)}$,
       \begin{equation*}
          (1-\epsilon)\norm{\vb_i-\vb_j}_2^2 \le \frac{\norm{\vb^{(m)}_i - \vb^{(m)}_j}_2^2}{k} \le(1+\epsilon)\norm{\vb_i-\vb_j}_2^2,
       \end{equation*}
    {\bf(2)} when $k>\frac{4\log\frac{4}{\delta}}{\epsilon^2-\epsilon^3}$, we have 
    with probability at least $1-\delta$ over the random sampling of matrix $\Ab^{(m)}$,
       \begin{equation*}
          \vb_i\cdot \vb_j -\epsilon \le \frac{\vb^{(m)}_i\cdot\vb^{(m)}_j}{k} \le \vb_i\cdot \vb_j +\epsilon. 
       \end{equation*}
 \end{theorem}
 
 The proof of Theorem~\ref{thm: main} is given in the supplementary. As stated in Theorem~\ref{thm: main}, random projections are likely to preserve the similarities among features, i.e., originally distant features ( e.g., one normal and one anomalous instance) will still be comparatively distant and features originally closely distributed ( e.g., normal data) tend to be near each other after projection. This lays the theoretical foundation of our SLA$^2$P framework in that, although the transformations are random, they still preserve the structure and inner relationship of the data points, which is coherent with handcrafted geometric transformations for 2D image data.
 
 \section{Experiments}\label{sec: exp}

 \subsection{Datasets}
 To manifest the generality and flexibility of SLA$^2$P, we empirically evaluate our framework on three image datasets and four tabular datasets. The image datasets are CIFAR-10~\citep{krizhevsky2009learning}, CIFAR-100~\citep{krizhevsky2009learning} and Caltech 101~\citep{fei2004learning}. The tabular datasets are composed of text datasets 20 Newsgroups~\citep{lang1995newsweeder}, Reuters-21578\footnote{\url{http://www.daviddlewis.com/resources/testcollections/reuters21578/}} and inherently tabular datasets Arrhythmia and KDDCUP99~\citep{Dua:2019}. For the image and text classification datasets, each class of data serves as inliers in turn, and the average result over all the classes is reported as the overall performance. The detailed descriptions are as follows.\\
 {\bf CIFAR-10} is composed of 60,000 32$\times$32 images of 10 classes, with 6,000 images per class. In each experiment, the inliers are 6,000 images from one class and the outliers are $6000\times p$ images randomly selected from the rest classes.\\
 {\bf CIFAR-100} consists of 60,000 32$\times$32 images labeled according to 100 distinct categories. The 100 categories are grouped into 20 superclasses, with 3,000 images per superclass. In each experiment, the inliers are 3,000 images from one superclass, and the outliers are $3000\times p$ images randomly chosen from the rest of superclasses.\\
 {\bf Caltech 101} contains 9,146 images of 101 different classes. The size of each image is $300\times 200$. Following the experiment setting of~\citep{lai2019robust}, we select 11 classes that contain at least 100 images and randomly sample 100 out of them for each class. In each experiment, the inliers are 100 images from one certain class and the outliers are $100\times p$ images selected from the rest 10 classes at random.\\
 {\bf 20 Newsgroups} is a text classification dataset with roughly 20,000 newsgroup documents, partitioned nearly evenly across 20 newsgroups, i.e., classes. We randomly sample 360 documents per class. In each experiment, the inliers are 360 documents from a certain class, and the outliers are $360\times p$ documents randomly chosen from the rest classes.\\
 {\bf Reuters-21578} is also a text classification dataset containing 90 text categories with multi-labels. We choose the five largest classes with single labels and sample 360 documents for each class at random. In each experiment, the inliers are the documents from a fixed class, and $360\times p $ outliers are sampled randomly from the remaining four classes. \\
 {\bf Arrhythmia} is a small-scale medical dataset  containing attributes on the diagnosis of cardiac arrhythmia in patients. The dataset contains $16$ classes. We construct the anomalous dataset using the smallest classes $3,4,5,7,8,9,14,15$ and the normal set using the rest. There are $452$ data instances in total, among which $66$ are anomalies.\\
 {\bf KDDCUP99} is a large-scale intrusion detection dataset. Following~\citep{zong2018deep} we use the entire UCI $10\%$ dataset, where the non-attack classes are treated as anomalies. There are $97278$ abnormal instances and $396743$ normal ones.

 \subsection{Experimental Setup}

 We compare SLA$^2$P with six SOTA anomaly detection methods:
 IF ~\citep{liu2008isolation}, 
 OCSVM ~\citep{scholkopf2000support,amer2013enhancing}, 
 DAGMM ~\citep{zong2018deep}, 
 E$^3$Outlier~\citep{wang2019effective}, 
 RSRAE ~\citep{lai2019robust}
 and PANDA~\citep{reiss2021panda}. We adopt the commonly used Area under the Receiver Operating Characteristic curve (AUROC) and Area under the Precision-Recall curve (AUPR) as evaluation metric. For AUPR, we consider outliers as ``positive'' when computing. All the experiments, including baseline methods, are run $5$ times independently using the same random seeds $0,1,2,3, 4$ for reproducibility and fair comparison and the averaged scores are reported in the main paper. The source code and more experimental details (including each single experiment result) are provided in the supplementary.

 \paragraph{Setup of SLA$^2$P.} For image data, we use ResNets~\citep{he2016deep} pretrained on ImageNet without the last fully connected layer to extract embeddings. To match the image input shapes, we first resize raw images to $224 \times 224$ using Bilinear Interpolation and then feed it into the extractor network. For text data, we apply TFIDF transformer and Hashing-vector~\citep{rajaraman2011mining} sequentially to pre-process raw data into vectors, which is the same process as in~\citep{lai2019robust}. For inherently tabular data, we directly use raw data vectors as the input. On Reuters dataset, we set $M=512$ and $k=128$ due to high dimensionality. On KDDCup99 dataset, we set $M=64$ and $k=128$ because of high computational burden. On the rest of the datasets, we set $M=256$ and $k=256$. For the early stopping threshold, we set $\mu=0.75$ for 20 Newsgroups, $\mu=0.3$ for Reuters and $\mu=0.6$ for all the rest datasets. We choose the perturbation magnitude as $\eta=10$ for 20 Newsgroups, $\eta=1e2$ for Reuters, $\eta=1e4$ for CIFAR-100 and $\eta=1e3$ for all the rest datasets. In fact, setting $\mu=0.6$ and $\eta=1e3$ is universally adequate for different UAD tasks and there is an implicit relationship between these two hyperparameters (see ablation study). Assuming the input dimension of classifier network is $q$, for all the experiments, the classifier network is a $3$-layer fully connected network with structure FC($q$, $2q$)-FC($2q$, $4q$)-FC($4q$, $M$). Batch normalization~\citep{ioffe2015batch} is applied to each layer and LeakyReLU~\citep{xu2015empirical} is employed as activation functions. We use Adam~\citep{kingma2014adam} to optimize the classifier network parameters with learning rate $1e$-$3$ and weight decay $5e$-$4$.

 \paragraph{Setup of baseline methods.} Our experiments indicate that all the benchmark methods, except E$^3$outlier (which directly manipulates images), works better using extracted representative embeddings as input than using raw data as input. Hence for fair comparison, we compare our method to the baseline methods with extracted features as input in the sequel. For the implementation of the benchmarks, we adapt the code from package scikit-learn~\citep{pedregosa2011scikit} for IF and  OCSVM. We implemented DAGMM and E$^3$Outlier using the code\footnote{\url{https://github.com/demonzyj56/E3Outlier}} from~\citep{wang2019effective} with minimal modifications such that they are applicable to our datasets and adapt to our experimental protocal. RSRAE results are obtained running the official public code~\footnote{\url{https://github.com/dmzou/RSRAE}}. We adapt PANDA for UAD setting by setting the training set and the testing set to be the same with both inliers and outliers using the official public code~\footnote{\url{https://github.com/talreiss/PANDA}}. All the data preprocessing processes follow the original papers.

 \subsection{Results}
 \setlength{\tabcolsep}{3pt}
 \captionsetup{width=0.95\textwidth}
 \begin{table*}[t]
    \centering
    \small
    \scalebox{0.8}{
    \begin{tabular}{cccccccccc}
    \toprule
    Dataset   & p  & IF  &  OCSVM & DAGMM & E$^3$Outlier & RSRAE & PANDA &  SLA (ours) & SLA$^2$P (ours) \\ \midrule 
                & 0.1& 83.24~/~43.55 & 85.19~/~47.57 & 59.86~/~15.73 & 85.89~/~45.09 & 83.58~/~42.48 &86.69~/~59.43& \underline{90.81}~/~\underline{59.46} & {\bf 91.56}~/~{\bf 62.91}  \\
    CIFAR-10   & 0.3& 79.36~/~56.95 & 79.52~/~56.69 & 67.04~/~37.81& 80.58~/~58.53 & 75.05~/~50.90 &83.63~/~64.20& \underline{87.68}~/~{\bf 70.08} & {\bf 87.69}~/~\underline{70.03}\\
                & 0.5& 75.94~/~62.25 & 74.83~/~60.04 & 62.73~/~44.84& 77.44~/~64.26 & 71.27~/~56.92 &82.75~/~68.20& \underline{84.64}~/~\underline{73.25} & {\bf 84.71}~/~{\bf 73.66}\\ \hline
                & 0.1& 77.35~/~33.35 & 79.98~/~35.58 & 54.47~/~12.09 & 80.36~/~33.61 & \underline{89.21}~/~\underline{55.94} &84.70~/~58.33& 87.51~/~52.13 & {\bf 91.45}~/~{\bf 71.70}\\
    CIFAR-100  & 0.3& 73.90~/~49.65 & 74.86~/~49.62 & 56.78~/~29.59 & 78.60~/~53.15 & \underline{85.33}~/~\underline{66.40} &82.32~/~64.21& 85.03~/~66.39 & {\bf 85.48}~/~{\bf 69.29}\\
                & 0.5& 70.81~/~56.51 & 71.21~/~55.96 & 55.13~/~38.16 & 76.22~/~61.00 & 81.68~/~69.46 &77.83~/~68.88& {\bf 82.48}~/~{\bf 71.21} & \underline{81.98}~/~\underline{70.76}\\ \hline
                & 0.1& 89.88~/~59.72 & 92.98~/~67.16 & 72.61~/~34.02& 90.72~/~62.29 & 94.28~/~74.44 &96.32~/~80.39& \underline{96.36}~/~\underline{80.98} & {\bf 96.42}~/~{\bf 81.12}\\
    Caltech 101& 0.3& 87.12~/~69.86 & 86.80~/~69.40 & 73.53~/~49.20& 87.37~/~70.97& 87.52~/~69.94 &94.67~/~85.72& \underline{95.09}~/~\underline{86.17} & {\bf 95.35}~/~{\bf 87.70}\\
                & 0.5& 84.29~/~72.97 & 81.67~/~69.82 & 69.14~/~53.70& 84.82~/~73.93& 82.59~/~71.78  &\underline{93.97}~/~\underline{88.91}& 93.57~/~87.73 & {\bf 96.26}~/~{\bf 94.31}\\ 
    \bottomrule
    \end{tabular}
    }
    \caption{AUROC/AUPR (\%) results on image datasets for UAD with ResNet-50 as the feature extractor. The best in bold and the second best underlined. The mean scores over $5$ independent runs are reported.} 
    \label{table: image-res50}
 \end{table*}
 We report average AUROC and AUPR of all the methods on all the datasets. To exhibit the generality of our approach, we present its performance under varying anomaly ratios $p=10\%, 30\%, 50\%$. We name our approach without adversarial perturbation as SLA and include its performance as well. Results on image datasets using pretrained ResNet-50 network are given in Tab.~\ref{table: image-res50}. The results using pretrained ResNet-101 are provided in the supplementary for completeness, where slightly better performances are achieved owing to the usage of deeper pretrained DNN extractor. We also summarize the results on text datasets and inherently tabular datasets in Tables~\ref{table: text} and~\ref{table: inherently} respectively.

 \paragraph{Peformance on image datasets.} As indicated in Tab.~\ref{table: image-res50}, the performance of our two proposed methods, especially SLA$^2$P, consistently outperform existing methods on all the image datasets, including CIFAR-10 and CIFAR-100, which are abidingly considered the most challenging benchmark datasets for UAD~\citep{wang2019effective}. In addition, SLA$^2$P is robust to the anomaly ratio and can still attain exceptional performance when $p$ is large, with AUROC $84.71\%$ on CIFAR-10, $82.48\%$ on CIFAR-100 and $96.26\%$ on Caltech 101 when the number of anomalies reaches half the number of normal samples. This is commendable as when the number of anomalies increase, no matter what UAD method is used, it is inevitable that the adverse effect of the training of outliers to the training of inliers will increase accordingly~\citep{lai2019robust,lai2020novelty}. 
 \begin{table*}[t]
   \centering
   \scalebox{0.85}{
   \begin{tabular}{ccccccccc}
   \toprule
    Dataset & p & IF  &  OCSVM & DAGMM & RSRAE  &  SLA (ours) &SLA$^2$P (ours) \\
   \midrule
    \multirow{3}{*}{20 Newsgroups} & 0.1 & 56.45~/~15.09 & 82.04~/~43.82 & 54.84~/~12.76 & 89.03~/~49.79 & \underline{90.77}~/~\underline{50.57} & {\bf 93.20}~/~{\bf 71.77}  \\
     & 0.3 & 58.35~/~34.56 & 75.07~/~52.28 & 54.07~/~27.38 & \underline{88.86}~/~\underline{68.61} & 86.29~/~63.58 & {\bf 90.16}~/~{\bf 74.21} \\
     & 0.5 & 57.00~/~43.77 & 71.33~/~58.49 & 51.94~/~35.93 & {\bf 86.73}~/~\underline{73.55}  & 84.48~/~71.18 & \underline{86.55}~/~{\bf 78.13} \\
    \hline
    \multirow{3}{*}{Reuters-21578}  & 0.1 & 59.30~/~18.74 & 90.95~/~61.69 & 65.07~/~22.33 & 92.44~/~64.44  & \underline{93.51}~/~\underline{64.58} & {\bf 95.77}~/~{\bf 81.40} \\
     & 0.3 & 58.86~/~35.99 & 83.24~/~63.25 & 59.33~/~33.61 & \underline{88.70}~/~\underline{73.07}  & 88.07~/~71.95 & {\bf 95.52}~/~{\bf 89.68}  \\
     & 0.5 & 55.85~/~42.10 & 74.88~/~62.36 & 62.01~/~46.54 & \underline{83.39}~/~\underline{72.93}  & 82.87~/~72.64 & {\bf 86.09}~/~{\bf 79.92} \\
    \bottomrule
   \end{tabular}
   }
   \caption{AUROC/AUPR (\%) results on text datasets for UAD. The mean scores over $5$ independent runs are reported.}
   \label{table: text}
\end{table*}

\captionsetup{width=0.\textwidth}
\begin{table}[t]
   \centering
   \scalebox{0.95}{
   \begin{tabular}{cccccc}
   \toprule
    & \multicolumn{2}{c}{Arrhythmia}  && \multicolumn{2}{c}{KDDCUP99}   \\
    \cmidrule{2-3} \cmidrule{5-6}
    & AUROC (\%) & AUPR (\%) && AUROC (\%) & AUPR (\%) \\
    \midrule
    IF& 79.78$_{\pm 0.93}$ & 46.29$_{\pm 1.30}$ && \underline{92.72$_{\pm 1.17}$} & 69.06$_{\pm 3.91}$ \\
    OCSVM& 79.47$_{\pm 0.00}$  & \underline{47.75$_{\pm 0.00}$}  && 51.98$_{\pm 0.00}$ & 43.23$_{\pm 0.00}$ \\
    DAGMM& 67.16$_{\pm 4.14}$ & 30.68$_{\pm 1.98}$ && 77.55$_{\pm 5.90}$ & 47.28$_{\pm 5.80}$ \\
    RSRAE& \underline{80.31$_{\pm 2.89}$}  & 42.03$_{\pm 2.84}$ && 60.09$_{\pm 2.30}$ & 27.36$_{\pm 1.22}$  \\
    SLA (ours)& 79.85$_{\pm 0.69}$ & 46.78$_{\pm 1.22}$ && 97.00$_{\pm 0.46}$ & \underline{88.89$_{\pm 0.92}$}  \\
    SLA$^2$P (ours)& {\bf 98.93}$_{\pm \bf 0.08}$ & {\bf 90.59}$_{\pm \bf 0.66}$ && {\bf 97.59}$_{\pm \bf 0.34}$  & {\bf 90.59}$_{\pm \bf 0.86}$  \\
   \bottomrule
   \end{tabular}
   }
   \caption{UAD performance on inherently tabular datasets.}
   \label{table: inherently}
\end{table}

 \paragraph{Peformance on tabular datasets.} Besides image datasets, our framework is also able to exhibit superior performance on tabular datasets as demonstrated in Tables~\ref{table: text} and~\ref{table: inherently}. SLA$^2$P achieves SOTA results on 20 Newsgroups, significantly outperforming current SOTA RSRAE by  $3\%$-$4\%$ in AUROC and around $10\%$ in AUPR for different anomaly ratios. On Reuters, our method works much better than current SOTA, with up to $6.8\%$ gain in AUROC and $16.6\%$ gain in AUPR when $p=0.3$ over RSRAE. On inherently tabular datasets, SLA$^2$P exhibits dominating performance with over $40 \%$ AUPR gain on Arrhythmia, over $20\%$ AUPR gain on KDDCUP99, and the AUROCs and AUPRs of these two classical datasets all exceed $90\%$. The extraordinarily excellent performance of SLA$^2$P on the large-scale dataset KDDCUP99 reveals its practicality .

 \captionsetup[sub]{font=tiny}
 \captionsetup{width=\textwidth}
 
 \begin{figure*}[t]
    \centering
    \begin{subfigure}[b]{0.2\linewidth}
       \includegraphics[width=\linewidth]{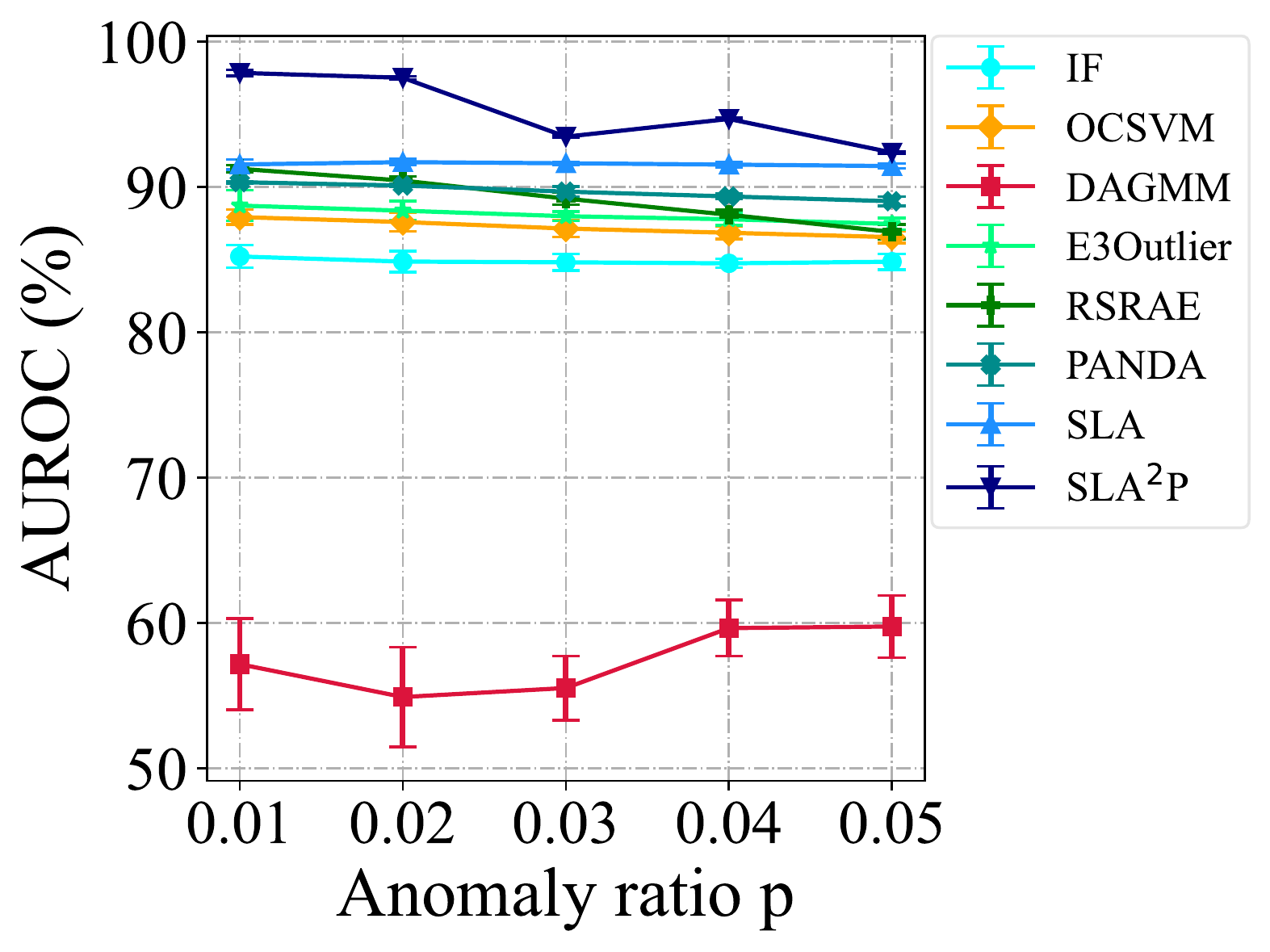}
       \caption{AUROC on CIFAR-10.}
    \end{subfigure}%
    \hfill
    \begin{subfigure}[b]{0.2\linewidth}
       \includegraphics[width=\linewidth]{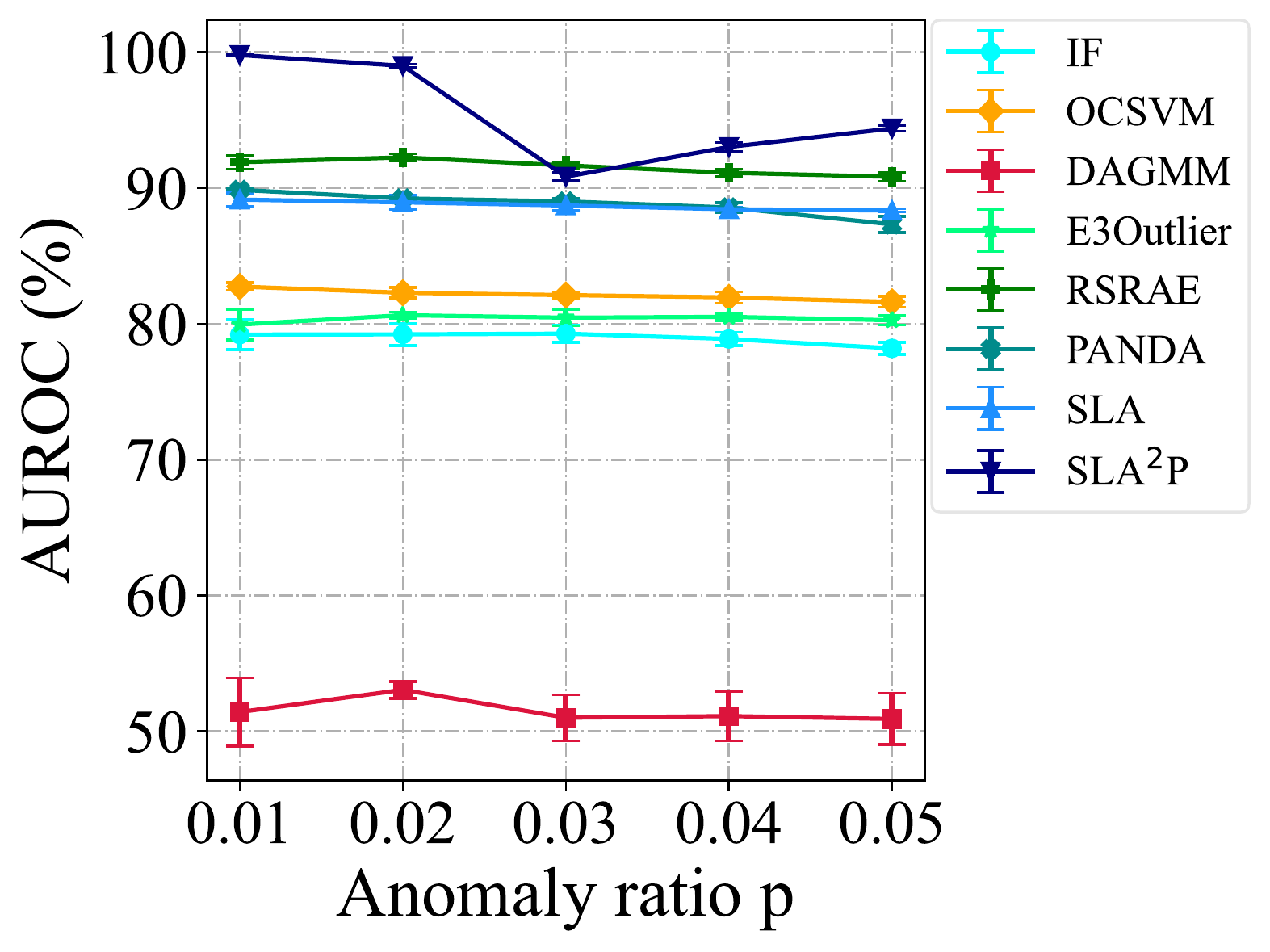}
       \caption{AUROC on CIFAR-100.}
    \end{subfigure}%
    \hfill
    \begin{subfigure}[b]{0.2\linewidth}
       \includegraphics[width=\linewidth]{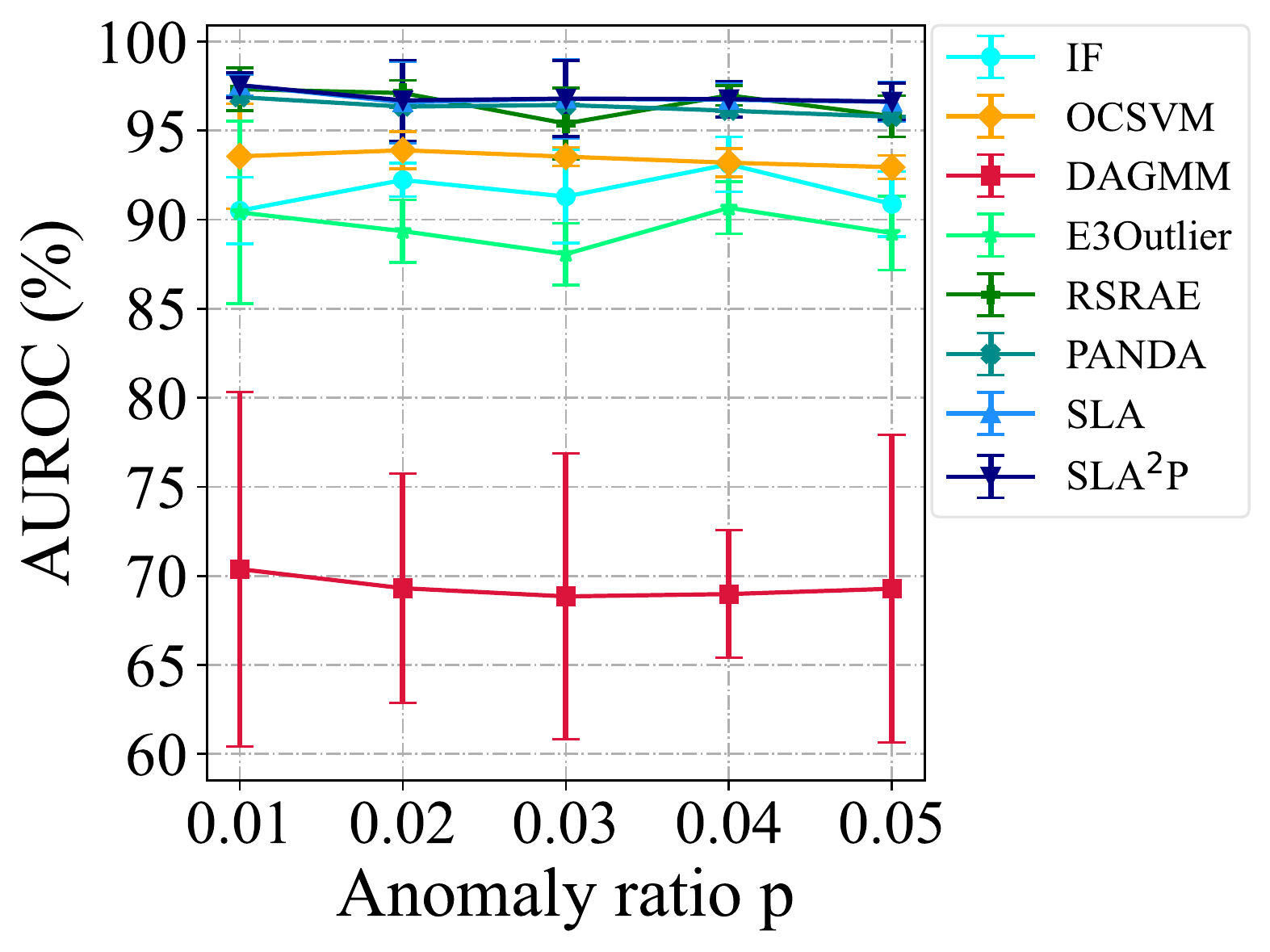}
       \caption{AUROC on Caltech.}
    \end{subfigure}%
    \hfill
    \begin{subfigure}[b]{0.2\linewidth}
       \includegraphics[width=\linewidth]{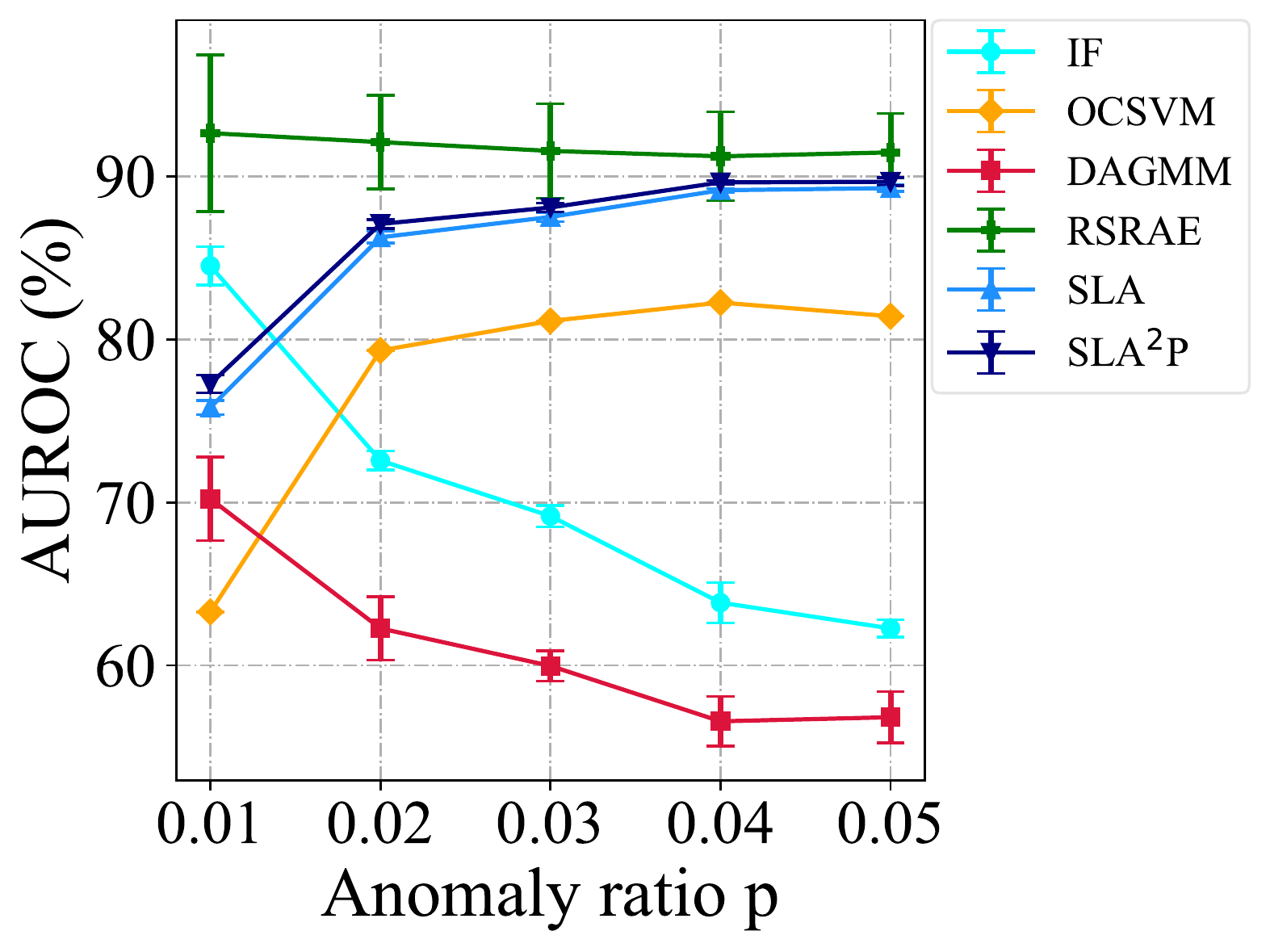}
       \caption{AUROC on 20 Newsgroups.}
    \end{subfigure}%
    \hfill
    \begin{subfigure}[b]{0.2\linewidth}
       \includegraphics[width=\linewidth]{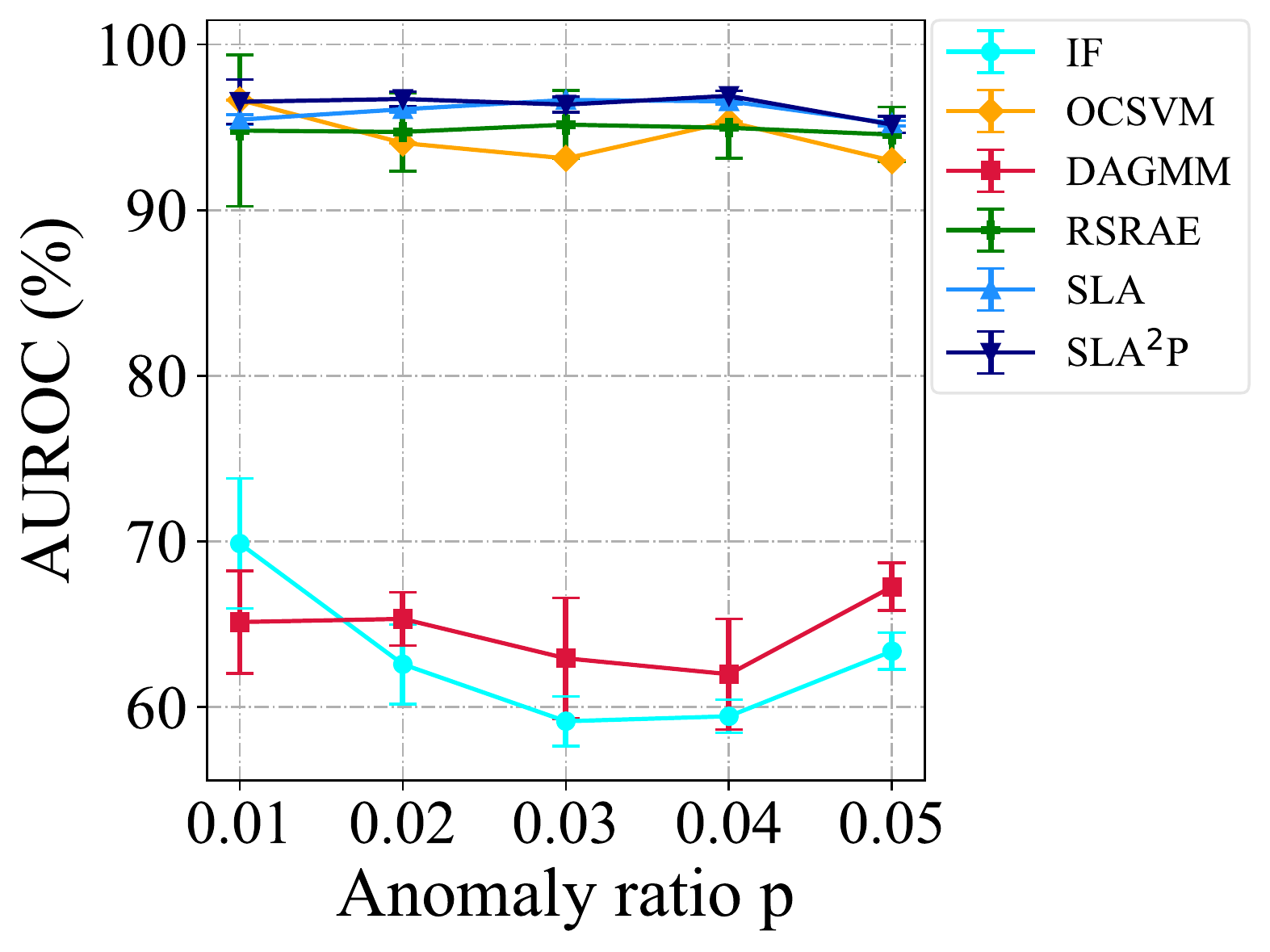}
       \caption{AUROC on Reuters.}
    \end{subfigure}%
    \quad
    \begin{subfigure}[b]{0.2\linewidth}
       \includegraphics[width=\linewidth]{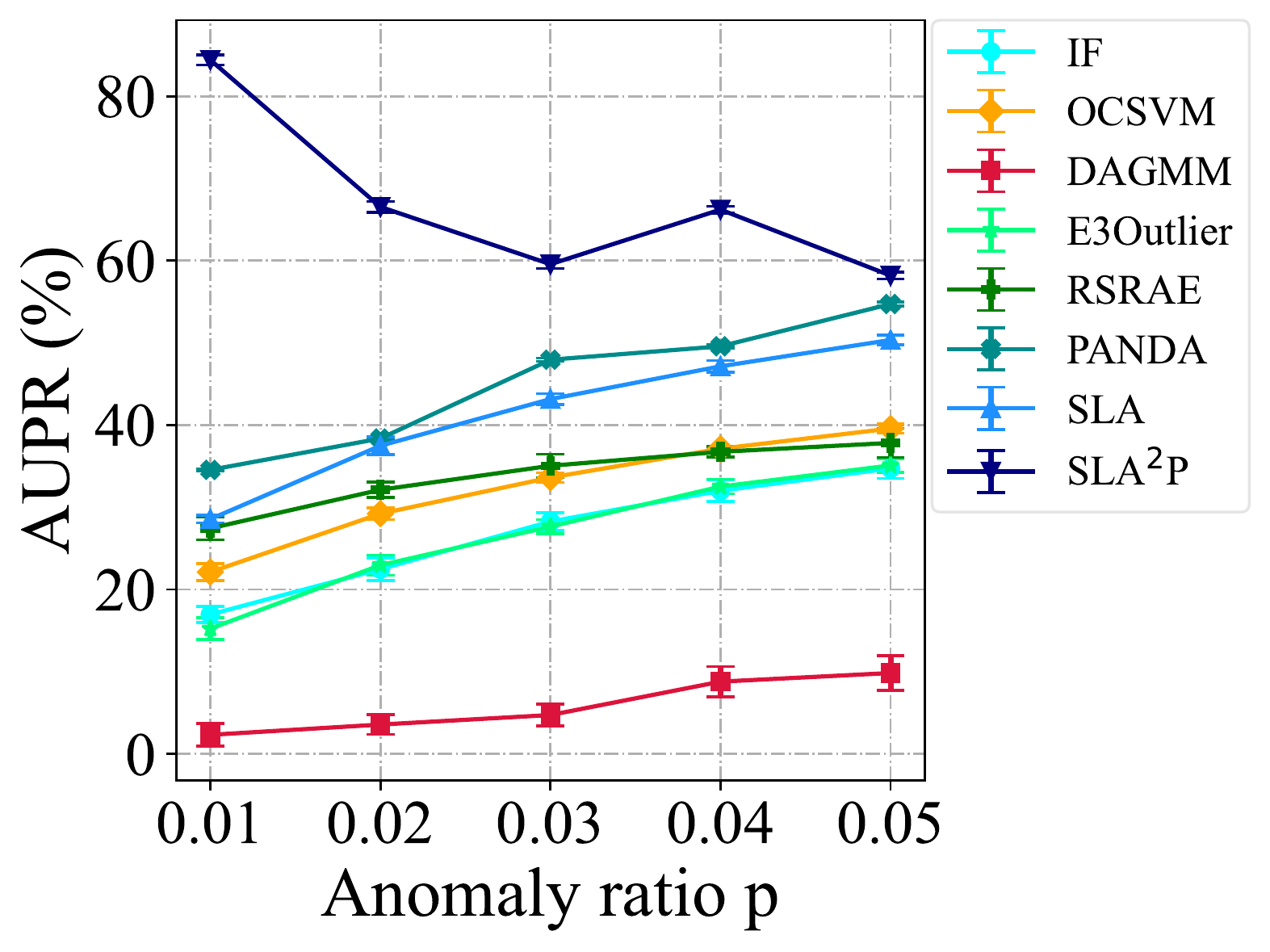}
       \caption{AUPR on CIFAR-10.}
    \end{subfigure}%
    \hfill
    \begin{subfigure}[b]{0.2\linewidth}
       \includegraphics[width=\linewidth]{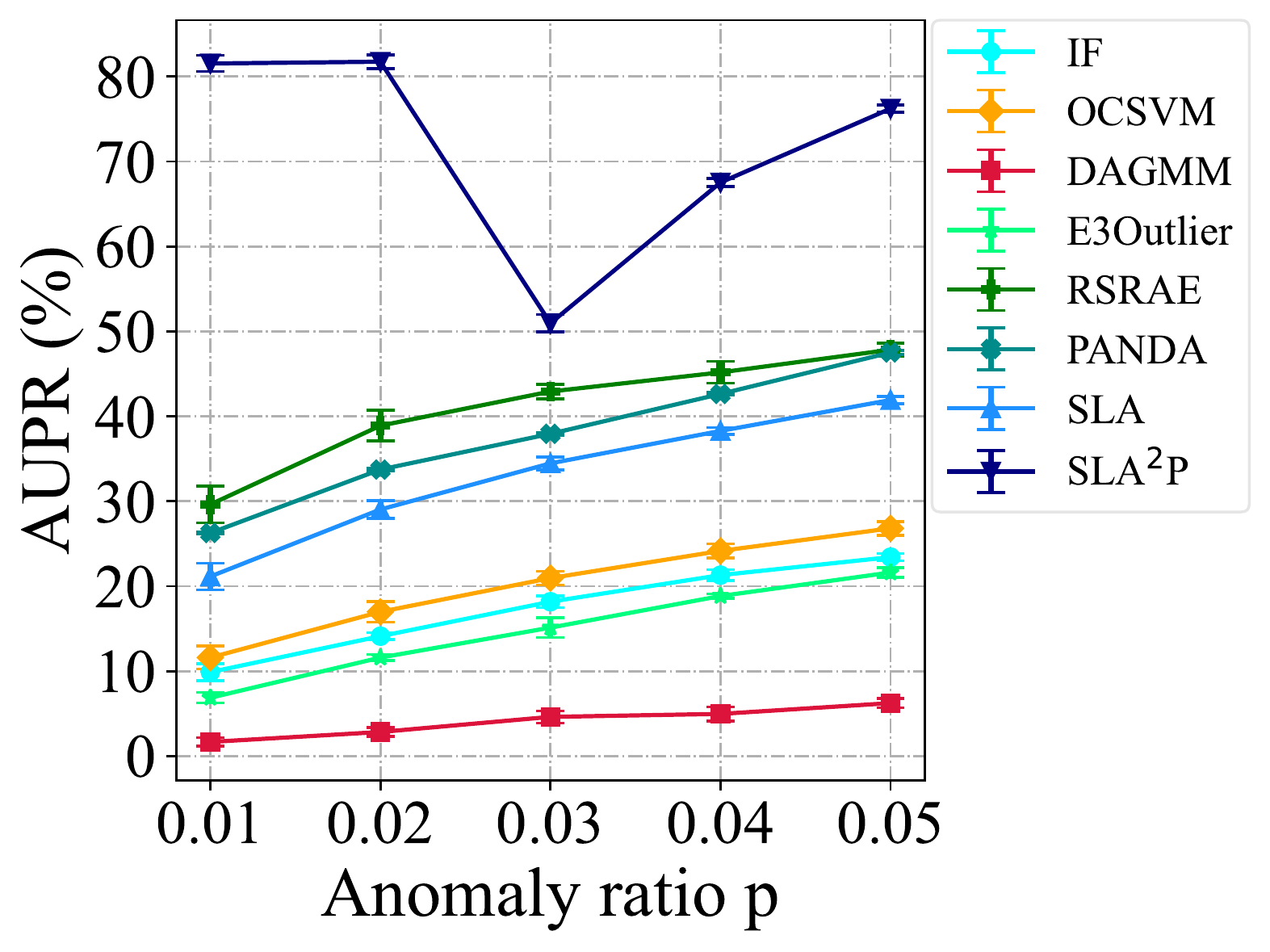}
       \caption{AUPR on CIFAR-100.}
    \end{subfigure}%
    \hfill
    \begin{subfigure}[b]{0.2\linewidth}
       \includegraphics[width=\linewidth]{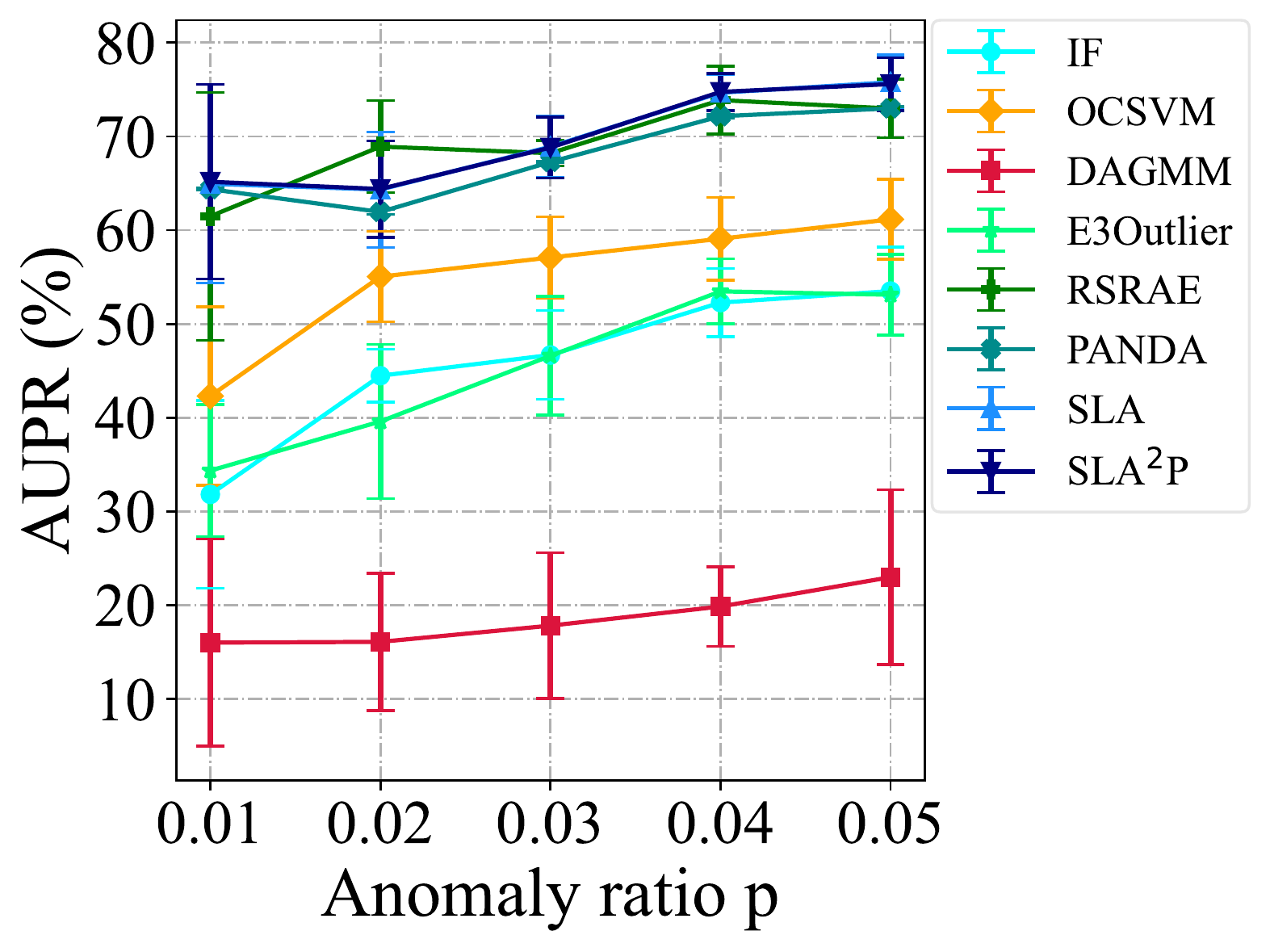}
       \caption{AUPR on Caltech.}
    \end{subfigure}%
    \hfill
    \begin{subfigure}[b]{0.2\linewidth}
       \includegraphics[width=\linewidth]{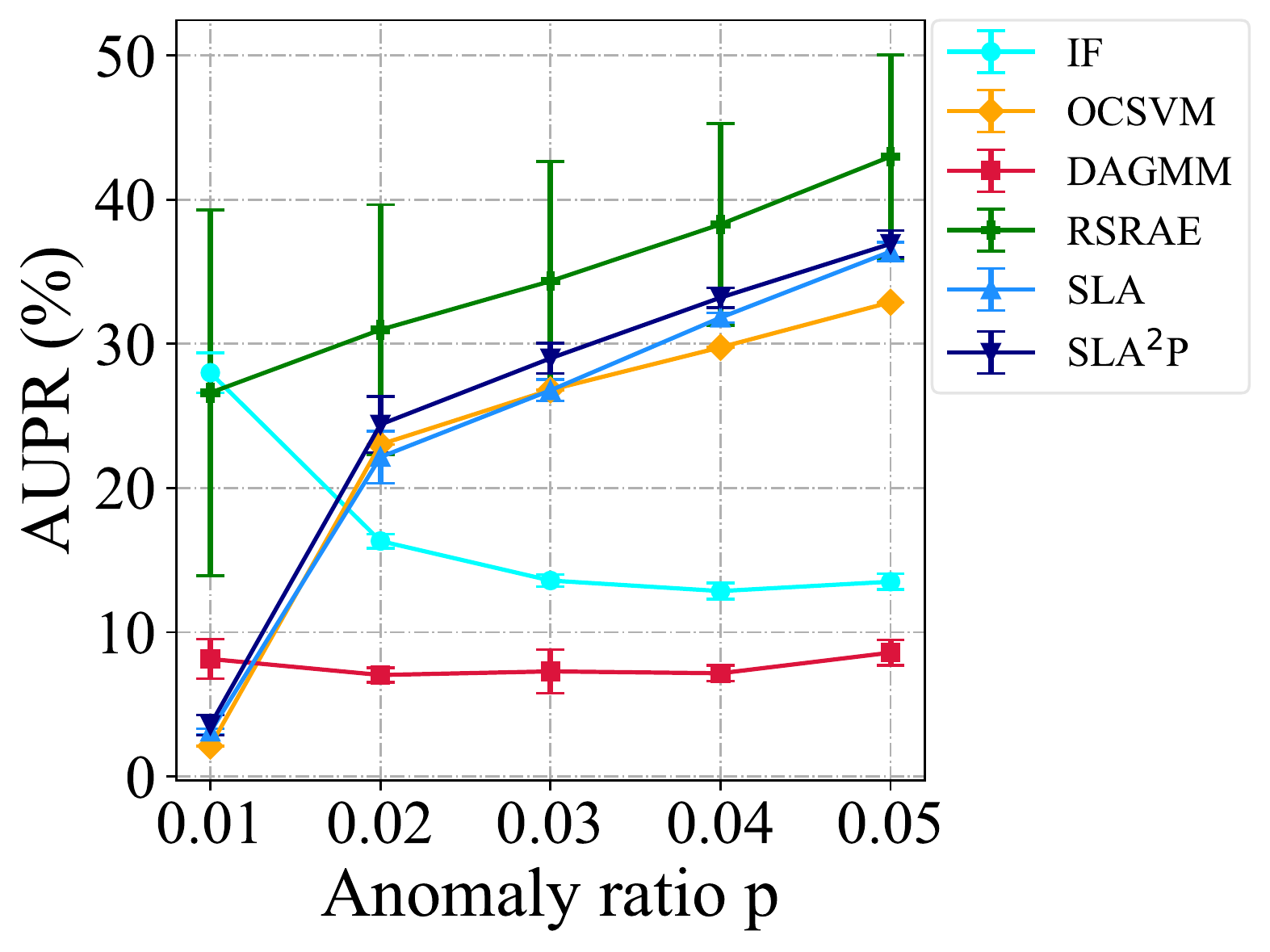}
       \caption{AUPR on 20 Newsgroups.}
    \end{subfigure}%
    \hfill
    \begin{subfigure}[b]{0.2\linewidth}
       \includegraphics[width=\linewidth]{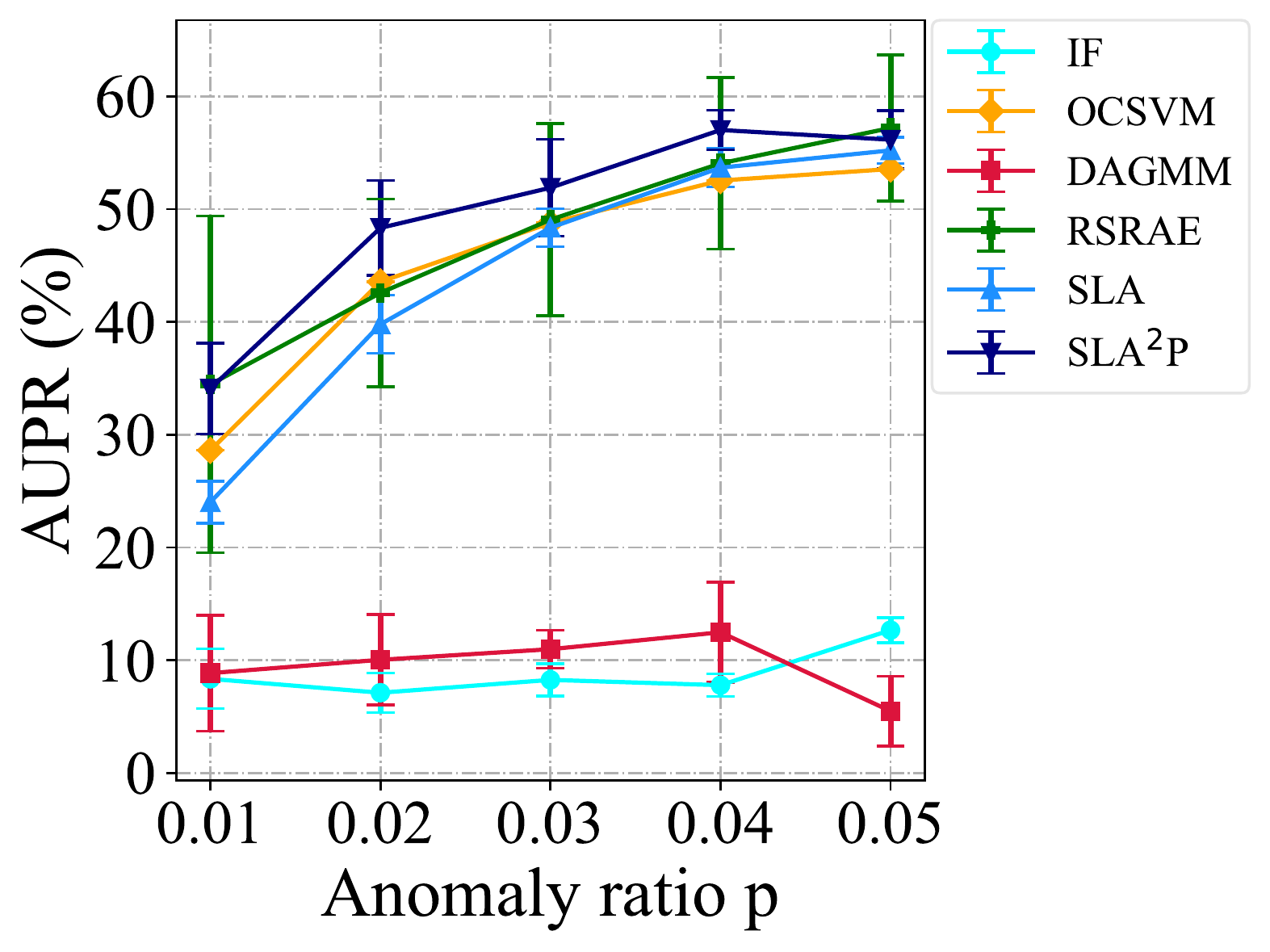}
       \caption{AUPR on Reuters.}
    \end{subfigure}%
    \caption{AUROC and AUPR scores with standard deviation bars when the anomaly ratio is tiny. Best viewed in color.}
    \label{fig: tiny}
 \end{figure*}
 \paragraph{Peformance under tiny anomaly ratios.} The anomaly ratio can be extremely small in some UAD tasks and circumstances. Therefore we further compare our methods to the baseline methods when the anomaly rates are tiny: $p=1\%, 2\%, 3\%, 4\%, 5\%$. Clearly seen from Fig.~\ref{fig: tiny}, our SLA$^2$P still achieves favorable performance compared to other baseline methods. For all the aforementioned image and text datasets except 20 Newsgroups, SLA$^2$P dominates in AUROC and AUPR with minimal numerical deviations. Furthermore, as demonstrated by the steadiness of the SLA$^2$P curve in Fig.~\ref{fig: tiny}, SLA$^2$P is remarkably stable with the alternation of $p$, especially in AUROC, despite the fact that we require sampling random matrices in our pipeline.

 \subsection{Ablation Study and Sensitivity Analysis}\label{subsec: ablation}

 \captionsetup[sub]{font=tiny}
 \begin{figure*}[t]
    \centering
    \begin{subfigure}[b]{0.19\linewidth}
       \includegraphics[width=\linewidth]{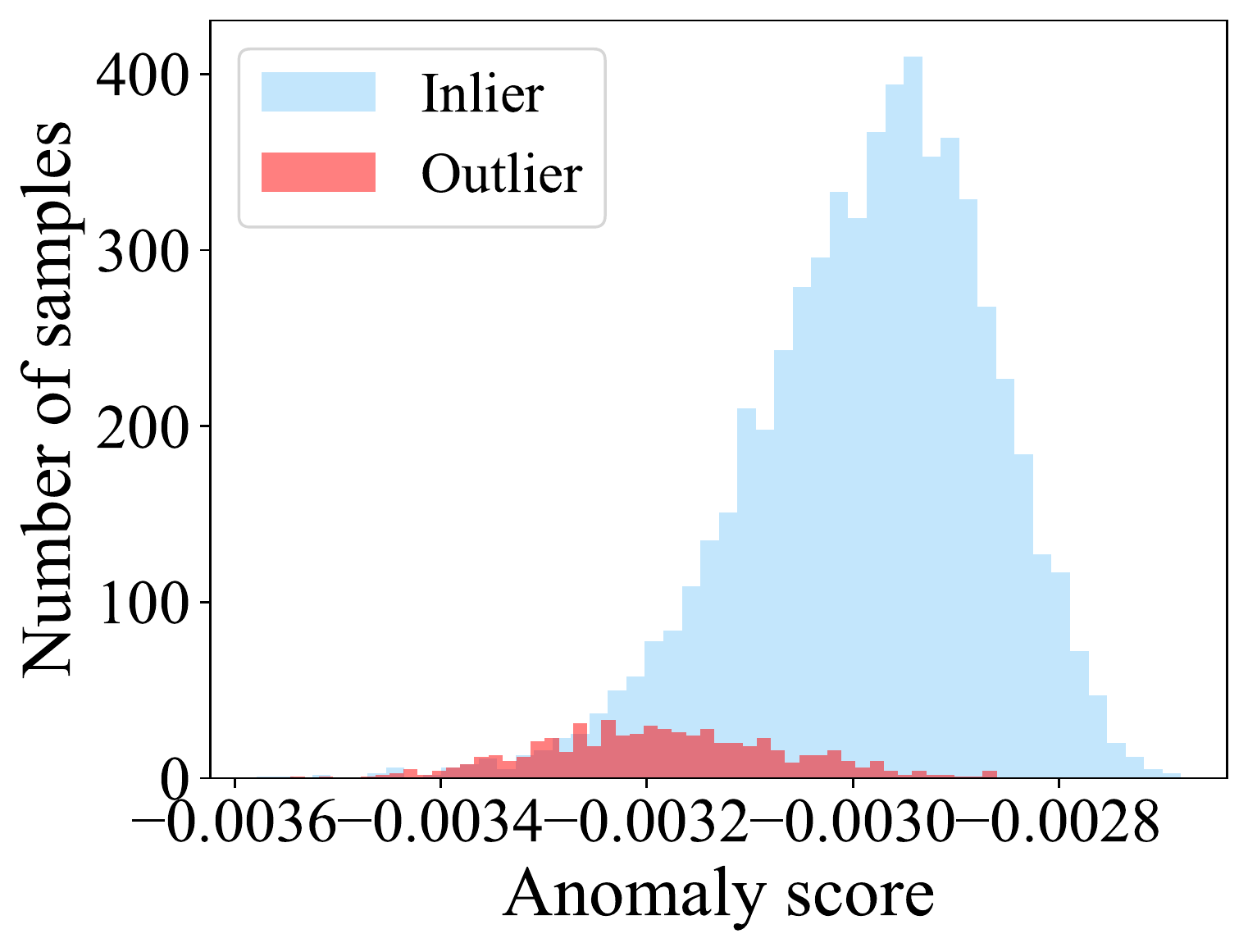}
       \caption{CIFAR-10, inlier 'cat' \protect\\ (w/o perturbation).}
    \end{subfigure}%
    \hfill
    \begin{subfigure}[b]{0.19\linewidth}
       \includegraphics[width=\linewidth]{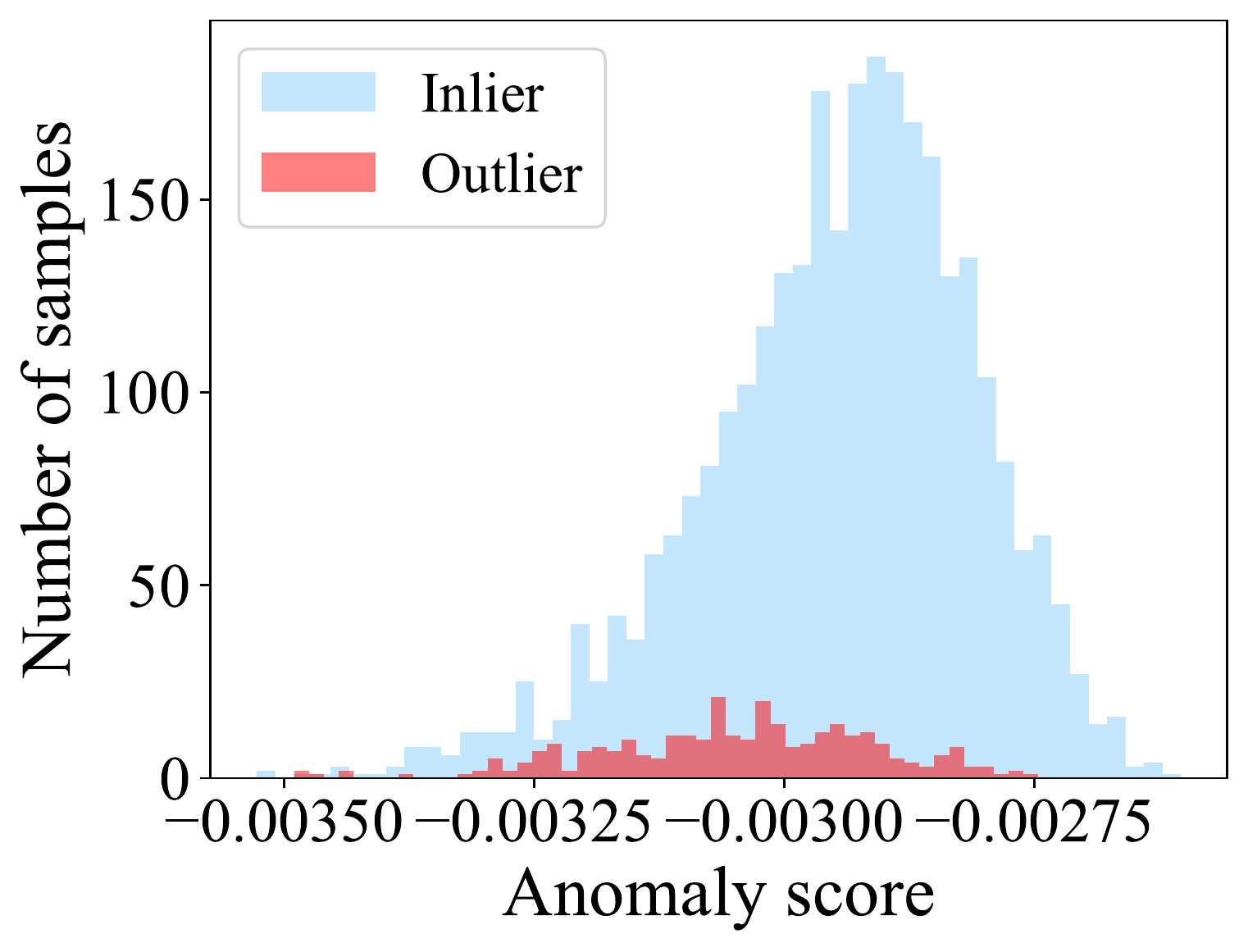}
       \caption{CIFAR-100, inlier 'non-insect invertebrates' (w/o perturbation).}
    \end{subfigure}%
    \hfill
    \begin{subfigure}[b]{0.19\linewidth}
       \includegraphics[width=\linewidth]{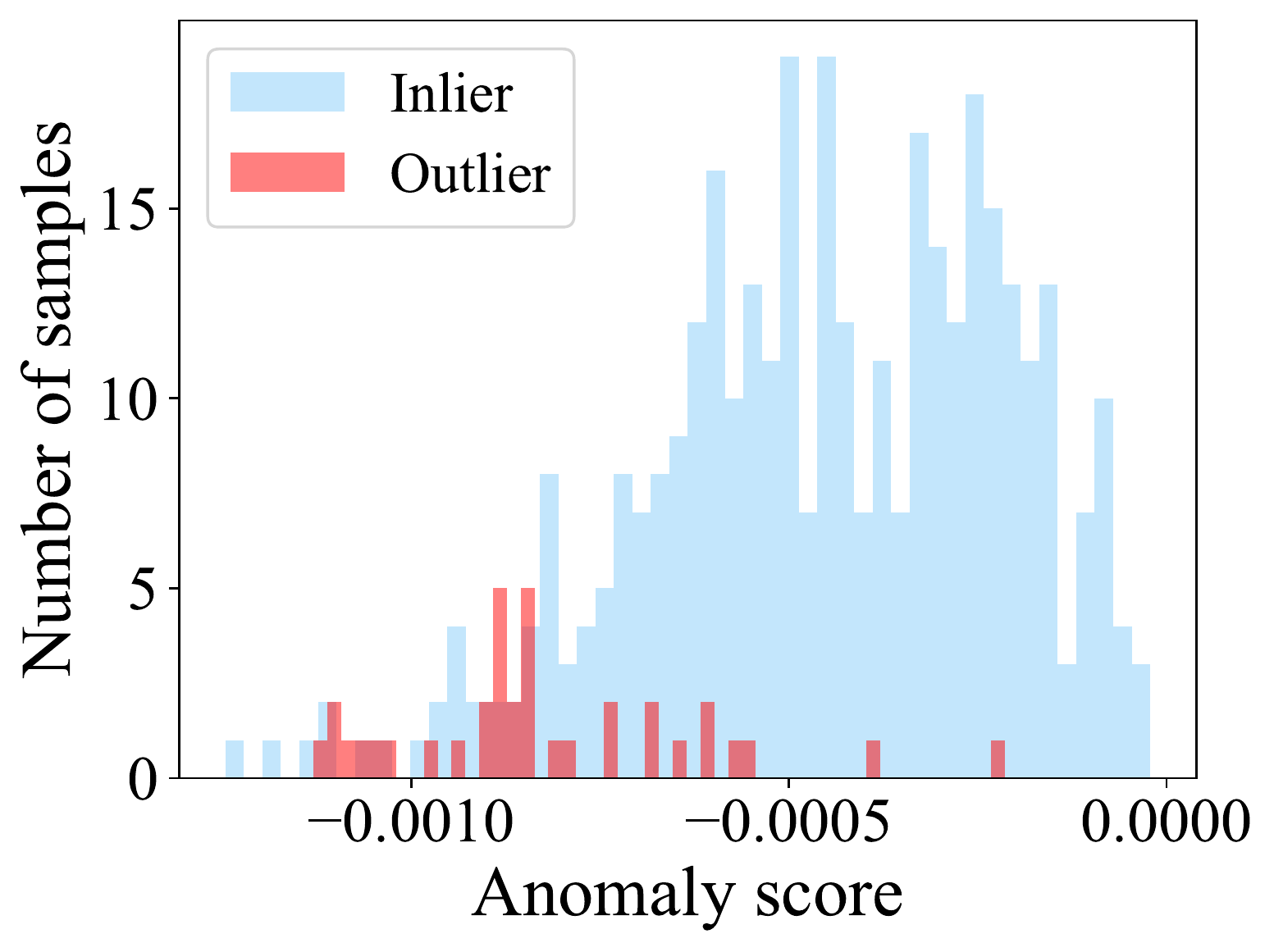}
       \caption{20, 'comp.sys.mac.hardware' \protect\\(w/o perturbation).}
    \end{subfigure}%
    \hfill
    \begin{subfigure}[b]{0.19\linewidth}
       \includegraphics[width=\linewidth]{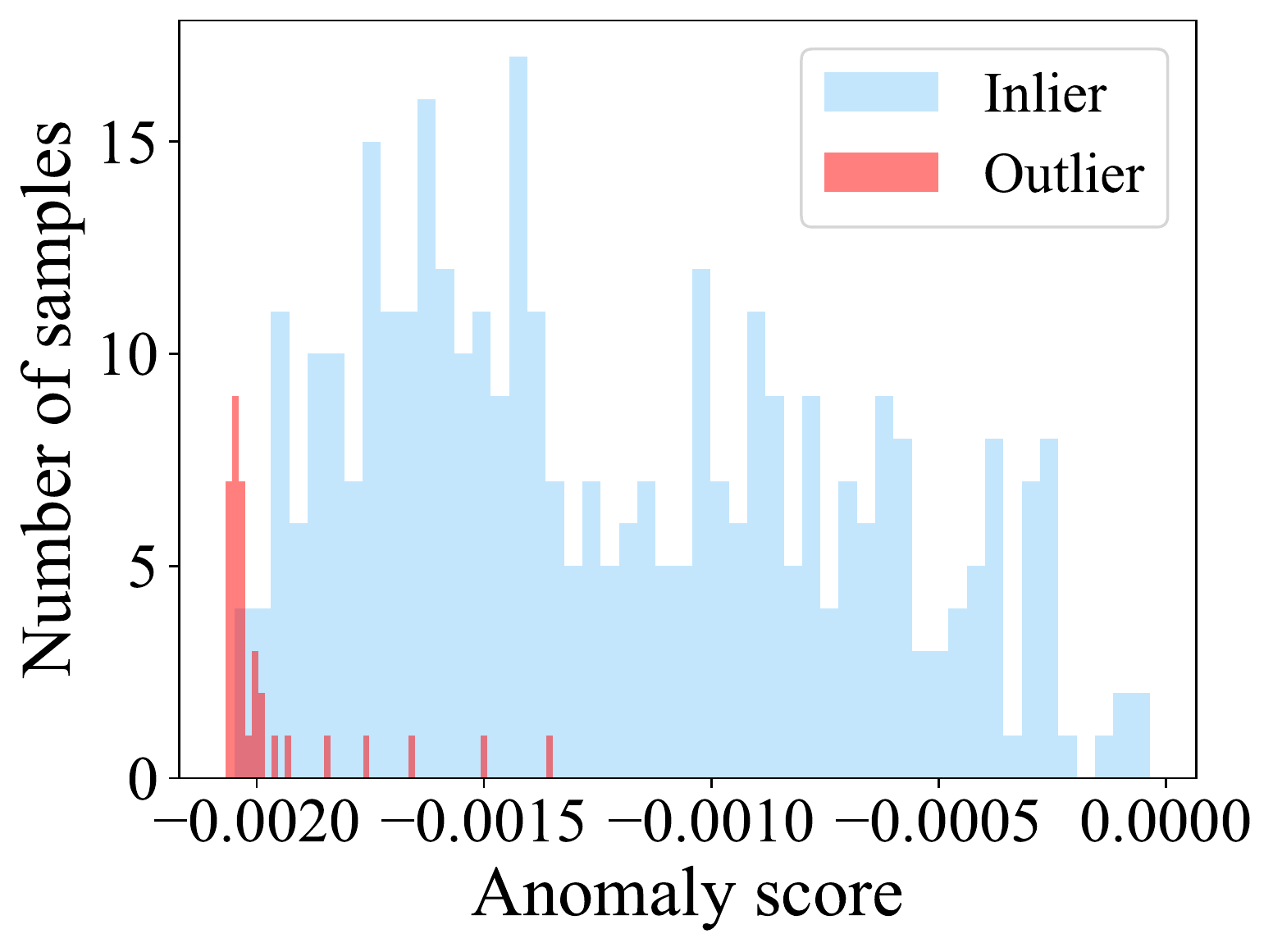}
       \caption{Reuters, inlier 'topics' \protect\\ (w/o perturbation).}
    \end{subfigure}%
    \hfill
    \begin{subfigure}[b]{0.19\linewidth}
       \includegraphics[width=\linewidth]{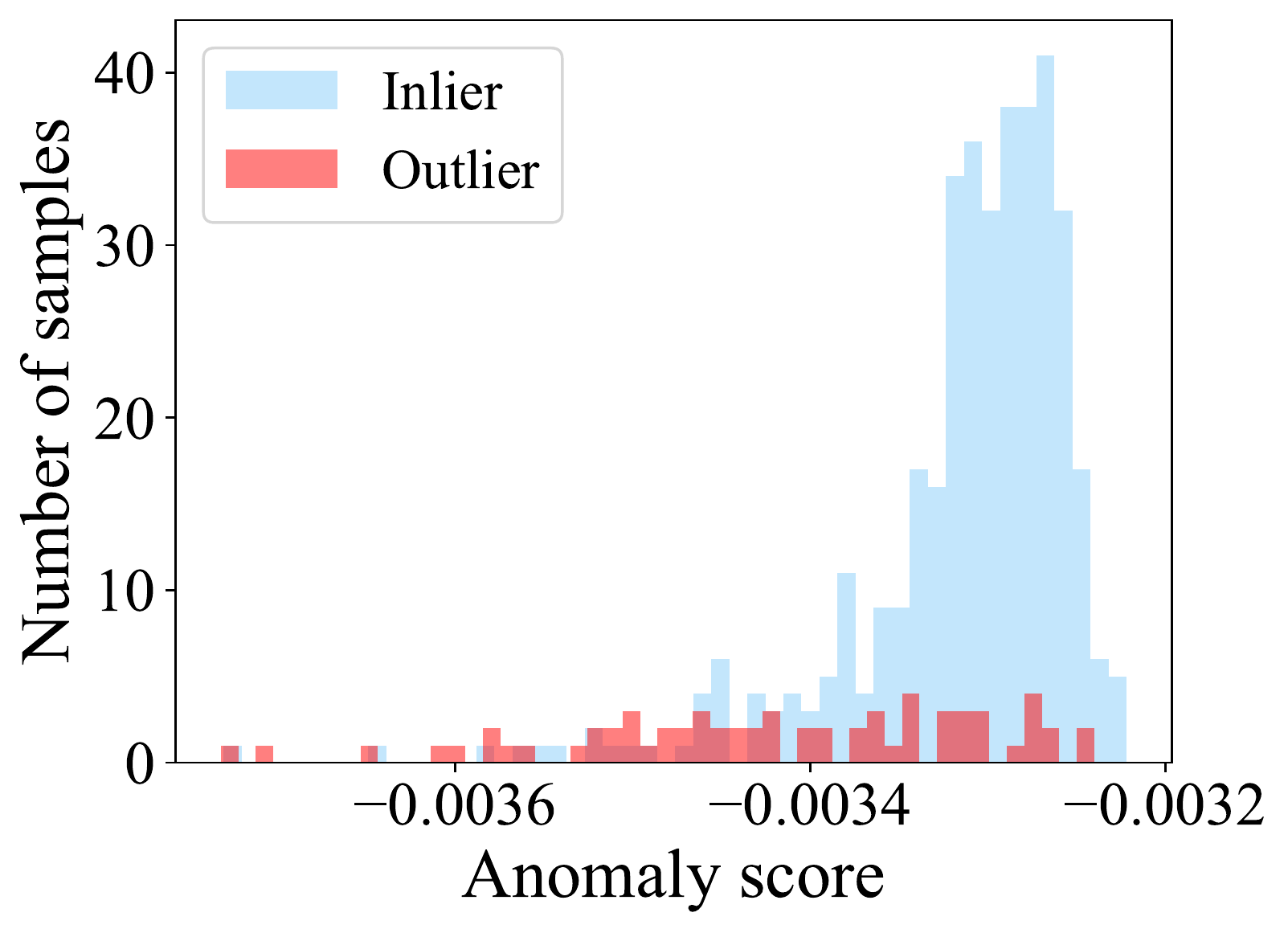}
       \caption{Arrhythmia \protect\\(w/o perturbation).}
    \end{subfigure}%
    \quad
    \begin{subfigure}[b]{0.19\linewidth}
       \includegraphics[width=\linewidth]{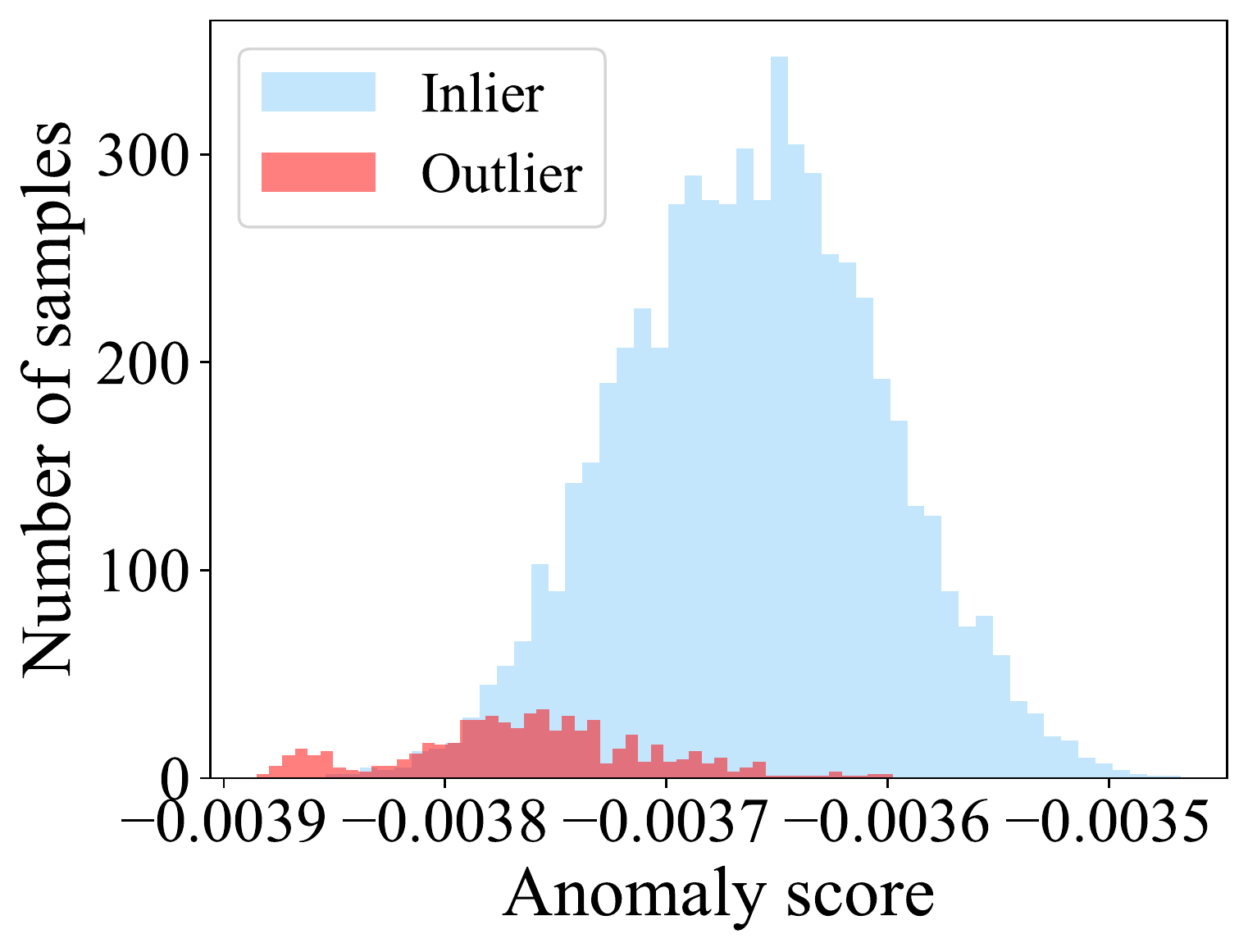}
       \caption{CIFAR-10, inlier 'cat' \protect\\ (w/ perturbation).}
    \end{subfigure}%
    \hfill
    \begin{subfigure}[b]{0.19\linewidth}
       \includegraphics[width=\linewidth]{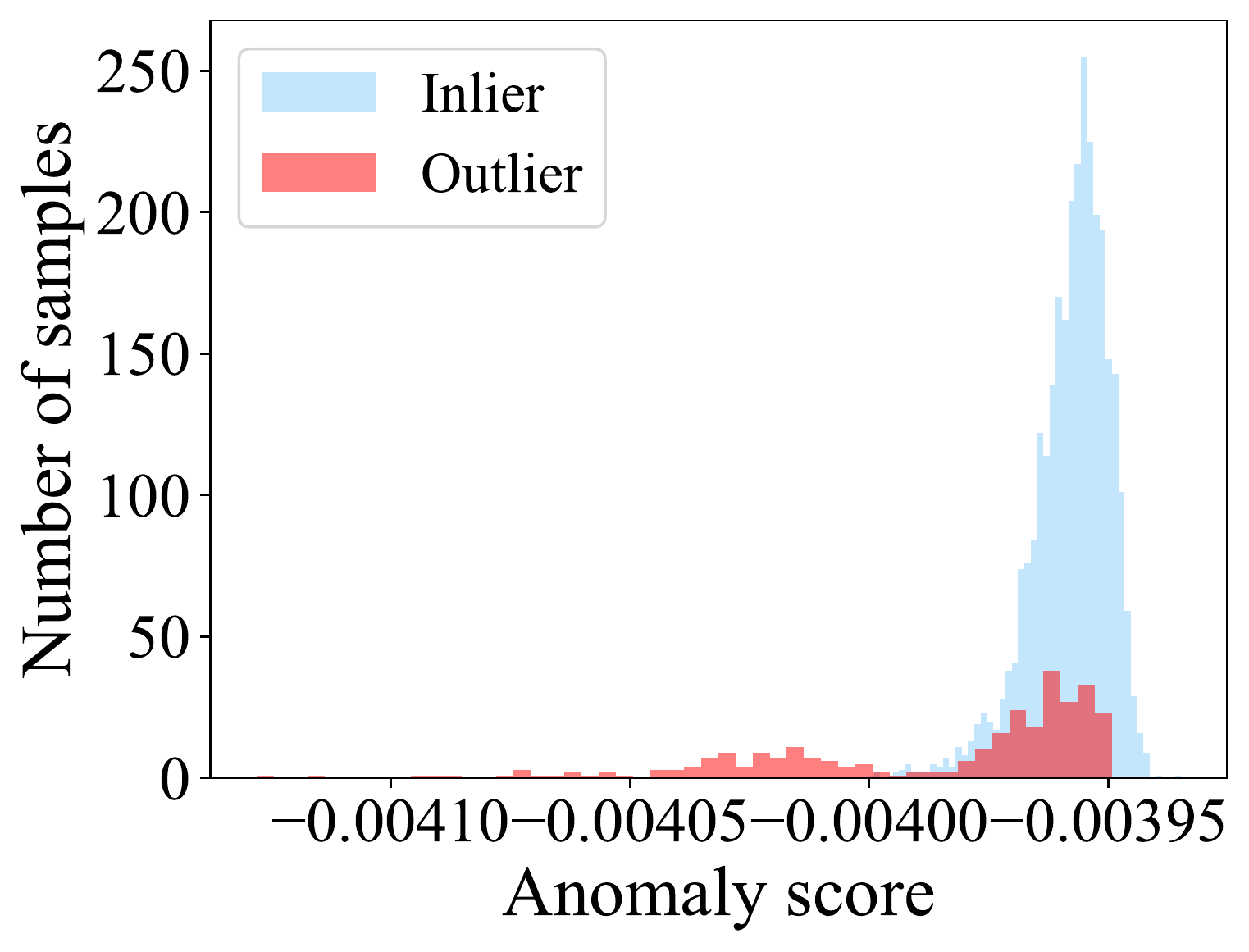}
       \caption{CIFAR-100, inlier 'non-insect invertebrates' (w/ perturbation).}
    \end{subfigure}%
    \hfill
    \begin{subfigure}[b]{0.19\linewidth}
       \includegraphics[width=\linewidth]{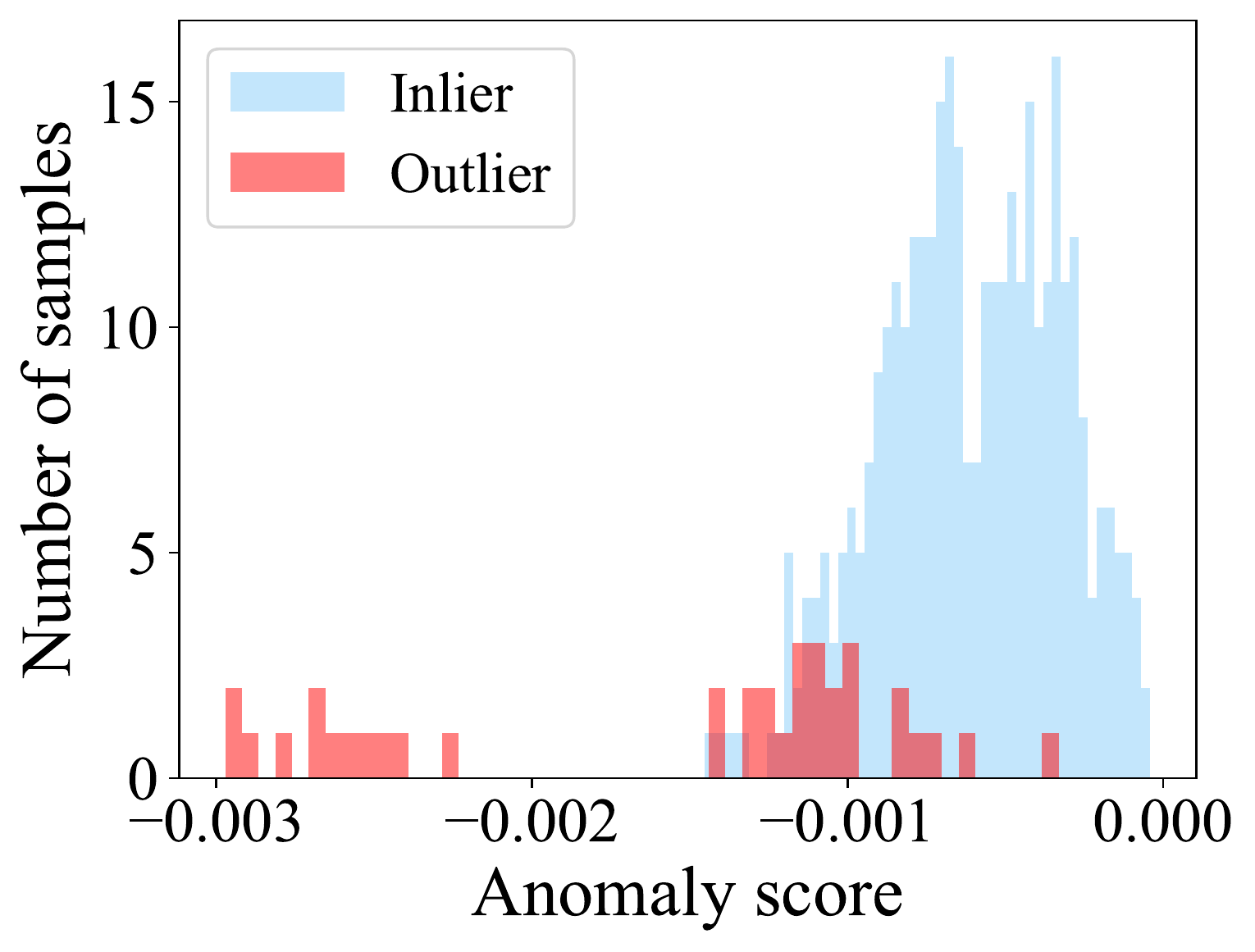}
       \caption{20, 'comp.sys.mac.hardware' \protect\\ (w/ perturbation).}
    \end{subfigure}%
    \hfill
    \begin{subfigure}[b]{0.19\linewidth}
       \includegraphics[width=\linewidth]{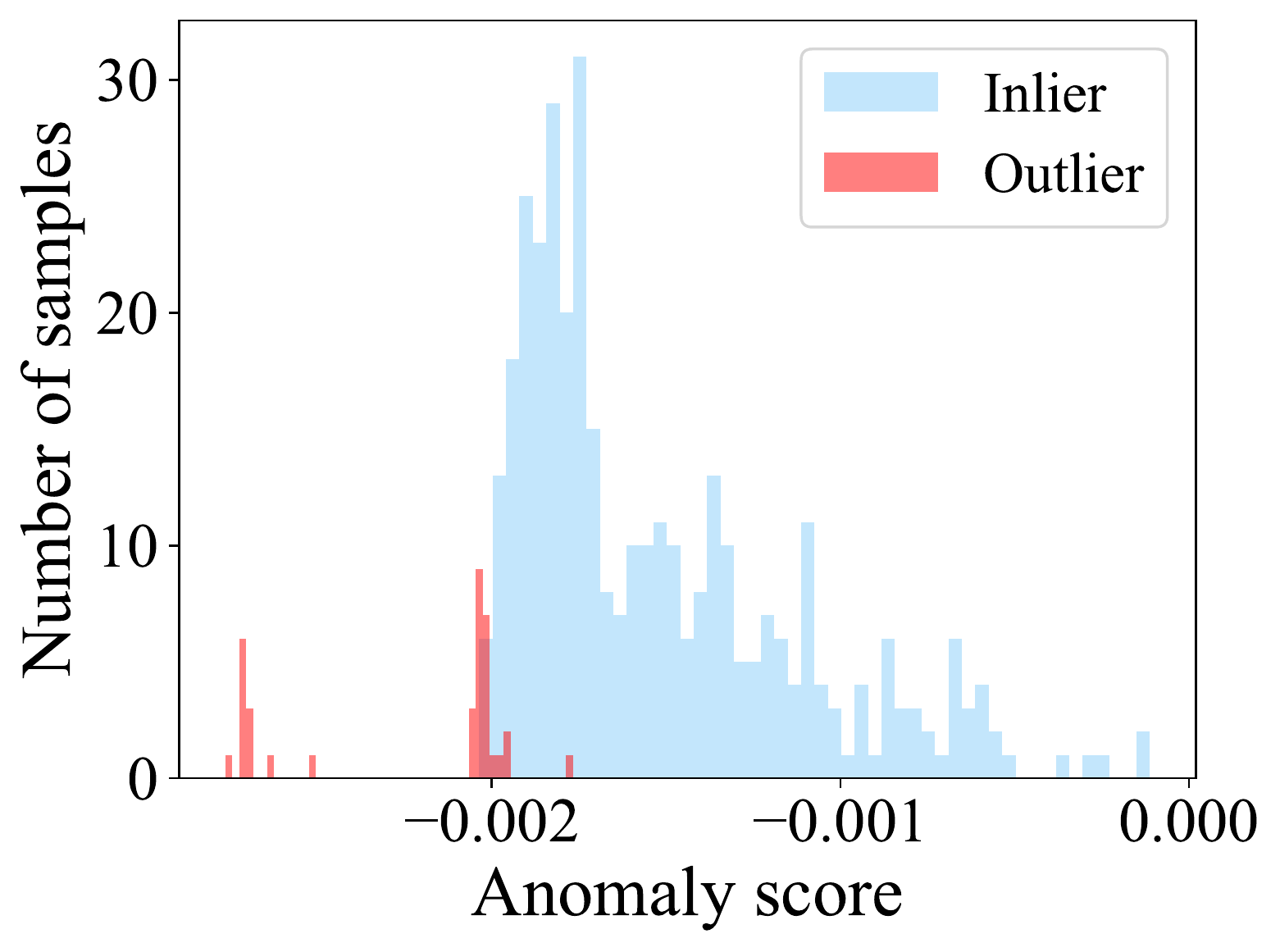}
       \caption{Reuters, inlier 'topics' \protect\\ (w/ perturbation).}
    \end{subfigure}%
    \hfill
    \begin{subfigure}[b]{0.19\linewidth}
       \includegraphics[width=\linewidth]{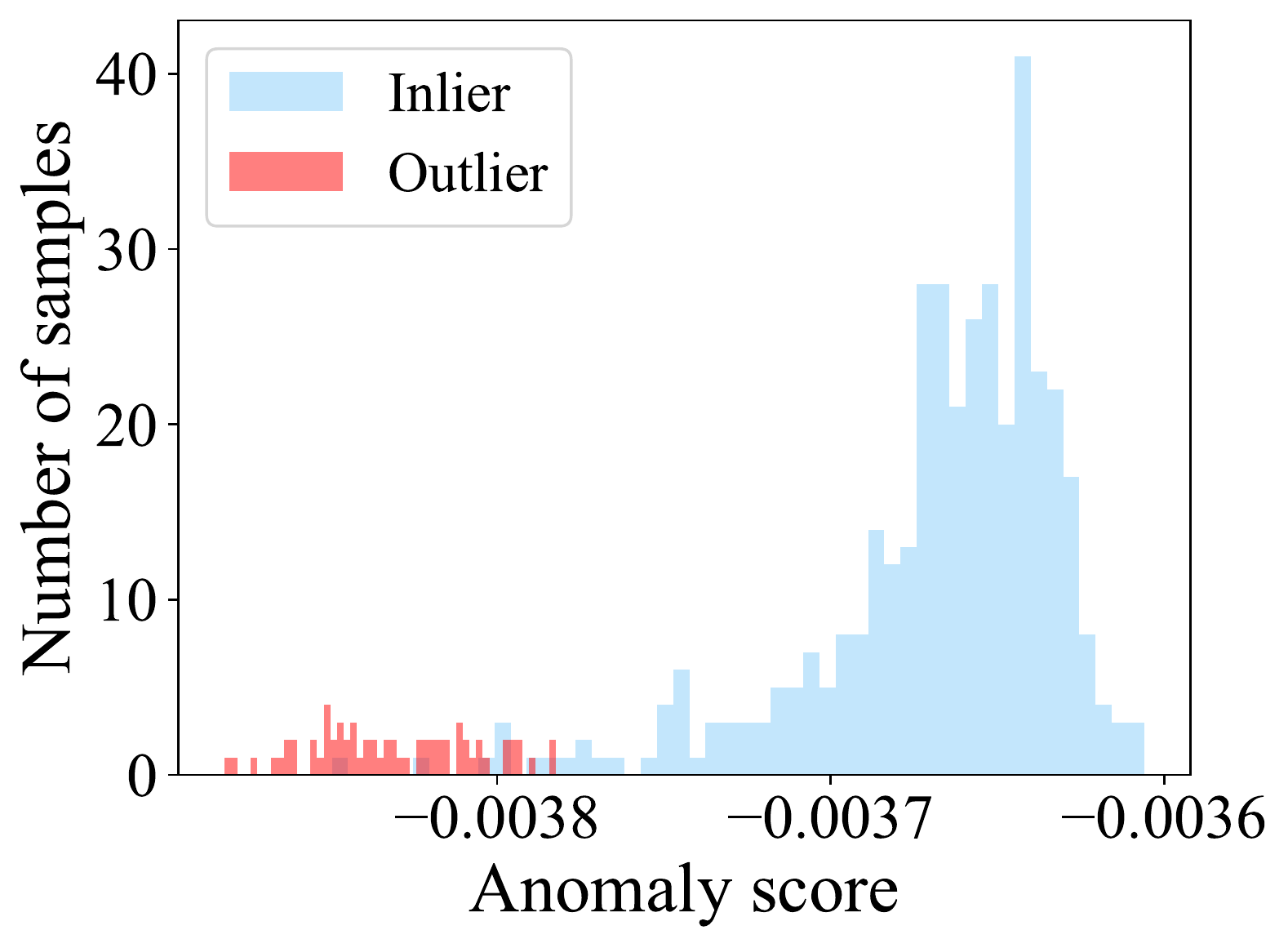}
       \caption{Arrhythmia \protect\\(w/ perturbation).}
    \end{subfigure}%
    \caption{Distribution visualization of the anomaly scores before and after adversarial perturbation of our framework. Best viewed in color.}
    \label{fig: distribution}
 \end{figure*}
 \captionsetup[sub]{font=small}
 
 \captionsetup{width=\linewidth}
 
 \begin{figure}[t]
    \centering
    \begin{subfigure}[b]{0.333\linewidth}
       \includegraphics[width=\linewidth]{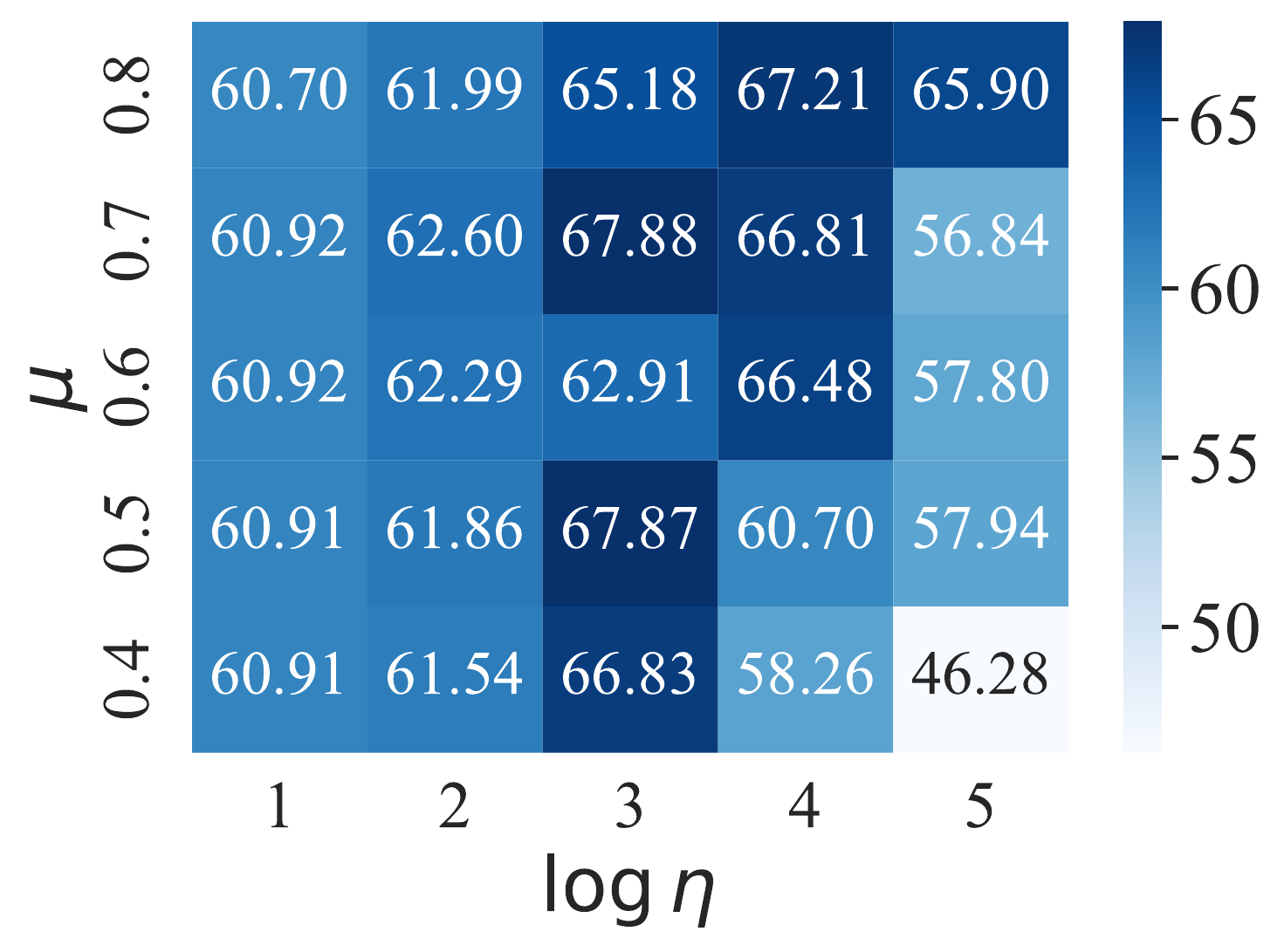}
       \caption{CIFAR-10.}
    \end{subfigure}%
    \hfill
    \begin{subfigure}[b]{0.333\linewidth}
       \includegraphics[width=\linewidth]{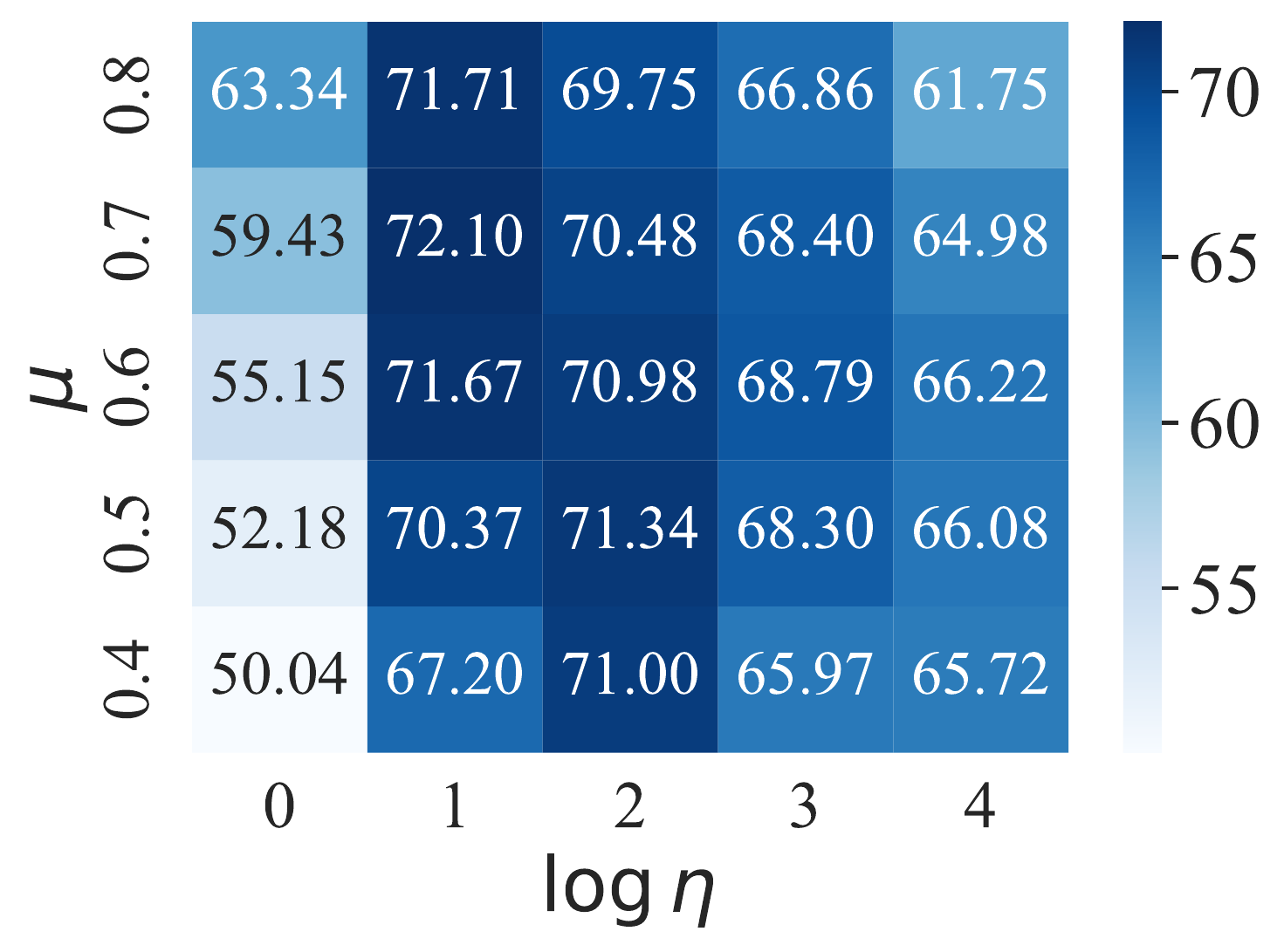}
       \caption{20 Newsgroups.}
    \end{subfigure}%
    \hfill
    \begin{subfigure}[b]{0.333\linewidth}
       \includegraphics[width=\linewidth]{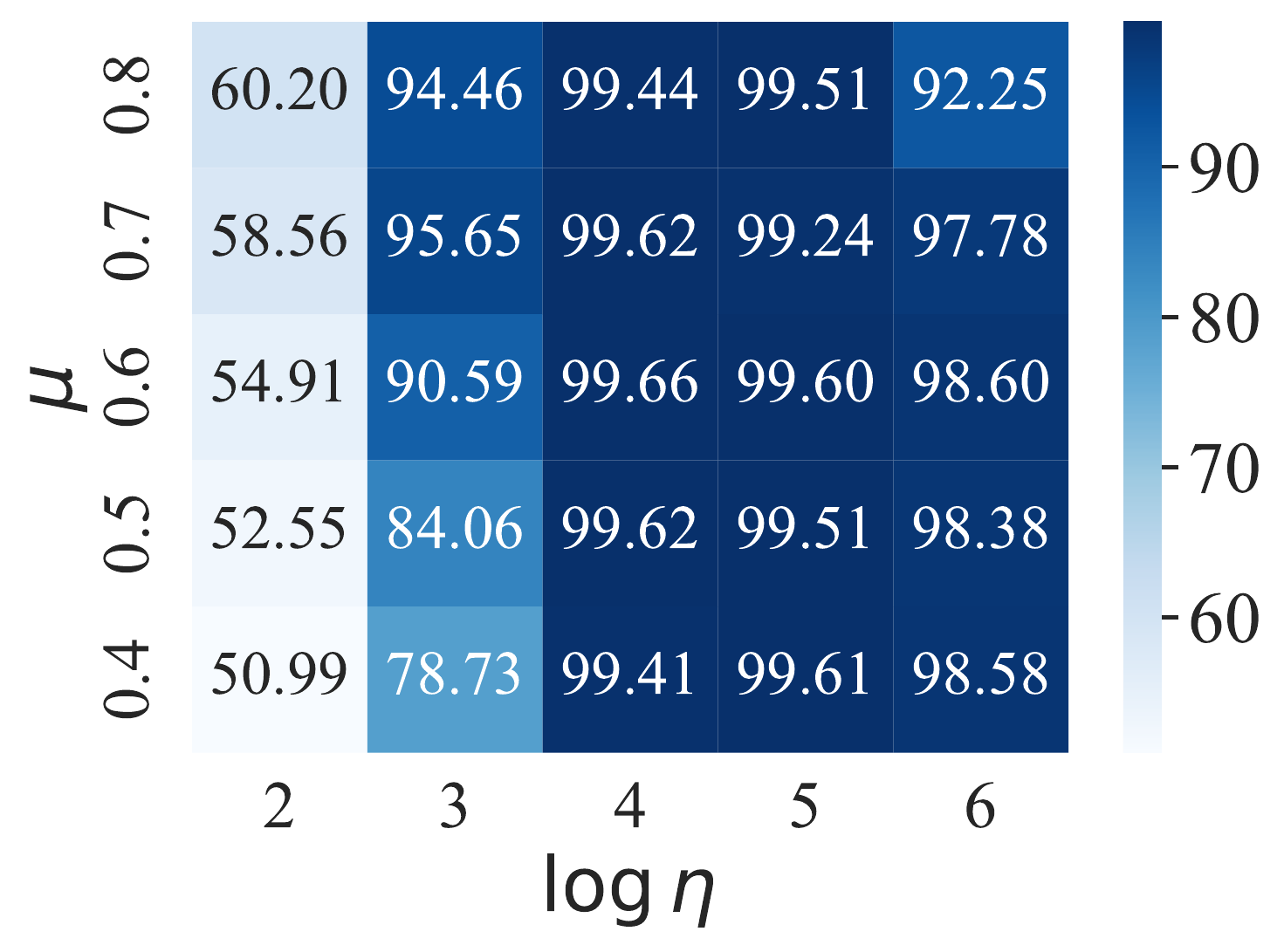}
       \caption{Arrhythmia.}
    \end{subfigure}%
    \caption{AUPR heatmap of our SLA$^2$P. Best viewed in color.}
    \label{fig: heatmap}
 \end{figure}

 \paragraph{Effect of adversarial perturbation.} To further demonstrate the function of the adversarial perturbations on the transformed features, we plot the anomaly score distributions before and after the perturbations. We choose experiments with inlier class 'cat' of CIFAR-10, 'non-insect invertebrates' of CIFR-100, 'comp.sys.mac.hardware' of 20 Newsgroups, 'topics' of Reuters. We also comprise Arrhythmia experiment. As shown in Fig.~\ref{fig: distribution}, despite our self-supervised SLA framework, there still exist outliers whose anomaly scores interlaces with those of inliers. In contrast, after the perturbation, the scores of both normal and anomalous data decreases but those of the inliers are more robust to the perturbations, making the anomalies more separable. 

\begin{figure}[t]
    \centering
    \begin{subfigure}[b]{0.5\linewidth}
       \includegraphics[width=\linewidth]{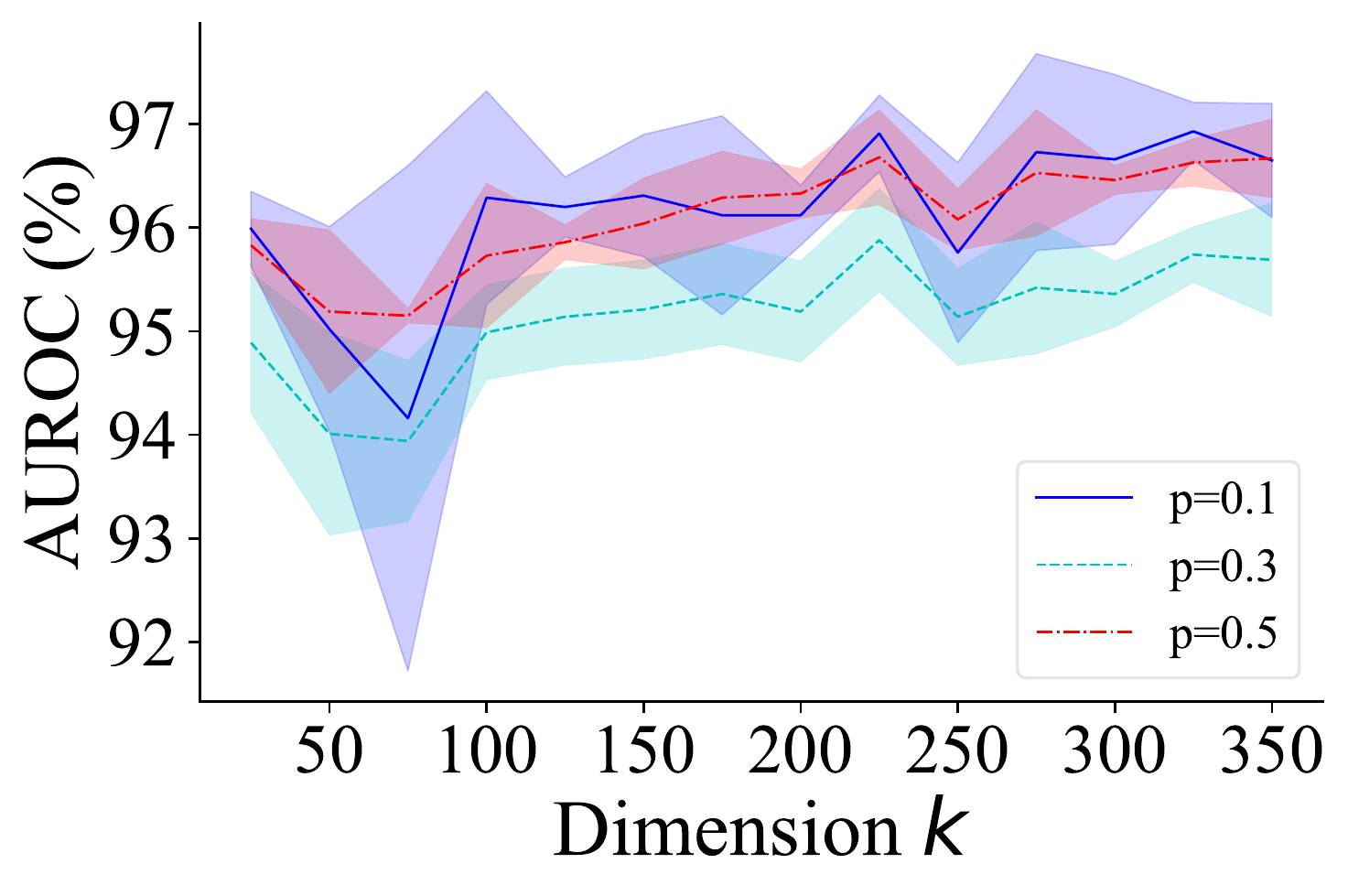}
       \caption{AUROC w.r.t. $k$.}
    \end{subfigure}%
    \hfill
    \begin{subfigure}[b]{0.5\linewidth}
       \includegraphics[width=\linewidth]{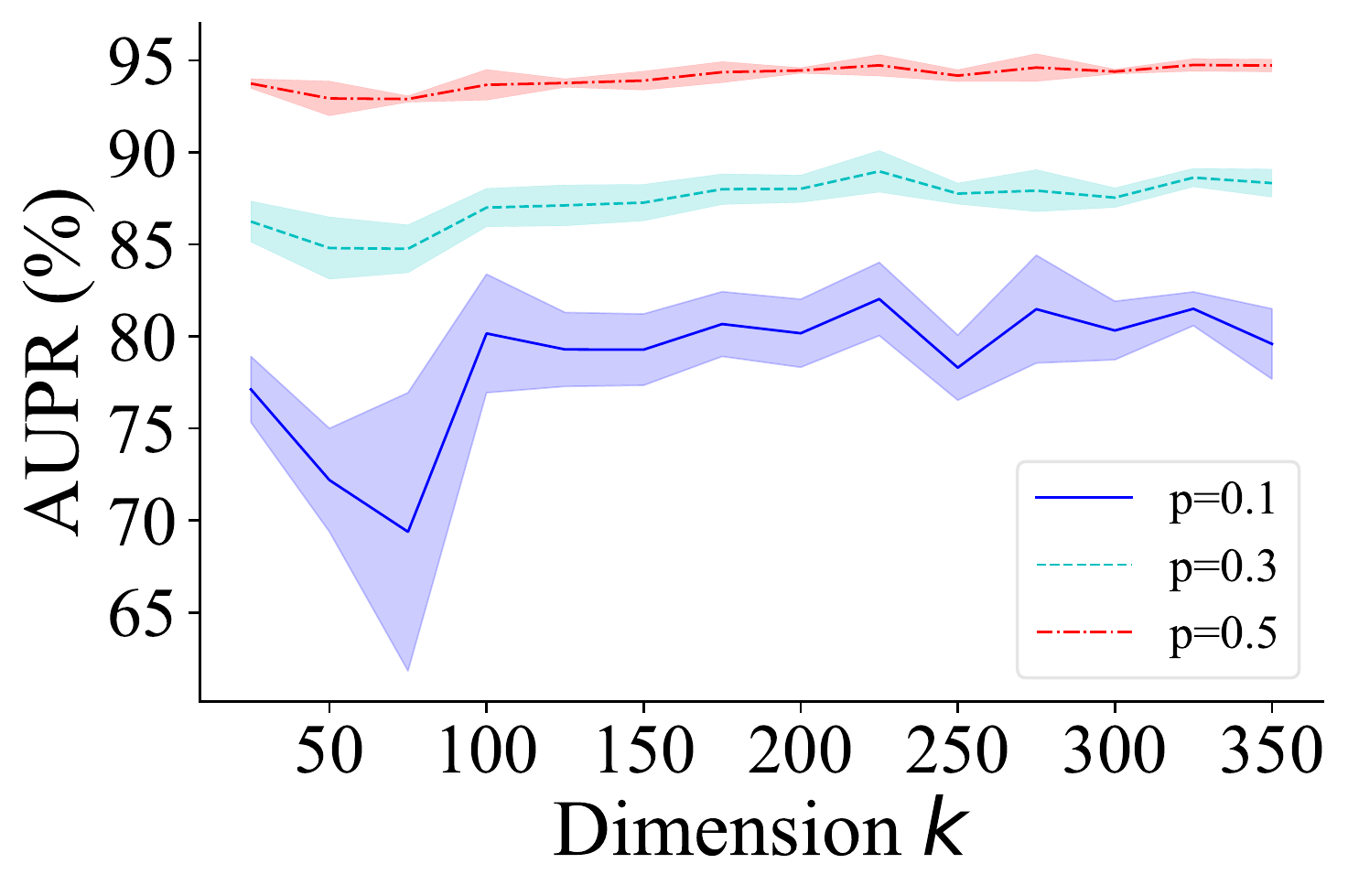}
       \caption{AUPR w.r.t. $k$.}
    \end{subfigure}%
    \caption{UAD performance on Caltech 101 with varying $k$ of our proposed SLA$^2$P method. Best viewed in color.}
    \label{fig: ablation_k}
 \end{figure}

 \paragraph{Relationship between $\mu$ and $\eta$.} We provide fine-grained AUPR results when $p=0.1$ on CIFAR-10, 20 Newsgroups and Arrhythmia tuning $\mu \in \{0.4, 0.5, 0.6, 0.7, 0.8\}$ and $\eta$ in log scale. As illustrated in Fig.~\ref{fig: heatmap}, SLA$^2$P is highly robust w.r.t. the early stopping threshold $\mu$ and the perturbation magnitude $\eta$. There is a wide range of hyperparameter permutations that make the AUPR score of SLA$^2$P over $60\%$ on CIFAR-10, over $55\%$ on 20 Newsgroups and over $50\%$ on Arrhythmia (all better than current SOTA baselines). Also, within the range of the hyperparameter values exhibiting competitive AD performance, higher $\mu$ typically requires larger $\eta$ to guarantee satisfactory detection performance. This makes sense by intuition as the more transformed samples are classified well, the larger the stepsize of the perturbations should be to separate the anomalies out.
 \paragraph{Robustness to transformed dimension $k$.} We investigate the robustness of AD performance of SLA$^2$P to the transformation matrix dimension $k$. We conduct ablation study on Caltech 101 dataset with $k$ varing in the interval $[25,350]$. The experiment is repeated for $5$ times independently and the AUROC and AUPR curve w.r.t. dimension $k$ are depicted in Fig.~\ref{fig: ablation_k}. From the figure we can draw the conclusion that UAD performance of SLA$^2$P is generally robust w.r.t. $k$ with moderate growth when $k$ increases and the growth tends to be flatten out when $k$ becomes comparatively larger. This phenomenon may be elucidated by Theorem~\ref{thm: main} as $k$ needs to be large enough to preserve the similarities between transformed features with high probability.

 \section{Conclusion}
 This paper proposes a novel framework, SLA$^2$P, for unsupervised anomaly detection. We employ random projections as a feature-level self-supervised approach and incorporate adversarial perturbation into it to design discriminative anomaly scores.  We further justify SLA$^2$P's effectiveness both theoretically and empirically.  We hope this work may also shed light to pretext task based perturbation techniques for other unsupervised learning problems.

\newpage

{\small
\bibliographystyle{ims}
\bibliography{sla2p_bib}
}

\newpage
\appendix

\section{Proofs}
    \subsection{Proof of Proposition 1}
       \begin{proof}
          We first prove $\Ab\cV$ is still a linear subspace of $\RR^k$.
 
          Suppose $\yb_1$ and $\yb_2$ are two vectors in $\Ab \cV$, then by definition there exist $\xb_1, \xb_2 \in \cV$ s.t. $\Ab\xb_1=\yb_1, \Ab\xb_2=\yb_2$. Hence 
          \begin{align}\label{eq: plus}
             \yb_1 + \yb_2 &= \Ab\xb_1 + \Ab\xb_2 \nonumber\\
             & = \Ab(\xb_1 +\xb_2).
          \end{align}
          Considering $\xb_1, \xb_2 \in \cV$ and $\cV$ is a linear subspace, we have $\xb_1 +\xb_2 \in \cV$. Combining this with \eqref{eq: plus} we obtain 
          \begin{equation}\label{eq: plus-in}
             \yb_1 + \yb_2 \in \Ab\cV.
          \end{equation}
 
          For any $c\in\RR$, $c\yb_1 = c\Ab\xb_1 = \Ab(c\xb_1)$. As $\xb_1 \in \cV$ and $\cV$ is a linear subspace, we have $c\xb_1 \in \cV$, therefore we get
          \begin{equation}\label{eq: multiply-in}
             c\yb_1 \in \cV.
          \end{equation}
          Combining \eqref{eq: plus-in} and \eqref{eq: multiply-in} we conclude $\Ab\cV$ is a linear subspace of $\RR^k$.
 
          Next we prove $\Ab\cU$ is still a convex set in $\RR^k$.
 
          Suppose $\yb'_1$ and $\yb'_2$ are two vectors in $\Ab \cU$, then by definition there exist $\xb'_1, \xb'_2 \in \cU$ s.t. $\Ab\xb'_1=\yb'_1, \Ab\xb'_2=\yb'_2$. Hence for any $0<\alpha<1$, we have
          \begin{align}\label{eq: convex}
             \alpha \yb'_1+(1-\alpha)\yb'_2 & = \alpha\Ab\xb'_1 + (1-\alpha)\Ab\xb'_2 \nonumber \\
             & = \Ab\big(\alpha\xb'_1 + (1-\alpha)\xb'_2\big).
          \end{align}
          Since $\cU$ is a convex set in $\RR^d$ and $\xb'_1, \xb'_2 \in \cU$, we have $\alpha\xb'_1 + (1-\alpha)\xb'_2 \in \cU$. Then by \eqref{eq: convex} we get $\alpha \yb'_1+(1-\alpha)\yb'_2 \in \Ab\cU$. Thereby we proved the convexity of set $\Ab\cU$.
       \end{proof}
    \subsection{Proof of Theorem 1}
       \begin{proof}
       For notational simplity, we use $\norm{\cdot}$ to denote L2 norm $\norm{\cdot}_2$ and we denote $\Ab^{(m)} = (a_{ij})$ in the following. Some of the proof techniques are adapted from Johnson–Lindenstrauss lemma~\citep{arriaga1999algorithmic}. We first prove property ($1$).
 
       Consider $\forall \vb \in \RR^d$, by our definition we have $\vb^{(m)} = \Ab^{(m)}\vb$, we have
       \begin{align}\label{eq: div-ine}
          & P\Big(\norm{\vb^{(m)}}^2 \ge (1+\epsilon)k\norm{\vb}^2\Big) \nonumber\\
          =& P\Big(\norm{\Ab^{(m)} \vb}^2 \ge (1+\epsilon)k\norm{\vb}^2\Big) \nonumber\\
          =& P\bigg(\frac{\norm{\Ab^{(m)} \vb}^2}{\norm{\vb}^2} \ge (1+\epsilon)k\bigg).
       \end{align}
       Let 
       \begin{align*}
          W \triangleq \frac{\norm{\Ab^{(m)} \vb}^2}{\norm{\vb}^2},
       \end{align*}
       and 
       \begin{align*}
          w_i \triangleq \frac{\sum_j a_{ij} v_j}{\norm{\vb}},
       \end{align*}
       it is straightforward that
       \begin{align*}
          W = \sum_{i=1}^k w_i^2.
       \end{align*}
       Because $a_{ij} \sim \cN(0,1)$ and $\{a_{ij}\}$are i.i.d., we have $w_i \sim \cN(0,1)$ and $\{w_i\}$ are i.i.d. Hence by \eqref{eq: div-ine}, for any $\alpha>0$, we have
       \begin{align}\label{eq: main}
          & P\Big(\norm{\vb^{(m)}}^2 \ge (1+\epsilon)k\norm{\vb}^2\Big) \nonumber\\
          =& P\big(W\ge (1+\epsilon)k\big) \nonumber\\
          =& P\Big(\eexp^{\alpha W} \ge \eexp^{(1+\epsilon)k\alpha}\Big) \nonumber\\
          \mathop{\le}^{\rm (i)} &\frac{\EE\Big(\eexp^{\alpha W}\Big)}{\eexp^{(1+\epsilon)k\alpha}} \nonumber\\
          \mathop{=}^{\rm (ii)} & \frac{\prod_{i=1}^k \EE\big(\eexp^{\alpha w_i^2}\big)}{\eexp^{(1+\epsilon)k\alpha}} \nonumber\\
          = & \Bigg(\frac{\EE\big(\eexp^{\alpha w_1^2}\big)}{\eexp^{(1+\epsilon)\alpha}}\Bigg)^k,
       \end{align}
       where (i) comes from Markov's inequality and (ii) uses the fact that $\{w_i\}$ are i.i.d.. Notice that when $0\le \alpha <\frac{1}{2}$ we have
       \begin{align}\label{eq: calculate-expectation}
          \EE\big(\eexp^{\alpha w_1^2}\big) &= \int_{-\infty}^{\infty} \eexp^{\alpha w_1^2} \frac{1}{\sqrt{2\pi}} \eexp^{-\frac{w_1^2}{2}} \mathrm{d}w_1 \nonumber\\
          & = \int_{-\infty}^{\infty} \frac{1}{\sqrt{2\pi}} \eexp^{-\frac{w_1^2}{2}(1-2\alpha)} \mathrm{d}w_1 \nonumber\\
          & = \frac{1}{\sqrt{1-2\alpha}} \int_{-\infty}^{\infty} \frac{\sqrt{1-2\alpha}}{\sqrt{2\pi}} \eexp^{-\frac{w_1^2}{2}(1-2\alpha)} \mathrm{d}w_1 \nonumber\\
          & \mathop{=}^{\rm (i)} \frac{1}{\sqrt{1-2\alpha}},
       \end{align}
       where (i) uses the integral of random variable with distribution $\cN(0,\frac{1}{1-2\alpha})$. Now plugging \eqref{eq: calculate-expectation} into \eqref{eq: main}, we get
       \begin{align*}
          &P\Big(\norm{\vb^{(m)}}^2 \ge (1+\epsilon)k\norm{\vb}^2\Big) \le \bigg(\frac{\eexp^{-2(1+\epsilon)\alpha}}{1-2\alpha}\bigg)^{\frac{k}{2}}.
       \end{align*}
       Take $\alpha=\frac{\epsilon}{2(1+\epsilon)}$, we have 
       \begin{align}\label{eq: right-bound}
          P\Big(\norm{\vb^{(m)}}^2 \ge (1+\epsilon)k\norm{\vb}^2\Big) 
          \le & \big((1+\epsilon)\eexp^{-\epsilon}\big)^{\frac{k}{2}} \nonumber\\
          = & \big(\eexp^{\log(1+\epsilon) - \epsilon}\big)^{\frac{k}{2}} \nonumber\\
          \mathop{\le}^{\rm (i)} & \big(\eexp^{-(\frac{\epsilon^2}{2} - \frac{\epsilon^3}{3})}\big)^{\frac{k}{2}} \nonumber\\
          \le & \eexp^{-\frac{k}{4}(\epsilon^2 - \epsilon^3)},
       \end{align}
       where (i) employs inequality $\log(1+x)<x-\frac{x^2}{2} + \frac{x^3}{3}$.
 
       Similarly, we have
       \begin{align*}
          &P\Big(\norm{\vb^{(m)}}^2 \le (1-\epsilon)k\norm{\vb}^2\Big) \\
          = &P\big(W\le(1-\epsilon)k\big) \\
          \le & P\Big(\eexp^{-\alpha W} \ge \eexp^{-(1-\epsilon)\alpha k}\Big) \\
          \le &\frac{\EE \Big(\eexp^{-\alpha W}\Big)}{\eexp^{-(1-\epsilon)\alpha W}} \\
          = & \frac{\prod_{i=1}^k \EE \big(\eexp^{-\alpha w_i^2}\big)}{\eexp^{-(1-\epsilon)\alpha k}} \\
          = & \bigg(\frac{\EE\big(\eexp^{-\alpha w_1^2}\big)}{\eexp^{-(1-\epsilon)\alpha}}\bigg)^k \\
          = & \bigg(\frac{\eexp^{2(1-\epsilon)\alpha}}{1+2\alpha}\bigg)^{\frac{k}{2}}.
       \end{align*}
       Take $\alpha = \frac{\epsilon}{2(1-\epsilon)}$, we obtain
       \begin{align}\label{eq: left-bound}
          P\Big(\norm{\vb^{(m)}}^2 \le (1-\epsilon)k\norm{\vb}^2\Big) &\le \big((1-\epsilon)\eexp^{\epsilon}\big)^\frac k2 \nonumber\\
          & = \big(\eexp^{\log (1-\epsilon) + \epsilon}\big)^{\frac{k}{2}} \nonumber\\
          & \le \big( \eexp^{-\frac{\epsilon^2}{2} + \frac{\epsilon^3}{3}} \big)^{\frac{k}{2}} \nonumber\\
          & \le \eexp^{-\frac{k}{4}(\epsilon^2 - \epsilon^3)}.
       \end{align}
       Combining \eqref{eq: right-bound} and \eqref{eq: left-bound}  and setting $\eexp^{-\frac{k}{4}(\epsilon^2 - \epsilon^3)}<\frac{\delta}{2}$, we have: when $k>\frac{4\log\frac{2}{\delta}}{\epsilon^2 - \epsilon^3}$, with probability at least $1-\delta$,
       \begin{equation}\label{eq: main}
          (1-\epsilon)\norm{\vb}^2 \le \frac{\norm{\vb^{(m)}}^2}{k} \le (1+\epsilon)\norm{\vb}^2.
       \end{equation}
       Now for any vectors $\vb_i, \vb_j \in \RR^d$, substituting $\vb$ in \eqref{eq: main} by $\vb_i - \vb_j$, we hereby proved property (1).
 
       Next we prove the property {\bf (2)}.
 
       For any index pair $(i,j)$, applying \eqref{eq: main} to vectors $\vb_1+\vb_2$ and $\vb_1-\vb_2$, we have when $k>\frac{4\log \frac{4}{\delta}}{\epsilon^2 - \epsilon^3}$, with probability at least $1-\delta$, 
       \begin{align}
          k(1-\epsilon)\norm{\vb_i + \vb_j}^2 &\le \norm{\vb^{(m)}_i + \vb^{(m)}_j}^2 \le k(1+\epsilon)\norm{\vb_i + \vb_j}^2, \label{eq: lege-1}\\ 
          k(1-\epsilon)\norm{\vb_i - \vb_j}^2 &\le \norm{\vb^{(m)}_i - \vb^{(m)}_j}^2 \le k(1+\epsilon)\norm{\vb_i - \vb_j}^2. \label{eq: lege-2}
       \end{align}
       Hence we have
       \begin{align}\label{eq: last-1}
          \vb_i^{(m)}\cdot\vb_j^{(m)}&=\frac{\norm{\vb_i^{(m)} + \vb_j^{(m)}}^2 - \norm{\vb_i^{(m)} - \vb_j^{(m)}}^2}{4} \nonumber\\
          & \mathop{\ge}^{\rm (i)} \frac{k(1-\epsilon)\norm{\vb_i+\vb_j}^2 - k(1+\epsilon)\norm{\vb_i - \vb_j}^2}{4} \nonumber\\
          & \mathop{=}^{\rm (ii)} \frac{k}{4}\Big(4\vb_i\cdot\vb_j - 2\epsilon\big(\norm{\vb_i}^2+\norm{\vb_j}^2\big)\Big) \nonumber\\
          & = k\Big(\vb_i\cdot\vb_j -\epsilon\Big),
       \end{align}
       where (i) uses the left hand side of \eqref{eq: lege-1} and right hand side of \eqref{eq: lege-2}, and (ii) utilizes $\norm{\vb_i} = \norm{\vb_j}=1$ due to L2 normalization. Likewise, we get 
       \begin{align}\label{eq: last-2}
          \vb_i^{(m)}\cdot\vb_j^{(m)}&=\frac{\norm{\vb_i^{(m)} + \vb_j^{(m)}}^2 - \norm{\vb_i^{(m)} - \vb_j^{(m)}}^2}{4} \nonumber\\
          & \le \frac{k(1+\epsilon)\norm{\vb_i+\vb_j}^2 - k(1-\epsilon)\norm{\vb_i - \vb_j}^2}{4} \nonumber\\
          & = \frac{k}{4}\Big(4\vb_i\cdot\vb_j + 2\epsilon\big(\norm{\vb_i}^2+\norm{\vb_j}^2\big)\Big) \nonumber\\
          & = k\Big(\vb_i\cdot\vb_j +\epsilon\Big),
       \end{align}
       Combining \eqref{eq: last-1} and \eqref{eq: last-2}, we get as long as  $k>\frac{4\log \frac{4}{\delta}}{\epsilon^2 - \epsilon^3}$, with probability at least $1-\delta$,
       \begin{align*}
          \vb_i\cdot\vb_j - \epsilon \le \frac{\vb_i^{(m)}\cdot\vb_j^{(m)}}{k} \le \vb_i\cdot\vb_j + \epsilon.
       \end{align*}
    \end{proof}

\section{Complete Results}
In this section we provide the UAD performance comparison on image tasks using ResNet-101 as feature extractor. As illustrated above in Fig.~\ref{table: image-res101}, our methods still achieve the best performance consistently.

We also provide the detailed UAD performance results on CIFAR-10 and text tabular datasets. Note that our reported results in the main paper is the averaged AUROC and AUPR, as we have mentioned that in each dataset each class serves as normal data in turn. The 20 superclasses of CIFAR-100 are 'fish', 'flowers', 'food containers', 'fruit and vegetables', 'household electrical devices', 'household furniture', 'insects', 'large carnivores', 'large man-made outdoor things', 'large natural outdoor scenes', 'large omnivores and herbivores', 'medium-sized mammals', 'non-insect invertebrates', 'people', 'reptiles', 'small mammals', 'trees', 'vehicles' and 'vehicles 2'. The 11 selected classes of Caltech are 'watch', 'ketch', 'hawksbill', 'grand piano', 'chandelier', 'car side', 'bonsai', 'airplanes', 'motorbikes', 'leopards' and 'faces'. All the experients are run 5 times independently using seeds $0,1,2,3,4$, and averaged scores$\pm$standard deviations are reported.

\setlength{\tabcolsep}{3pt}
\captionsetup{width=0.95\textwidth}
\begin{table*}[t]
   \centering
   \small
   \scalebox{0.85}{
   \begin{tabular}{cccccccccc}
   \toprule
   Dataset   & p  & IF  &  OCSVM & DAGMM  & RSRAE  & PANDA &  SLA (ours) & SLA$^2$P (ours) \\ \midrule 
               & 0.1& 85.44~/~45.60 & 87.30~/~50.23 & 64.27~/~19.60 & 84.33~/~45.95 & 88.68~/~60.11 &\underline{92.74}~/~\underline{65.59}& {\bf 94.08}~/~{\bf 73.62}  \\
   CIFAR-10   & 0.3& 81.45~/~58.85 & 81.29~/~58.14 & 68.34~/~38.61 & 75.72~/~52.42 & 87.36~/~65.35 &\underline{90.08}~/~\underline{74.93}& {\bf 91.66}~/~{\bf 78.28}   \\
               & 0.5& 78.65~/~64.72 & 76.32~/~60.83 & 68.68~/~50.61 & 72.06~/~58.15 & 83.06~/~71.27 &\underline{87.11}~/~\underline{76.56}& {\bf 89.18}~/~{\bf 80.69}  \\ \hline
               & 0.1& 79.12~/~34.93 & 81.69~/~37.44 & 53.52~/~11.95 & 91.40~/~60.68 & 87.90~/~\underline{58.41} &\underline{89.94}~/~58.38& {\bf 93.04}~/~{\bf 75.05}   \\
   CIFAR-100  & 0.3& 75.91~/~51.41 & 76.23~/~50.37 & 53.99~/~27.21 & 87.35~/~69.41 & 86.65~/~64.35 &\underline{87.63}~/~\underline{71.31}& {\bf 87.92}~/~{\bf 73.40}  \\
               & 0.5& 73.29~/~58.90 & 72.35~/~56.26 & 55.36~/~39.21 & 84.12~/~72.61 & 81.81~/~69.62 &{\bf 85.36}~/~{\bf 75.37}& \underline{84.61}~/~\underline{74.41}   \\ \hline
               & 0.1& 91.11~/~62.71 & 93.27~/~70.29 & 72.26~/~33.32 & 95.24~/~78.83 & 96.40~/~81.11 &\underline{96.69}~/~\underline{82.87}& {\bf 96.98}~/~{\bf 84.73}   \\
   Caltech 101& 0.3& 87.41~/~70.92 & 88.60~/~74.02 & 74.01~/~51.49 & 89.47~/~73.90 & 95.16~/~88.38 &\underline{96.23}~/~\underline{89.44}& {\bf 96.48}~/~{\bf 90.94}   \\
               & 0.5& 84.68~/~73.63 & 83.50~/~71.99 & 70.09~/~56.17 & 84.95~/~75.30 & 94.27~/~90.34 &\underline{94.89}~/~\underline{90.40}& {\bf 97.25}~/~{\bf 95.76}  \\
   \bottomrule
   \end{tabular}
   }
   \caption{AUROC/AUPR (\%) results on image datasets for UAD with ResNet-101 as the feature extractor. The best in bold and the second best underlined. The mean scores over $5$ independent runs are reported.} 
   \label{table: image-res101}
   \vspace{-1mm}
\end{table*}

\begin{table*}[htbp]
   \centering
   \scalebox{0.8}{
   \begin{tabular}{ccccccccc}
   \toprule
     Inlier class name & IF & OCSVM & DAGMM& E$^3$Outlier & RSRAE& PANDA  & SLA (ours) & SLA$^2$P (ours) \\ \hline
    airplane & 84.82$\pm$0.76 & 85.21$\pm$0.62  & 68.51$\pm$0.01 & 78.61$\pm$0.69 & 83.95$\pm$0.61 & 87.48$\pm$0.59 & 88.84$\pm$0.60& {\bf 90.50$\pm$0.58}\\
    automobile & 92.33$\pm$0.56 & 94.68$\pm$0.40  & 66.64$\pm$0.02 & 95.47$\pm$0.65 & 93.12$\pm$0.59 & 96.34$\pm$0.31 & 96.77$\pm$0.28& {\bf 97.21$\pm$0.30}\\
    bird & 74.39$\pm$0.88 & 72.87$\pm$1.01  & 46.31$\pm$0.01 & 79.08$\pm$0.61 & 72.90$\pm$1.63 & 77.77$\pm$0.97& 81.30$\pm$1.19& {\bf 83.83$\pm$1.33}\\
    cat & 75.37$\pm$2.23 & 80.87$\pm$0.45  & 74.71$\pm$0.00 & 73.92$\pm$1.23 & 67.11$\pm$3.12 & 83.58$\pm$0.46 & 86.49$\pm$0.92& {\bf 88.59$\pm$0.81}\\
    deer & 87.01$\pm$1.24 & 87.88$\pm$0.26  & 76.41$\pm$0.04 & 84.91$\pm$1.01 & 84.94$\pm$1.57 & 90.87$\pm$0.34 & 90.65$\pm$0.38& {\bf 92.13$\pm$0.44} \\
    dog & 71.14$\pm$4.30 & 74.42$\pm$0.41  & 47.20$\pm$0.01 & 86.76$\pm$0.97 & 80.01$\pm$0.88 & 75.38$\pm$0.96 & 87.99$\pm$0.85& {\bf 89.86$\pm$1.04}\\
    frog & 87.02$\pm$0.60 & 86.23$\pm$0.79  & 76.09$\pm$0.02 & 87.21$\pm$0.63 & 81.87$\pm$2.14 & 86.59$\pm$0.58 & 91.68$\pm$0.37& {\bf 93.05$\pm$0.39} \\
    horse & 79.89$\pm$2.61 & 83.68$\pm$0.81  & 47.81$\pm$0.03 & 92.19$\pm$0.93 & 88.92$\pm$1.43 & 85.16$\pm$0.29 & 93.32$\pm$0.26& {\bf 94.16$\pm$0.29} \\
    ship & 89.47$\pm$1.31 & 91.23$\pm$0.20  & 78.24$\pm$5.61 & 92.62$\pm$0.51 & 89.91$\pm$1.71 & 90.06$\pm$0.47 & 94.64$\pm$0.50& {\bf 95.76$\pm$0.48} \\
    truck & 90.61$\pm$1.40 & 92.90$\pm$0.49  & 62.33$\pm$3.55 & 90.90$\pm$1.04 & 93.10$\pm$0.35 & 93.60$\pm$0.32 & 96.44$\pm$0.48& {\bf 96.98$\pm$0.31} \\ 
    \midrule
    {\it average} & 83.24$\pm$0.91 & 85.19$\pm$0.13  & 59.86$\pm$0.03 & 85.89$\pm$0.08 & 83.58$\pm$0.21 & 86.69$\pm$0.57 & 90.81$\pm$0.15 & {\bf 91.56$\pm$0.12} \\ \bottomrule
   \end{tabular}%
   }
   \caption{AUROC (\%) on CIFAR-10 for UAD using ResNet-50 as extractor when $p=0.1$. The best performance is in bold.}
\end{table*}

\begin{table*}[htbp]
   \centering
   \scalebox{0.8}{
   \begin{tabular}{ccccccccc}
   \toprule
    Inlier class name & IF & OCSVM & DAGMM& E$^3$Outlier & RSRAE& PANDA & SLA (ours) & SLA$^2$P (ours)  \\ \hline
    airplane & 81.51$\pm$1.50 & 80.18$\pm$0.38  & 76.62$\pm$0.03 & 76.22$\pm$0.71 & 77.04$\pm$0.75 & 83.39$\pm$0.88 & 85.45$\pm$0.27& {\bf 86.96$\pm$0.41} \\
    automobile & 90.01$\pm$1.58 & 91.28$\pm$0.12  & 87.76$\pm$0.01 & 94.32$\pm$0.52 & 88.78$\pm$0.43 & 94.63$\pm$0.54& 95.62$\pm$0.30& {\bf 96.21$\pm$0.26}\\
    bird & 67.18$\pm$1.28 & 67.23$\pm$0.53  & 67.29$\pm$0.01 & 71.18$\pm$2.39 & 62.03$\pm$1.48 & 73.37$\pm$2.11 & 76.36$\pm$0.53& {\bf 77.66$\pm$0.38} \\
    cat & 70.71$\pm$1.10 & 75.08$\pm$0.60  & 77.23$\pm$0.02 & 62.54$\pm$1.97 & 52.22$\pm$1.19 & 81.97$\pm$0.74 & 82.00$\pm$0.99& {\bf 83.38$\pm$0.88} \\
    deer & 84.11$\pm$0.18 & 83.12$\pm$0.29  & 82.01$\pm$0.04 & 80.76$\pm$1.36 & 76.23$\pm$1.86 & 88.46$\pm$0.62 & 88.16$\pm$0.47& {\bf 89.73$\pm$0.58} \\
    dog & 64.42$\pm$3.11 & 67.10$\pm$0.38  & 52.12$\pm$0.01 & 76.67$\pm$1.05 & 71.12$\pm$0.83 & 69.70$\pm$1.78 & 82.58$\pm$1.13& {\bf 85.42$\pm$0.57}\\
    frog & 83.13$\pm$1.53 & 78.79$\pm$0.51  & 75.78$\pm$0.02 & 80.33$\pm$2.00 & 69.91$\pm$2.19 & 82.36$\pm$0.82 & 87.46$\pm$0.54& {\bf 89.75$\pm$0.52} \\
    horse & 73.82$\pm$2.10 & 77.11$\pm$0.31  & 52.79$\pm$0.01 & 87.31$\pm$0.87 & 80.31$\pm$0.81 & 81.97$\pm$0.51 & 91.46$\pm$0.17& {\bf 92.38$\pm$0.29}\\
    ship & 85.73$\pm$1.76 & 85.85$\pm$0.38  & 72.72$\pm$18.01 & 90.52$\pm$1.21 & 84.04$\pm$1.19 & 88.63$\pm$0.73 & 92.51$\pm$0.17& {\bf 93.85$\pm$0.33} \\
    truck & 89.08$\pm$0.68 & 88.69$\pm$0.18  & 84.87$\pm$4.69 & 90.13$\pm$0.78 & 89.40$\pm$0.64 & 91.77$\pm$0.67 & 95.17$\pm$0.33 & {\bf 95.78$\pm$0.29}\\ 
    \midrule 
    {\it average} & 79.36$\pm$0.81 & 79.52$\pm$0.10  & 67.04$\pm$2.19 & 80.58$\pm$0.35 & 75.05$\pm$0.32 & 83.63$\pm$0.94 &87.68$\pm$0.25 &{\bf 87.69$\pm$0.18} \\ \bottomrule
   \end{tabular}%
   }
   \caption{AUROC (\%) on CIFAR-10 for UAD using ResNet-50 as extractor when $p=0.3$. The best performance is in bold.}
\end{table*}

\begin{table*}[htbp]
   \centering
   \scalebox{0.8}{
   \begin{tabular}{cccccccccc}
   \toprule
    Inlier class name & IF & OCSVM & DAGMM& E$^3$Outlier & RSRAE& PANDA& SLA (ours) & SLA$^2$P (ours) \\ \hline
    airplane & 79.51$\pm$1.78 & 75.67$\pm$0.42  & 79.02$\pm$3.22 & 73.79$\pm$1.01 & 73.68$\pm$0.51 & 83.61$\pm$1.15 & 81.76$\pm$0.43& {\bf 84.17$\pm$1.14}  \\
     automobile & 87.57$\pm$1.39 & 86.88$\pm$0.28  & 86.11$\pm$0.02 & 93.32$\pm$0.54 & 86.02$\pm$0.87 & 92.92$\pm$0.77 & 94.38$\pm$0.44& {\bf 95.31$\pm$0.33} \\
     bird & 65.67$\pm$2.08 & 63.29$\pm$0.43  & 72.02$\pm$2.41 & 65.53$\pm$2.19 & 57.62$\pm$0.87 & 72.25$\pm$2.37 & 72.76$\pm$0.69& {\bf 73.20$\pm$0.79}\\
     cat & 67.61$\pm$2.60 & 71.20$\pm$0.43  & 78.35$\pm$2.03 & 55.78$\pm$1.70 & 48.51$\pm$1.41 & 79.67$\pm$1.24 & 77.74$\pm$0.40& {\bf 79.95$\pm$0.18} \\
    deer & 82.13$\pm$1.56 & 79.14$\pm$0.15  & 77.44$\pm$2.31 & 76.01$\pm$1.41 & 72.58$\pm$1.28 & {\bf 88.24$\pm$0.96} & 85.74$\pm$0.47&  88.04$\pm$0.33\\
     dog & 59.58$\pm$3.01 & 62.38$\pm$0.18  & 59.76$\pm$0.28 & 72.12$\pm$3.14 & 63.19$\pm$0.76 & 70.48$\pm$2.12 & 77.76$\pm$0.85& {\bf 81.12$\pm$1.14}\\
    frog & 80.52$\pm$1.11 & 73.02$\pm$0.45  & 85.67$\pm$7.02 & 74.59$\pm$0.73 & 64.81$\pm$1.00 & 79.89$\pm$1.18 & 81.83$\pm$0.64& {\bf 85.26$\pm$1.69}\\
     horse & 71.08$\pm$1.37 & 72.17$\pm$0.16  & 53.93$\pm$11.51 & 85.38$\pm$1.04 & 78.03$\pm$1.12 & 80.92$\pm$0.95 & 90.16$\pm$0.26&{\bf 91.50$\pm$0.40}  \\
    ship & 83.20$\pm$1.91 & 80.62$\pm$0.33  & 84.14$\pm$1.97 & 89.51$\pm$0.58 & 80.42$\pm$0.89 & 89.22$\pm$1.38 & 90.12$\pm$0.42& {\bf 92.38$\pm$0.41} \\
     truck & 86.32$\pm$0.89 & 83.29$\pm$0.07  & 75.03$\pm$11.55 & 87.58$\pm$0.78 & 87.02$\pm$0.44 & 90.32$\pm$0.74 & 94.17$\pm$0.25&{\bf 95.35$\pm$0.54}\\ \midrule
     {\it average} & 75.94$\pm$0.91 & 74.83$\pm$0.14  & 62.73$\pm$0.81 & 77.44$\pm$0.62 & 71.27$\pm$0.33 & 82.75$\pm$1.11 &84.64$\pm$0.10 & {\bf 84.71$\pm$0.27} \\ \bottomrule
   \end{tabular}%
   }
   \caption{AUROC (\%) on CIFAR-10 for UAD using ResNet-50 as extractor when $p=0.5$. The best performance is in bold.}
\end{table*}

\begin{table*}[htbp]
    \centering
    \scalebox{0.8}{
    \begin{tabular}{cccccccc}
    \toprule
    Dataset& Inlier class name & IF & OCSVM & DAGMM & RSRAE & SLA (ours) & SLA$^2$P (ours)  \\ \hline
    \multirow{21}{*}{20 Newsgroups}& alt.atheism & 55.29$\pm$3.70 & 95.59$\pm$0.00  & 59.36$\pm$3.59 & 94.52$\pm$1.24& 96.11$\pm$0.19 & {\bf 96.20$\pm$0.34} \\
    & comp.graphics & 68.14$\pm$2.53 & 65.23$\pm$0.00  & 58.41$\pm$3.04 & 84.23$\pm$2.55 & 83.73$\pm$0.71 & {\bf 88.62$\pm$0.33}  \\
    & comp.os.ms-windows.misc & 64.75$\pm$1.71 & 82.64$\pm$0.00  & 56.84$\pm$4.48 & 88.45$\pm$2.51 & 89.35$\pm$0.87 & {\bf 91.98$\pm$0.27}\\
    & comp.sys.ibm.pc.hardware & 63.77$\pm$2.39 & 73.21$\pm$0.00  & 61.36$\pm$5.04& 84.36$\pm$5.44 & 86.89$\pm$0.90 & {\bf 90.16$\pm$0.14}\\
    & comp.sys.mac.hardware & 60.83$\pm$2.33 & 73.96$\pm$0.00  & 56.03$\pm$2.13& 88.69$\pm$2.80 & 88.06$\pm$0.68 & {\bf 91.86$\pm$0.21}\\
    & comp.windows.x & 65.17$\pm$5.17 & 69.78$\pm$0.00 & 54.70$\pm$5.14& 86.52$\pm$1.91 & 91.21$\pm$0.50 & {\bf 94.07$\pm$0.30}\\
    & misc.forsale & 67.01$\pm$1.57 & 73.47$\pm$0.00 & 58.39$\pm$5.40& 86.99$\pm$4.28 & 85.76$\pm$0.82 & {\bf 90.74$\pm$0.53}\\
    & rec.autos & 55.60$\pm$3.88 & 70.76$\pm$0.00 & 57.86$\pm$5.02& 87.45$\pm$4.69 & 88.33$\pm$0.97 & {\bf 91.46$\pm$0.38}\\
    & rec.motorcycles & 52.12$\pm$4.00 & 87.94$\pm$0.00 & 54.78$\pm$3.63& 92.51$\pm$1.90 & 93.40$\pm$0.50 & {\bf 95.66$\pm$0.23}\\
    & rec.sport.baseball & 62.46$\pm$2.82 & 82.73$\pm$0.00 & 61.20$\pm$3.65& 91.39$\pm$1.68 & 92.91$\pm$0.37 & {\bf 95.29$\pm$0.18}\\
    & rec.sport.hockey & 59.82$\pm$2.13 & 92.38$\pm$0.00 & 53.36$\pm$10.76& 94.98$\pm$0.99 & 95.75$\pm$0.47 & {\bf 97.12$\pm$0.14}\\
    & sci.crypt & 46.89$\pm$2.39 & 93.50$\pm$0.00 & 61.73$\pm$2.82& 97.23$\pm$0.55 & 95.58$\pm$0.29 & {\bf 96.14$\pm$0.14}\\
    & sci.electronics & 59.80$\pm$2.38 & 61.22$\pm$0.00 & 58.41$\pm$4.79& 76.29$\pm$3.05 & 79.99$\pm$1.03 & {\bf 86.73$\pm$0.38}\\
    & sci.med & 56.97$\pm$2.58 & 77.48$\pm$0.00 & 47.21$\pm$5.49& 83.93$\pm$0.95 & 89.64$\pm$0.22 & {\bf 91.89$\pm$0.48}\\
    & sci.space & 53.04$\pm$3.65 & 87.06$\pm$0.00 & 48.34$\pm$4.43& 90.84$\pm$1.93 & 92.81$\pm$0.39 & {\bf 93.71$\pm$0.24}\\
    & soc.religion.christian & 51.60$\pm$3.59 & 91.81$\pm$0.00 & 51.52$\pm$8.33& 90.45$\pm$1.72 & 93.00$\pm$0.35 & {\bf 94.38$\pm$0.26}\\
    & talk.politics.guns & 49.52$\pm$2.51 & 92.25$\pm$0.00 & 56.55$\pm$7.24& 92.67$\pm$1.67 & 93.42$\pm$0.29 & {\bf 95.44$\pm$0.12}\\
    & talk.politics.mideast & 43.62$\pm$3.10 & 96.00$\pm$0.00 & 45.74$\pm$6.15& 94.92$\pm$2.20 & 96.71$\pm$0.14 & {\bf 97.53$\pm$0.12}\\
    & talk.politics.misc & 43.73$\pm$2.04 & 86.44$\pm$0.00 & 48.90$\pm$6.42& 90.04$\pm$1.24 & 91.84$\pm$0.41 & {\bf 93.20$\pm$0.29}\\
    & talk.religion.misc & 48.89$\pm$5.79 & 87.45$\pm$0.00 & 46.01$\pm$5.13& 84.17$\pm$3.43 & 90.84$\pm$0.35 & {\bf 91.89$\pm$0.17} \\ \cline{2-8} 
    & {\it average} & 56.45$\pm$0.67 & 82.04$\pm$0.00 & 54.84$\pm$2.27& 89.03$\pm$2.34 & 90.77$\pm$0.12 & {\bf 93.20$\pm$0.07}\\ \midrule
    \multirow{5}{*}{Reuters-21578}& exchanges & 75.79$\pm$0.94 & 96.81$\pm$0.00 & 77.35$\pm$7.09& 96.66$\pm$1.34 & 96.89$\pm$0.25 & {\bf 98.27$\pm$0.12} \\
    & organizations & 66.01$\pm$3.94 & 69.98$\pm$0.00 & 62.53$\pm$5.03& 76.50$\pm$4.25 & 80.73$\pm$0.31 & {\bf 86.72$\pm$0.57}\\
    & people & 58.43$\pm$3.70 & 97.81$\pm$0.00 & 63.61$\pm$8.67& 95.67$\pm$0.14 & 97.77$\pm$0.08 & {\bf 98.23$\pm$0.15}\\
    & places & 52.74$\pm$1.89 & 96.70$\pm$0.00 & 59.74$\pm$7.47& 96.47$\pm$2.09 & 97.12$\pm$0.07 & {\bf 97.93$\pm$0.11}\\
    & topics & 43.53$\pm$3.31 & 93.46$\pm$0.00 & 62.11$\pm$8.28& 96.90$\pm$0.70 & 95.03$\pm$0.24 & {\bf 97.71$\pm$0.08}\\ \cline{2-8} 
    & {\it average} & 59.30$\pm$0.82 & 90.95$\pm$0.00 & 65.07$\pm$4.12& 92.44$\pm$1.70 & 93.51$\pm$0.09& {\bf 95.77$\pm$0.12}\\ \bottomrule
    \end{tabular}%
    }
    \caption{AUROC (\%) on text datasets for UAD when $p=0.1$. The best performance is in bold.}
\end{table*}

\begin{table*}[htbp]
    \centering
    \scalebox{0.8}{
    \begin{tabular}{cccccccc}
    \toprule
    Dataset& Inlier class name & IF& OCSVM & DAGMM & RSRAE & SLA (ours) & SLA$^2$P (ours)  \\ \hline
    \multirow{21}{*}{20 Newsgroups}& alt.atheism & 12.41$\pm$0.30 & 78.86$\pm$0.00  & 13.12$\pm$1.92& 73.39$\pm$3.24 & 79.37$\pm$1.62 & {\bf 83.86$\pm$0.93}\\
    & comp.graphics & 18.61$\pm$1.73& 18.66$\pm$0.00  & 14.40$\pm$2.31& 36.56$\pm$5.32 & 31.62$\pm$3.30 & {\bf 63.21$\pm$1.62}\\
    & comp.os.ms-windows.misc & 16.01$\pm$0.77 & 34.31$\pm$0.00 & 11.70$\pm$1.42& 40.66$\pm$5.97 & 43.19$\pm$3.50 & {\bf 67.91$\pm$1.50}\\
    & comp.sys.ibm.pc.hardware & 21.65$\pm$1.63 & 26.66$\pm$0.00 & 18.60$\pm$4.95& 35.72$\pm$9.73 & 37.26$\pm$2.86 & {\bf 64.24$\pm$1.50}\\
    & comp.sys.mac.hardware & 21.46$\pm$3.09 & 28.15$\pm$0.00 & 13.96$\pm$6.04& 41.90$\pm$9.32 & 37.85$\pm$3.00 & {\bf 69.89$\pm$0.62}\\
    & comp.windows.x & 16.19$\pm$2.55 & 19.63$\pm$0.00 & 13.50$\pm$4.18& 36.58$\pm$2.78 & 40.45$\pm$3.01 & {\bf 66.94$\pm$0.88}\\
    & misc.forsale & 23.36$\pm$1.56 & 29.37$\pm$0.00 & 15.40$\pm$3.30 & 45.05$\pm$6.36 & 35.80$\pm$3.90 & {\bf 65.81$\pm$2.33}\\
    & rec.autos & 14.24$\pm$0.78 & 29.01$\pm$0.00 & 16.74$\pm$8.13& 41.00$\pm$9.34 & 39.49$\pm$3.19 & {\bf 65.68$\pm$1.22}\\
    & rec.motorcycles & 13.65$\pm$2.01 & 62.64$\pm$0.00 & 13.33$\pm$2.48& 68.45$\pm$5.61 & 67.31$\pm$3.54 & {\bf 78.87$\pm$1.11}\\
    & rec.sport.baseball & 16.72$\pm$1.20 & 50.43$\pm$0.00 & 13.77$\pm$2.66& 54.76$\pm$3.15 & 55.98$\pm$1.14 & {\bf 76.46$\pm$1.14}\\
    & rec.sport.hockey & 17.89$\pm$0.74 & 54.61$\pm$0.00 & 14.66$\pm$7.16& 63.95$\pm$4.02 & 60.47$\pm$2.78 & {\bf 78.86$\pm$0.82}\\
    & sci.crypt & 9.45$\pm$0.77 & 64.29$\pm$0.00 & 13.91$\pm$1.25& 72.41$\pm$3.19 & 67.49$\pm$1.77 & {\bf 78.35$\pm$0.67}\\
    & sci.electronics & 22.18$\pm$2.45& 17.22$\pm$0.00 & 13.66$\pm$3.73& 26.17$\pm$2.08 & 26.27$\pm$4.36 & {\bf 57.72$\pm$0.63}\\
    & sci.med & 12.97$\pm$1.28 & 34.09$\pm$0.00 & 9.07$\pm$1.22& 36.44$\pm$2.60 & 43.05$\pm$3.53 & {\bf 66.93$\pm$0.68}\\
    & sci.space & 11.58$\pm$1.85 & 48.89$\pm$0.00 & 9.58$\pm$2.61& 48.53$\pm$4.43 & 55.44$\pm$3.65 & {\bf 73.91$\pm$0.65}\\
    & soc.religion.christian & 12.45$\pm$0.73 & 49.12$\pm$0.00 & 11.03$\pm$1.98& 49.60$\pm$3.50 & 48.09$\pm$2.01 & {\bf 70.85$\pm$1.19}\\
    & talk.politics.guns & 10.80$\pm$0.87 & 64.36$\pm$0.00 & 12.34$\pm$4.22& 58.73$\pm$4.68 & 66.14$\pm$1.67 & {\bf 77.53$\pm$0.40}\\
    & talk.politics.mideast & 9.51$\pm$1.56& 67.09$\pm$0.00 & 9.59$\pm$2.47& 68.82$\pm$8.68 & 70.68$\pm$1.24 & {\bf 83.68$\pm$1.30}\\
    & talk.politics.misc & 9.83$\pm$0.88 & 52.73$\pm$0.00 & 8.58$\pm$1.22& 56.43$\pm$4.94 & 53.83$\pm$2.23 & {\bf 72.10$\pm$0.55}\\
    & talk.religion.misc & 10.92$\pm$1.26 & 46.30$\pm$0.00 & 8.21$\pm$0.93& 40.69$\pm$6.71 & 51.59$\pm$1.56 & {\bf 70.65$\pm$0.77}\\
    \cline{2-8} 
    & {\it average} & 15.09$\pm$0.22 & 43.82$\pm$0.00 & 12.76$\pm$1.06& 49.79$\pm$5.28 & 50.57$\pm$0.65 & {\bf 71.77$\pm$0.34}\\ \midrule
    \multirow{5}{*}{Reuters-21578}& exchanges & 36.21$\pm$0.64& 71.23$\pm$0.00 & 38.72$\pm$11.34& 73.37$\pm$9.35 & 71.77$\pm$3.63 & {\bf 87.68$\pm$1.00} \\
    & organizations & 25.86$\pm$4.09& 20.11$\pm$0.00 & 18.23$\pm$5.66& 32.19$\pm$6.35 & 34.19$\pm$0.78 & {\bf 60.47$\pm$0.87}\\
    & people & 11.94$\pm$1.39 & 71.16$\pm$0.00 & 17.14$\pm$5.29& 67.57$\pm$4.98 & 71.39$\pm$1.83 & {\bf 85.48$\pm$0.86}\\
    & places & 11.85$\pm$1.38 & 68.71$\pm$0.00 & 15.65$\pm$5.04& 70.04$\pm$6.79 & 68.31$\pm$3.35 & {\bf 85.00$\pm$0.92}\\
    & topics & 7.82$\pm$0.45 & 77.23$\pm$0.00 & 21.93$\pm$12.22& 79.04$\pm$6.10 & 77.26$\pm$1.53 & {\bf 88.35$\pm$0.82} \\
    \cline{2-8} 
    & {\it average} & 18.74$\pm$0.93 & 61.69$\pm$0.00 & 22.33$\pm$3.62& 64.44$\pm$6.71 & 64.58$\pm$1.34 & {\bf 81.40$\pm$0.32}\\
        \bottomrule
    \end{tabular}%
    }
    \caption{AUPR (\%) on text datasets for UAD when $p=0.1$. The best performance is in bold.}
\end{table*}

\begin{table*}[htbp]
    \centering
    \scalebox{0.8}{
    \begin{tabular}{cccccccc}
    \toprule
    Dataset& Inlier class name & IF & OCSVM & DAGMM & RSRAE & SLA (ours) & SLA$^2$P (ours) \\ \hline
    \multirow{21}{*}{20 Newsgroups}& alt.atheism & 56.28$\pm$1.11 & 88.77$\pm$0.00  & 58.88$\pm$8.15& 94.79$\pm$0.83 & 94.01$\pm$0.20 & {\bf 95.36$\pm$0.19}\\
    & comp.graphics & 67.45$\pm$1.50 & 64.61$\pm$0.00 & 56.44$\pm$3.78& 81.69$\pm$1.27 & 78.01$\pm$0.84 & {\bf 85.31$\pm$0.78}\\
    & comp.os.ms-windows.misc & 65.54$\pm$3.20 & 75.41$\pm$0.00 & 58.71$\pm$4.46& 88.43$\pm$0.72 & 86.04$\pm$0.31 & {\bf 89.69$\pm$0.30}\\
    & comp.sys.ibm.pc.hardware & 66.70$\pm$2.86 & 67.15$\pm$0.00 & 54.50$\pm$2.96& 84.51$\pm$1.28 & 81.46$\pm$0.85 & {\bf 86.95$\pm$0.93}\\
    & comp.sys.mac.hardware & 66.15$\pm$1.20 & 67.72$\pm$0.00 & 56.47$\pm$2.82& 87.35$\pm$1.71 & 82.56$\pm$0.49 & {\bf 88.18$\pm$0.41}\\
    & comp.windows.x & 64.89$\pm$1.87 & 63.71$\pm$0.00 & 56.31$\pm$1.03& 84.89$\pm$0.82 & 81.89$\pm$0.53 & {\bf 87.55$\pm$0.32}\\
    & misc.forsale & 66.26$\pm$2.01 & 64.54$\pm$0.00 & 59.40$\pm$4.67& {\bf 86.67$\pm$1.01} & 78.77$\pm$0.83 & 85.26$\pm$0.74\\
    & rec.autos & 57.95$\pm$1.62 & 69.99$\pm$0.00 & 51.62$\pm$3.82& 86.59$\pm$1.73 & 83.80$\pm$0.43 & {\bf 88.41$\pm$0.55}\\
    & rec.motorcycles & 57.71$\pm$1.11& 73.54$\pm$0.00 & 50.54$\pm$3.15& 92.41$\pm$1.00 & 89.02$\pm$0.41 & {\bf 92.45$\pm$0.31}\\
    & rec.sport.baseball & 64.57$\pm$2.06 & 72.12$\pm$0.00 & 58.86$\pm$2.88& 90.86$\pm$1.92 & 88.19$\pm$0.59 & {\bf 91.75$\pm$0.34}\\
    & rec.sport.hockey & 61.35$\pm$1.78 & 85.87$\pm$0.00 & 56.12$\pm$3.93& 94.25$\pm$0.95 & 92.45$\pm$0.49 & {\bf 94.44$\pm$0.41}\\
    & sci.crypt & 53.63$\pm$1.62 & 90.63$\pm$0.00 & 55.95$\pm$5.21& {\bf 96.89$\pm$0.32} & 93.59$\pm$0.19 & 94.88$\pm$0.20\\
    & sci.electronics & 59.84$\pm$2.64 & 57.25$\pm$0.00 & 54.57$\pm$3.14& 77.03$\pm$2.45 & 72.75$\pm$0.11 & {\bf 80.84$\pm$0.36}\\
    & sci.med & 55.63$\pm$1.61 & 71.56$\pm$0.00 & 47.53$\pm$2.51& 83.20$\pm$1.87 & 84.28$\pm$0.71 & {\bf 88.71$\pm$0.61}\\
    & sci.space & 54.62$\pm$1.82 & 77.32$\pm$0.00 & 53.03$\pm$7.49& 91.16$\pm$0.98 & 89.11$\pm$0.16 & {\bf 91.80$\pm$0.25}\\
    & soc.religion.christian & 52.36$\pm$2.13 & 82.46$\pm$0.00 & 61.11$\pm$5.88& 90.11$\pm$1.59 & 87.99$\pm$0.44 & {\bf 90.39$\pm$0.47}\\
    & talk.politics.guns & 48.15$\pm$1.56 & 83.58$\pm$0.00 & 52.52$\pm$5.13& 92.55$\pm$0.88 & 91.46$\pm$0.19 & {\bf 93.57$\pm$0.32}\\
    & talk.politics.mideast & 47.20$\pm$3.43 & 86.40$\pm$0.00 & 43.65$\pm$5.28& {\bf 95.30$\pm$1.19} & 93.39$\pm$0.31 & 94.94$\pm$0.21\\
    & talk.politics.misc & 47.73$\pm$3.22 & 80.08$\pm$0.00 & 47.48$\pm$4.14& 91.81$\pm$1.58 & 89.22$\pm$0.40 & {\bf 91.84$\pm$0.34}\\
    & talk.religion.misc & 52.96$\pm$1.35 & 78.64$\pm$0.00 & 47.62$\pm$2.46& 86.63$\pm$1.93 & 87.86$\pm$0.14 & {\bf 90.87$\pm$0.21}\\ \cline{2-8} 
    & {\it average} & 58.35$\pm$0.45 & 75.07$\pm$0.00 & 54.07$\pm$0.57& 88.86$\pm$1.30 & 86.29$\pm$0.11 & {\bf 90.16$\pm$0.11}\\ \midrule
    \multirow{5}{*}{Reuters-21578}& exchanges & 76.45$\pm$1.50 & 90.49$\pm$0.00  & 71.75$\pm$5.28& 95.18$\pm$0.44 & 95.27$\pm$0.15 & {\bf 97.06$\pm$0.22}\\
    & organizations & 67.49$\pm$3.11 & 57.04$\pm$0.00 &  60.68$\pm$4.51& 63.66$\pm$2.88 & 60.48$\pm$0.32  & {\bf 89.91$\pm$0.75}\\
    & people & 58.40$\pm$2.89 & 89.15$\pm$0.00 & 52.18$\pm$6.94& 93.75$\pm$1.22 & 94.72$\pm$0.16 & {\bf 96.78$\pm$0.12}\\
    & places & 48.81$\pm$2.12 & 88.10$\pm$0.00 & 56.93$\pm$5.25& 94.64$\pm$1.00 & 94.82$\pm$0.10 & {\bf 97.06$\pm$0.16}\\
    & topics & 43.18$\pm$2.63 & 91.41$\pm$0.00 & 55.11$\pm$8.95& 96.25$\pm$0.49 & 95.04$\pm$0.16  & {\bf 96.79$\pm$0.17}\\ \cline{2-8} 
    & {\it average} & 58.86$\pm$0.74 & 83.24$\pm$0.00 & 59.33$\pm$2.68& 88.70$\pm$1.21 & 88.07$\pm$0.07 & {\bf 95.52$\pm$0.13}\\ \bottomrule
    \end{tabular}%
    }
    \caption{AUROC (\%) on text datasets for UAD when $p=0.3$. The best performance is in bold.}
\end{table*}

\begin{table*}[htbp]
    \centering
    \scalebox{0.8}{
    \begin{tabular}{cccccccc}
    \toprule
    Dataset& Inlier class name & IF& OCSVM & DAGMM & RSRAE & SLA (ours) & SLA$^2$P (ours) \\ \hline
    \multirow{21}{*}{20 Newsgroups}& alt.atheism & 32.34$\pm$1.21 & 79.18$\pm$0.00  & 30.32$\pm$7.71 & 87.05$\pm$1.62 & 85.45$\pm$1.17 & {\bf 89.27$\pm$0.87} \\
    & comp.graphics & 38.58$\pm$0.91 & 32.30$\pm$0.00  & 29.75$\pm$3.33 & 51.82$\pm$1.83 & 47.93$\pm$1.08 & {\bf 64.49$\pm$1.70}\\
    & comp.os.ms-windows.misc & 35.88$\pm$2.29 & 45.57$\pm$0.00  & 28.80$\pm$3.46 & 64.87$\pm$1.82 & 58.19$\pm$0.98 & {\bf 70.33$\pm$0.46} \\
    & comp.sys.ibm.pc.hardware & 46.80$\pm$3.94 & 38.06$\pm$0.00  & 28.80$\pm$1.53 & 57.72$\pm$1.67 & 52.38$\pm$0.87 & {\bf 67.26$\pm$1.83} \\
    & comp.sys.mac.hardware & 45.69$\pm$1.99 & 39.99$\pm$0.00  & 29.23$\pm$1.27 & 62.98$\pm$4.59 & 55.00$\pm$1.39 & {\bf 69.85$\pm$1.06}\\
    & comp.windows.x & 37.76$\pm$1.25 & 34.38$\pm$0.00 & 26.95$\pm$0.41 & 56.12$\pm$2.17 & 51.28$\pm$0.98 & {\bf 66.50$\pm$0.79}\\
    & misc.forsale & 44.86$\pm$2.90 & 38.19$\pm$0.00 & 32.07$\pm$3.83 & 63.22$\pm$1.39 & 50.60$\pm$1.76 & {\bf 66.44$\pm$1.98}\\
    & rec.autos & 34.70$\pm$1.98 & 44.40$\pm$0.00 & 24.94$\pm$2.31 & 61.43$\pm$3.65 & 56.68$\pm$1.01 & {\bf 68.83$\pm$1.00} \\
    & rec.motorcycles & 36.40$\pm$1.00 & 58.58$\pm$0.00 & 24.22$\pm$1.81& 80.87$\pm$2.31 & 73.08$\pm$1.46 & {\bf 82.14$\pm$0.62} \\
    & rec.sport.baseball & 38.35$\pm$3.57 & 51.58$\pm$0.00 & 30.98$\pm$3.30 & 74.09$\pm$4.16 & 68.20$\pm$1.85 & {\bf 77.87$\pm$0.79}\\
    & rec.sport.hockey & 38.20$\pm$1.91 & 63.70$\pm$0.00 & 31.01$\pm$3.55 & 78.36$\pm$2.29 & 73.33$\pm$2.04 & {\bf 81.09$\pm$1.77} \\
    & sci.crypt & 27.95$\pm$0.82 & 74.10$\pm$0.00 & 28.50$\pm$5.65 & {\bf 87.00$\pm$0.81} & 78.60$\pm$0.44 & 83.25$\pm$0.54 \\
    & sci.electronics & 40.90$\pm$2.68 & 32.59$\pm$0.00 & 29.02$\pm$3.64 & 47.55$\pm$3.10 & 41.80$\pm$0.81 & {\bf 57.49$\pm$1.23}\\
    & sci.med & 29.88$\pm$1.91 & 47.39$\pm$0.00 & 22.64$\pm$2.18 & 55.24$\pm$4.00 & 57.41$\pm$1.62 & {\bf 70.40$\pm$2.01}\\
    & sci.space & 28.94$\pm$1.02 & 55.17$\pm$0.00 & 25.43$\pm$4.01 & 68.36$\pm$1.66 & 68.30$\pm$1.12 & {\bf 77.34$\pm$1.41}\\
    & soc.religion.christian & 30.78$\pm$1.47 & 56.53$\pm$0.00 & 33.81$\pm$8.10 & 70.18$\pm$2.76 & 62.88$\pm$1.52 & {\bf 71.69$\pm$2.08}\\
    & talk.politics.guns & 25.36$\pm$1.23 & 67.53$\pm$0.00 & 25.25$\pm$5.08 & 77.79$\pm$1.58 & 76.87$\pm$0.33  & {\bf 83.04$\pm$0.51}\\
    & talk.politics.mideast & 23.78$\pm$1.88 & 67.66$\pm$0.00 & 20.52$\pm$2.50 & {\bf 83.25$\pm$3.11} & 77.06$\pm$0.81 & 82.88$\pm$0.67\\
    & talk.politics.misc & 25.65$\pm$1.41& 62.98$\pm$0.00 & 22.17$\pm$3.27 & 77.70$\pm$3.04 & 70.95$\pm$1.33 & {\bf 78.71$\pm$1.49}\\
    & talk.religion.misc & 28.43$\pm$1.25& 55.68$\pm$0.00 & 23.20$\pm$2.12 & 66.59$\pm$2.39 & 65.63$\pm$0.53 & {\bf 75.39$\pm$0.70}\\ \cline{2-8} 
    & {\it average} & 34.56$\pm$0.72 & 52.28$\pm$0.00 & 27.38$\pm$0.82 & 68.61$\pm$2.50 & 63.58$\pm$0.09 & {\bf 74.21$\pm$0.17}\\ \midrule
    \multirow{5}{*}{Reuters-21578}& exchanges & 57.39$\pm$3.64 & 71.58$\pm$0.00 & 51.14$\pm$8.66 & 83.75$\pm$2.41 & 81.89$\pm$0.87 & {\bf 90.39$\pm$0.48} \\
    & organizations & 49.37$\pm$1.98 & 28.15$\pm$0.00 & 33.60$\pm$4.60 & 36.99$\pm$4.06 & 31.43$\pm$0.88  & {\bf 84.52$\pm$1.07} \\
    & people & 26.28$\pm$2.34 & 69.27$\pm$0.00 & 27.53$\pm$4.12 & 78.52$\pm$2.78 & 79.64$\pm$1.30 & {\bf 90.09$\pm$0.41} \\
    & places & 26.17$\pm$1.90 & 67.45$\pm$0.00 & 28.68$\pm$3.25& 78.66$\pm$2.17 & 79.17$\pm$1.36 & {\bf 90.67$\pm$0.65} \\
    & topics & 20.71$\pm$1.80 & 79.80$\pm$0.00 & 27.12$\pm$6.92 & 87.45$\pm$1.56 & 87.62$\pm$0.75 & {\bf 92.73$\pm$0.39}\\ \cline{2-8} 
    & {\it average} & 35.99$\pm$0.70 & 63.25$\pm$0.00 & 33.61$\pm$2.77 & 73.07$\pm$2.60 & 71.95$\pm$0.42 & {\bf 89.68$\pm$0.20} \\ \bottomrule
    \end{tabular}%
    }
    \caption{AUPR (\%) on text datasets for UAD when $p=0.3$. The best performance is in bold.}
\end{table*}

\begin{table*}[htbp]
    \centering
    \scalebox{0.8}{
    \begin{tabular}{cccccccc}
    \toprule
    Dataset& Inlier class name & IF & OCSVM & DAGMM & RSRAE & SLA (ours) & SLA$^2$P (ours) \\ \hline
    \multirow{21}{*}{20 Newsgroups}& alt.atheism & 55.37$\pm$2.51 & 84.37$\pm$0.00  & 56.22$\pm$6.82 & {\bf 93.74$\pm$0.65} & 92.45$\pm$0.16 & 93.29$\pm$0.10 \\
    & comp.graphics & 66.44$\pm$2.36 & 55.84$\pm$0.00  & 52.66$\pm$5.06 & 77.90$\pm$1.89 & 75.69$\pm$0.61 & {\bf 79.34$\pm$0.43} \\
    & comp.os.ms-windows.misc & 67.75$\pm$2.34 & 68.66$\pm$0.00 & 54.80$\pm$6.22 & 86.54$\pm$0.82 & 84.54$\pm$0.36 & {\bf 86.62$\pm$0.37}  \\
    & comp.sys.ibm.pc.hardware & 65.91$\pm$1.58 & 65.63$\pm$0.00 & 59.71$\pm$4.35 & 82.21$\pm$1.60 & 79.41$\pm$0.57  & {\bf 82.21$\pm$0.45}  \\
    & comp.sys.mac.hardware & 66.34$\pm$1.31 & 63.12$\pm$0.00 & 56.00$\pm$2.59 & {\bf 85.15$\pm$1.24} & 80.53$\pm$0.47 & 82.87$\pm$0.27  \\
    & comp.windows.x & 63.82$\pm$1.15 & 61.06$\pm$0.00 & 55.17$\pm$2.02 & 81.52$\pm$1.27 & 78.96$\pm$0.18 & {\bf 81.92$\pm$0.36} \\
    & misc.forsale & 67.50$\pm$1.99 & 62.82$\pm$0.00 & 52.94$\pm$2.67 & {\bf 82.16$\pm$0.72} & 76.11$\pm$0.42 & 79.49$\pm$0.40   \\
    & rec.autos & 55.86$\pm$1.07 & 70.15$\pm$0.00 & 49.91$\pm$3.20 & {\bf 84.62$\pm$1.59} & 81.96$\pm$0.62  & 84.40$\pm$0.37  \\
    & rec.motorcycles & 55.01$\pm$0.55 & 70.27$\pm$0.00 & 52.22$\pm$3.01 & {\bf 89.93$\pm$2.14} & 86.92$\pm$0.31 & 88.64$\pm$0.35  \\
    & rec.sport.baseball & 61.85$\pm$2.65 & 69.67$\pm$0.00 & 56.25$\pm$6.17 & {\bf 88.45$\pm$1.81} & 86.01$\pm$0.27 & 87.81$\pm$0.19  \\
    & rec.sport.hockey & 59.34$\pm$2.73 & 76.95$\pm$0.00 & 59.53$\pm$3.58 & {\bf 92.81$\pm$0.88} & 90.10$\pm$0.15 & 91.25$\pm$0.12 \\
    & sci.crypt & 51.13$\pm$1.73 & 83.21$\pm$0.00 & 52.09$\pm$10.96 & {\bf 96.29$\pm$0.32} & 92.63$\pm$0.11 & 93.48$\pm$0.17 \\
    & sci.electronics & 59.41$\pm$2.32 & 59.95$\pm$0.00 & 49.29$\pm$3.30 & 73.16$\pm$2.59 & 71.19$\pm$0.36 & {\bf 75.72$\pm$0.39} \\
    & sci.med & 55.23$\pm$2.43 & 64.81$\pm$0.00 & 46.71$\pm$5.30 & 80.61$\pm$1.20 & 81.55$\pm$0.38 & {\bf 84.00$\pm$0.29} \\
    & sci.space & 52.55$\pm$1.61 & 74.08$\pm$0.00 & 48.36$\pm$3.08 & {\bf 89.45$\pm$0.70} & 87.76$\pm$0.41 & 89.27$\pm$0.24 \\
    & soc.religion.christian & 50.07$\pm$0.92 & 78.20$\pm$0.00 & 57.08$\pm$5.01 & {\bf 89.15$\pm$1.60} & 87.14$\pm$0.31 & 88.78$\pm$0.31  \\
    & talk.politics.guns & 44.45$\pm$1.68 & 80.81$\pm$0.00 & 46.17$\pm$5.30 & 91.27$\pm$1.20 & 90.20$\pm$0.32 & {\bf 91.44$\pm$0.17} \\
    & talk.politics.mideast & 46.62$\pm$2.87 & 82.79$\pm$0.00 & 46.96$\pm$4.29 & {\bf 94.02$\pm$1.08} & 91.90$\pm$0.18 & 92.80$\pm$0.13 \\
    & talk.politics.misc & 45.09$\pm$2.11 & 78.98$\pm$0.00 & 40.78$\pm$2.39 & 89.48$\pm$1.47 & 88.45$\pm$0.19 & {\bf 89.48$\pm$0.23} \\
    & talk.religion.misc & 50.29$\pm$2.74 & 75.32$\pm$0.00 & 45.86$\pm$6.69& 86.11$\pm$1.46 & 86.14$\pm$0.49 & {\bf 88.13$\pm$0.38} \\ \cline{2-8} 
    & {\it average} & 57.00$\pm$0.33 & 71.33$\pm$0.00 & 51.94$\pm$1.02 & {\bf 86.73$\pm$1.31} & 84.48$\pm$0.11 & 86.55$\pm$0.06 \\ \midrule
    \multirow{5}{*}{Reuters-21578}& exchanges & 74.68$\pm$2.24 & 80.03$\pm$0.00 & 67.68$\pm$5.85 & 92.53$\pm$1.68 & 91.76$\pm$0.23 & {\bf 93.83$\pm$0.22} \\
    & organizations & 65.09$\pm$1.57 & 48.58$\pm$0.00 & 62.84$\pm$4.04 & 51.99$\pm$2.66 & 46.62$\pm$0.33 & {\bf 55.99$\pm$0.54} \\
    & people & 53.49$\pm$3.23 & 79.78$\pm$0.00 & 60.71$\pm$5.49 & 89.64$\pm$3.78 & 91.05$\pm$0.15 & {\bf 93.25$\pm$0.16} \\
    & places & 48.26$\pm$1.21 & 79.49$\pm$0.00 & 60.54$\pm$4.41 & 87.54$\pm$5.76 & 90.53$\pm$0.14 & {\bf 92.54$\pm$0.12} \\
    & topics & 37.70$\pm$2.64 & 86.50$\pm$0.00 & 58.30$\pm$9.01 & {\bf 95.27$\pm$0.43} & 94.41$\pm$0.17 & 94.84$\pm$0.18 \\ \cline{2-8} 
    & {\it average} & 55.85$\pm$0.87 & 74.88$\pm$0.00 & 62.01$\pm$2.54 & 83.39$\pm$2.86 & 82.87$\pm$0.06 & {\bf 86.09$\pm$0.13} \\ \bottomrule
    \end{tabular}%
    }
    \caption{AUROC (\%) on text datasets for UAD when $p=0.5$. The best performance is in bold.}
\end{table*}

\begin{table*}[htbp]
    \centering
    \scalebox{0.8}{
    \begin{tabular}{cccccccc}
    \toprule
    Dataset& Inlier class name & IF & OCSVM & DAGMM & RSRAE & SLA (ours) & SLA$^2$P (ours) \\ \hline
    \multirow{21}{*}{20 Newsgroups}& alt.atheism & 41.80$\pm$1.28 & 79.72$\pm$0.00  & 38.23$\pm$5.56 & 89.16$\pm$0.83 & 88.14$\pm$0.76 & {\bf 90.11$\pm$0.50} \\ 
    & comp.graphics & 50.38$\pm$1.99 & 39.35$\pm$0.00 & 36.54$\pm$3.53 & 58.07$\pm$1.96 & 57.25$\pm$1.74 & {\bf 68.75$\pm$0.90} \\
    & comp.os.ms-windows.misc & 49.54$\pm$1.91 & 51.78$\pm$0.00 & 37.20$\pm$4.28 & 69.13$\pm$1.55 & 68.15$\pm$0.86 & {\bf 75.75$\pm$0.64} \\
    & comp.sys.ibm.pc.hardware & 55.09$\pm$1.86 & 49.65$\pm$0.00 & 44.41$\pm$4.83 & 64.54$\pm$3.11 & 61.92$\pm$0.97 & {\bf 71.14$\pm$0.72}\\
    & comp.sys.mac.hardware & 56.40$\pm$1.03 & 48.76$\pm$0.00 & 40.85$\pm$4.84 & 70.01$\pm$1.79 & 64.67$\pm$1.25 & {\bf 72.49$\pm$0.53} \\
    & comp.windows.x & 48.97$\pm$0.93 & 44.21$\pm$0.00  & 38.15$\pm$1.59 & 61.97$\pm$2.37 & 60.07$\pm$1.12 & {\bf 71.15$\pm$0.69} \\
    & misc.forsale & 55.78$\pm$2.54 & 46.97$\pm$0.00 & 37.19$\pm$3.33 & 66.20$\pm$1.36 & 58.45$\pm$1.07 & {\bf 69.10$\pm$0.81} \\
    & rec.autos & 42.72$\pm$1.51 & 56.07$\pm$0.00 & 32.65$\pm$2.24 & 68.25$\pm$2.42 & 66.11$\pm$1.85 & {\bf 74.91$\pm$0.90} \\
    & rec.motorcycles & 44.48$\pm$1.12 & 63.35$\pm$0.00 & 35.54$\pm$2.10 & 82.33$\pm$2.47 & 78.90$\pm$1.05 & {\bf 82.95$\pm$0.73}\\
    & rec.sport.baseball & 46.19$\pm$1.98 & 57.15$\pm$0.00 & 39.12$\pm$6.25 & 77.70$\pm$3.58 & 74.16$\pm$1.01 &  {\bf 80.34$\pm$0.42}\\
    & rec.sport.hockey & 46.73$\pm$3.28 & 64.72$\pm$0.00 & 42.38$\pm$2.78 & 83.08$\pm$1.32 & 78.62$\pm$0.50 & {\bf 83.97$\pm$0.37}\\
    & sci.crypt & 36.27$\pm$1.20 & 74.18$\pm$0.00 & 35.10$\pm$6.49 & {\bf 89.82$\pm$0.55} & 84.09$\pm$0.89 & 88.19$\pm$0.40\\
    & sci.electronics & 51.65$\pm$3.19 & 45.37$\pm$0.00 & 33.21$\pm$2.73 & 54.24$\pm$2.96 & 53.56$\pm$0.72 & {\bf 64.96$\pm$0.66}\\
    & sci.med & 39.89$\pm$1.90 & 51.69$\pm$0.00 & 32.02$\pm$3.27 & 63.48$\pm$1.64 & 65.62$\pm$0.45 & {\bf 73.90$\pm$0.45}\\
    & sci.space & 37.12$\pm$1.18 & 59.51$\pm$0.00 & 32.97$\pm$2.01 & 75.09$\pm$2.18 & 76.06$\pm$1.13 & {\bf 81.51$\pm$0.61}\\
    & soc.religion.christian & 37.47$\pm$1.58 & 62.21$\pm$0.00 & 40.12$\pm$6.40 & 76.40$\pm$2.17 & 71.70$\pm$1.39 & {\bf 78.83$\pm$0.63}\\
    & talk.politics.guns & 32.83$\pm$1.22 & 72.04$\pm$0.00 & 30.75$\pm$3.50 & 82.08$\pm$2.11 & 81.78$\pm$0.36 & {\bf 85.33$\pm$0.13}\\
    & talk.politics.mideast & 32.75$\pm$2.12 & 71.31$\pm$0.00 & 31.91$\pm$1.96 & 85.94$\pm$2.04 & 81.87$\pm$0.33 & {\bf 86.19$\pm$0.22} \\
    & talk.politics.misc &33.10$\pm$2.06 & 69.18$\pm$0.00 & 27.86$\pm$1.10 & 80.16$\pm$2.93 & 78.86$\pm$0.40 & {\bf 82.97$\pm$0.47}\\
    & talk.religion.misc & 36.14$\pm$2.19 & 62.59$\pm$0.00 & 32.33$\pm$3.58 & 73.37$\pm$2.50 & 73.69$\pm$1.48  & {\bf 80.12$\pm$0.77}\\\cline{2-8} 
    & {\it average} & 43.77$\pm$0.21& 58.49$\pm$0.00 & 35.93$\pm$0.62 & 73.55$\pm$2.09 & 71.18$\pm$0.22 & {\bf 78.13$\pm$0.11}\\ \midrule
    \multirow{5}{*}{Reuters-21578}& exchanges & 62.72$\pm$2.18 & 66.11$\pm$0.00 & 55.81$\pm$7.96 & 83.71$\pm$3.17 & 80.48$\pm$0.89 & {\bf 87.02$\pm$0.48}\\
    & organizations & 53.76$\pm$3.25 & 34.05$\pm$0.00 & 44.81$\pm$4.25 & 38.47$\pm$2.70 & 33.05$\pm$0.21 & {\bf 49.14$\pm$0.63}\\
    & people & 34.28$\pm$2.33 & 67.09$\pm$0.00 & 44.57$\pm$7.30 & 78.33$\pm$6.13 & 80.38$\pm$0.66  & {\bf 86.85$\pm$0.40}\\
    & places & 33.34$\pm$0.93 & 66.53$\pm$0.00 & 44.53$\pm$5.87 & 74.31$\pm$8.89 & 79.54$\pm$0.84 & {\bf 85.62$\pm$0.31}\\
    & topics & 26.38$\pm$1.08 & 78.00$\pm$0.00 & 43.00$\pm$9.53 & 89.82$\pm$0.51 & 89.74$\pm$0.35  & {\bf 90.97$\pm$0.30}\\ \cline{2-8} 
    & {\it average} & 42.10$\pm$0.92 & 62.36$\pm$0.00 & 46.54$\pm$2.09 & 72.93$\pm$4.28 & 72.64$\pm$0.15 & {\bf 79.92$\pm$0.19}\\ \bottomrule
    \end{tabular}%
    }
    \caption{AUPR (\%) on text datasets for UAD when $p=0.5$. The best performance is in bold.}
\end{table*}

\end{document}